\documentclass[eat,twocolumn]{jmlr}%
\usepackage{longtable}%

\usepackage{booktabs}
\usepackage{adjustbox}
\usepackage{float}

\usepackage{cprotect}
\theorembodyfont{\upshape}
\theoremheaderfont{\scshape}
\theorempostheader{:}
\theoremsep{\newline}

\jmlrvolume{1}
\firstpageno{1}

\jmlryear{2022}
\jmlrworkshop{Machine Learning for Health (ML4H) 2022}

\title[Model Evaluation in Medical Datasets Over Time]{Model Evaluation in Medical Datasets Over Time}

  \author{\Name{Helen Zhou\nametag{\thanks{equal contribution}}} \Email{hlzhou@andrew.cmu.edu}\\
  \Name{Yuwen Chen}$^{*}$ \Email{yuwenc2@andrew.cmu.edu}\\
  \Name{Zachary C. Lipton} \Email{zlipton@andrew.cmu.edu}\\
  \addr Carnegie Mellon University, Pittsburgh, PA, USA}

\begin{document}

\maketitle
\vspace{-2em}
\begin{abstract}
Machine learning 
models deployed in healthcare systems
face data drawn from continually evolving environments.
However, 
researchers proposing 
such models typically evaluate them
in a time-agnostic manner, 
with train and test splits
sampling patients throughout the entire study period.
We introduce the Evaluation on Medical Datasets Over Time (EMDOT) framework and Python package, 
which evaluates the performance of a model class over time. 
Across five medical datasets and a variety of models, we
compare
two training strategies:
(1) using all historical data, 
and
(2) using a window of the most recent data. 
We note 
changes in performance over time, and identify possible explanations for these shocks.
\end{abstract}

\section{Introduction}
\label{sec:intro}
As medical practices, healthcare systems, and community environments evolve over time, so does the distribution of collected data. Features are deprecated as new ones are introduced, data collection may fluctuate depending on hospital policies, and the underlying 
populations 
may shift. 

Amidst this ever-changing data landscape, models that perform well on one time period 
cannot 
be assumed to perform well in perpetuity. 
In the MIMIC-III critical care dataset, 
\citet{nestor2019feature} found that a 
change to the electronic health record (EHR) system 
in 2008 
coincided with %
sudden 
degradations 
in AUROC for mortality prediction. 
Shifts over time may also occur in a more gradual, continuous fashion.
In 
COVID-19 data
from the Centers for Disease Control and Prevention (CDC), 
\citet{cheng2021unpacking} noted that the age distribution among 
cases shifted continually throughout the pandemic, 
and that these continual shifts 
confounded estimates of improvements in mortality rate. 
We propose an evaluation framework to characterize model performance over time by simulating 
training procedures that practitioners could have executed up to each time point, and subsequently deployed in future time points. %
We argue that
standard time-agnostic evaluation is insufficient for selecting deployment-ready models, showing
that 
it 
over-estimates deployment performance. 
Instead, we advocate for EMDOT as a worthwhile pre-deployment step 
to help practitioners
gain confidence in the robustness of their models to distribution shifts that have happened in the past and may to some extent repeat in the future. 

While intuitive, evaluation of models over time remains far from standard practice in the development of machine learning models for healthcare (ML4H). One possible reason for this is lack of access---as noted by \citet{nestor2019feature}, 
it is common practice to remove timestamps 
when
de-identifying medical datasets for public use.
In this work, we identify five sources of medical data containing varying granularities of 
temporal information per-record, 
four of which are \emph{publicly available}.
We profile the performance of various training strategies and model classes across time, and 
identify possible sources of distribution shifts 
within each dataset.
We release the Evaluation on Medical Datasets Over Time (EMDOT) Python package\footnote{github.com/acmi-lab/EvaluationOverTime} (details in Appendix \ref{app:emdot_appendix})
to allow researchers to apply EMDOT
to their own datasets and
test techniques for
handling
shifts over time.

\section{Related work}
The promise of ML 
for improving healthcare 
has been explored in several domains, 
from 
cancer survival prediction \citep{hegselmann2018reproducible}, 
to diabetic retinopathy detection \citep{google_diabetic_retinopathy2016}, 
to antimicrobial stewardship \citep{kanjilal2020decision,boominathan2020}, 
to mortality prediction in liver transplant candidates \citep{
byrd2021predicting}.
Typically, these ML models are evaluated on a random held-out set of patients, and sometimes externally validated on other hospitals or newly collected data. However, what makes a model trustworthy enough to be deployed in the wild?
The medical community has a long history of utilizing (mostly) fixed, simple risk scores to inform patient care \citep{six2008chest,kamath2001model,wilson1998prediction,wells1995accuracy}. Risk scores often prioritize ease-of-use, are computed from 
few variables, verified by domain experts for clear causal 
connections 
to outcomes of interest,
and validated through use over time and across hospitals. Together, these factors give clinicians confidence that the model will perform reliably for years to come.
With increasingly complex models, however, trust and adoption 
may be hindered by a lack of 
confidence in robustness to changing environments.

As noted by \citet{d2020underspecification}, ML models often exhibit unexpectedly poor behavior when deployed in real-world domains. A key reason for this,
they argue, is \emph{under-specification}, where ML pipelines yield many predictors with equivalently strong held-out performance in the training domain, but such predictors can behave very differently in deployment. 
EMDOT could stress test models to help combat under-specification.

Although evaluation over time is far from standard in ML4H literature, some have begun to explore this idea.
To predict wound-healing, \cite{jung2015implications} 
found that when data were split by cutoff time instead of patients, benefits of model averaging and stacking disappeared.
\citet{pianykh2020continuous} found degradation in performance of a model for wait times dependent on how much historical data was trained on. To predict severe COVID-19, \citet{ahn_clinical_concepts} found that learned clinical concept features
performed more robustly over time than raw features. 

Closest to our work is \citet{nestor2019feature}, 
which
evaluated %
AUROC in MIMIC-III from 2003--2012, comparing training on 
just 2001--2002; the prior year; and the full history.
Using the full history and manual aggregations of clinical concepts, they bridged a big 
drop in performance
due to changing EHR systems.
Whereas \citet{nestor2019feature} %
considers 
three models %
per test year, EMDOT simulates model deployment every year and evaluates %
across \emph{all future years}. 

EMDOT shares
similarities to 
techniques for
cross-validation in time series forecasting \citep{BERGMEIR2012192,cerqueira2020evaluating}, but
instead of just reporting 
summary statistics 
that could conceal 
shifts over time, 
EMDOT
seeks to elucidate 
the nature and potential causes of 
fluctuating performance over time. Also, we evaluate models treating data as i.i.d. (not longitudinal).

\begin{table*}[t]
\caption{Summary of datasets used for analysis. Fore more details, see Appendices \ref{app:seer_data}--\ref{app:optn_data}.}
\label{tab:dataset_info}
\centering
    \resizebox{2.0\columnwidth}{!}{%
    \begin{tabular}{lccccc}
    \toprule
    Dataset name        &  Outcome          & Time Range (time point unit) & \# samples & \# positives \\
    \midrule
    SEER (Breast)  & 5-year Survival   & 1975--2013 (year)            & 462,023   & 378,758   \\
    SEER (Colon)   & 5-year Survival   & 1975--2013 (year)            & 254,112   & 135,065   \\
    SEER (Lung)    & 5-year Survival   & 1975--2013 (year)            & 457,695   & 49,997 \\
    CDC COVID-19        & Mortality         & Mar 2020--May 2022 (month)  & 941,140   & 190,786  \\
    SWPA COVID-19         & 90-day Mortality  & Mar 2020--Feb 2022 (month)  & 35,293    & 1,516  \\
    MIMIC-IV            & In-ICU Mortality  & 2009--2020 (year)            & 53,050    & 3,334  \\
    OPTN (Liver)          & 180-day Mortality &  2005--2017 (year)           & 143,709   & 4,635 \\
    \bottomrule
    \end{tabular}%
    }
\end{table*}

\section{Data}
We sought medical datasets that had (1) a timestamp for each record, (2) interesting prediction task(s), and (3) enough distinct time points to evaluate over. 
Five datasets 
satisfied these criteria:
SEER cancer data, national CDC COVID-19 data, COVID-19 data from a major provider in Southwestern Pennsylvania (SWPA), MIMIC-IV critical care data, and OPTN liver transplant data. All but SWPA are publicly accessible.

\begin{figure}%
\includegraphics[width=1.0\columnwidth]{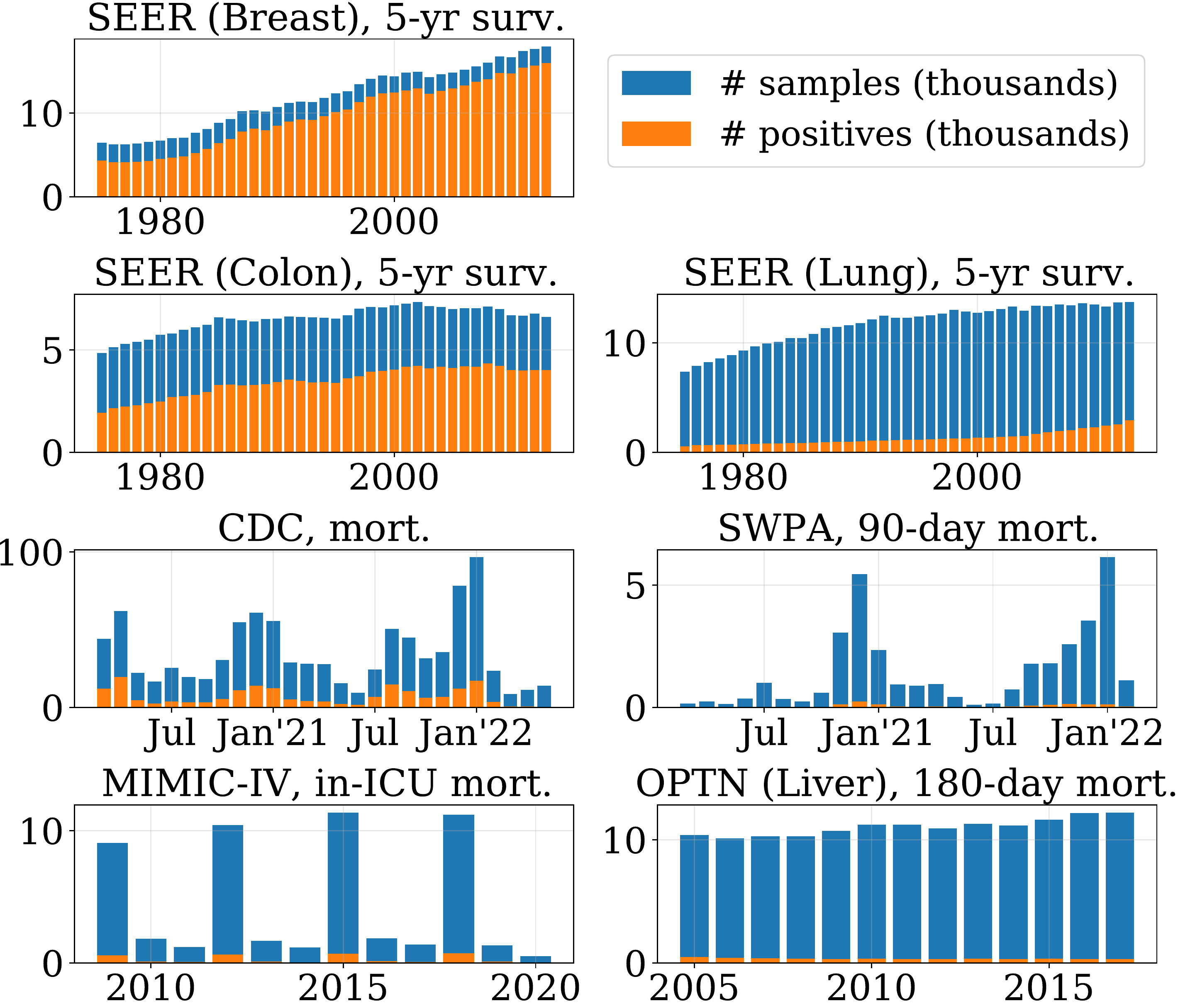}
\vspace{-2em}
\caption{Number of samples and positive outcomes per time point.}
\vspace{-1em}
\label{fig:bar_plot_number_of_samples_pos_outcomes}
\end{figure}

Table \ref{tab:dataset_info} summarizes the data. Figure \ref{fig:bar_plot_number_of_samples_pos_outcomes} visualizes 
data quantity over time.
Appendices \ref{app:seer_data}--\ref{app:optn_data} 
include cohort selection diagrams, cohort characteristics, features, heat maps of missingness, preprocessing steps, 
and additional details. Finally, each sample corresponds to a distinct patient.

\section{Methods}

We compare the \emph{reported} test performance of a model if it were trained using standard (all-period) time-agnostic splits to more realistic performance from simulating how a practitioner might have trained and deployed models in the past (evaluation over time). We consider both linear and more complex model classes, as well as two training regimes: (1) when models are trained on a recent window of the past, and (2) models are trained on all historical data. Finally, we describe diagnostic plots, which help detect reasons for changing model performance.

\paragraph{Prediction Task} All tasks are binary classification, evaluated by AUROC. %
Samples are treated in an `i.i.d' manner for training. 

\paragraph{All-Period Training}\label{sec:standard_split}
We mimic common practice by 
using time-agnostic data splits, which randomly place samples from the entire study time range into train, validation, and test sets (Appendix \ref{app:datasplit_details}).

\paragraph{Evaluation over Time Training}\label{sec:eot}

To simulate 
how practitioners train models 
and subsequently deploy 
them on 
future data, we define the EMDOT framework. At each time point (``simulated deployment date"), an \emph{in-period} set of data is available for model development. Evaluation is on both recent in-period and future \emph{out-of-period} data. 

\begin{figure}[t]
  \includegraphics[width=1.0\columnwidth]{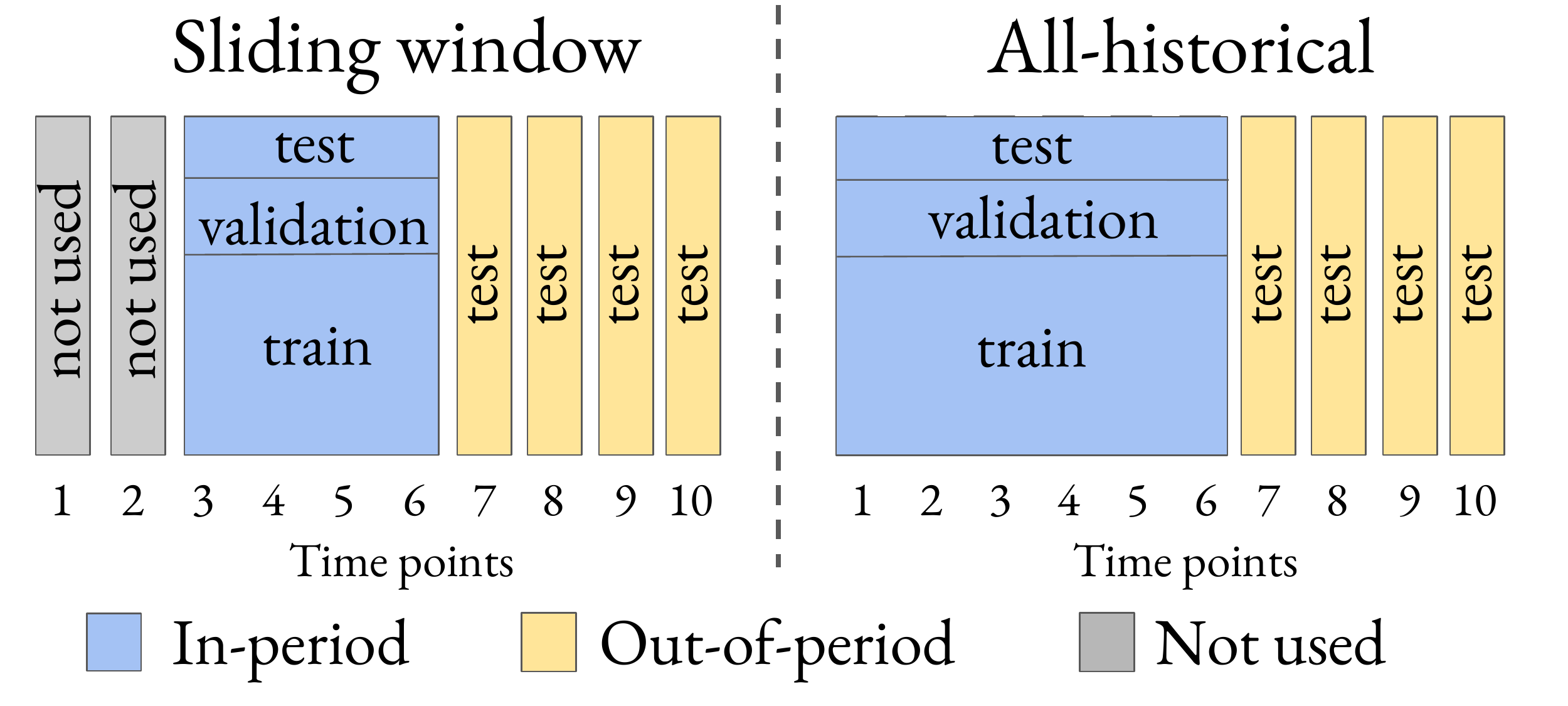}
  \vspace{-1.5em}
  \caption{EMDOT training regimes.}
  \vspace{-1em}
  \label{fig:training_regimes}
\end{figure}

In-period data is split into train, validation, and test sets 
(ratios and hyperparameters in Appendices \ref{app:datasplit_details} and \ref{app:hyperparameter_grids}). 
Recent in-period performance is evaluated on held-out test data from the most recent time point. Out-of-period performance is evaluated on all 
data from 
each future time point. E.g., a model trained up to time 6 is tested on data from 6, 7, 8, etc. (Figure \ref{fig:training_regimes}), and at time 8, the model is two time points \emph{stale}. 
This procedure can take $\leq T$ times more computation than all-period training for $T$ time points. 
Practitioners face a tradeoff between using %
recent data reflective of the present %
and using 
all available 
historical data for a larger sample size. %
Intuitively, the former may be appealing in modern applications with massive datasets, whereas the latter may be necessary in data-scarce applications. We explore these two training regimes, with different definitions of in-period data (Figure \ref{fig:training_regimes}): 
\begin{enumerate}
    \item \textbf{Sliding window}: The last $W$ time points are in-period. Here, $W$ = 4 for sufficient positive examples.%
    \item \textbf{All-historical}: Any data prior to the current time point is in-period.
\end{enumerate}

To decouple the effect of sample size from that of distribution shifts, we also perform comparisons with all-historical data that is \textbf{sub-sampled} to be the same size as the corresponding sliding window training set.

\paragraph{Models}
Logistic regression (LR), gradient boosted decision trees (GBDT) and feedforward neural networks (MLP) are compared. 

\paragraph{Detecting Sources of Change}
To better understand possible reasons for changing performance,
we create \emph{diagnostic plots} to track
importances and average 
values of the most important features over time. 
We 
generate these plots for logistic regression and define feature importance by the %
magnitudes of the coefficients, but note that other feature importance techniques 
could be used for more complex model classes. By highlighting sudden changes in the performance and the corresponding time periods, diagnostic plots surface shifts in the distribution of data that coincide with changing model performance.

\section{Results}
Here, we focus on three datasets with interesting results: SEER (Colon), CDC, and SWPA. Full results are in the Appendix.

\paragraph{All-period Training} In standard time-agnostic evaluation, on all datasets except MIMIC-IV, GBDT and MLP perform best (Table \ref{tab:AUROC_splitting_by_patient_abstract}, Appendix Table \ref{tab:AUROC_splitting_by_patient}). 
Top 10 coefficients of LR models are in Appendices \ref{app:seer_data}--\ref{app:optn_data}.

\begin{table}[htbp]
\caption{Test AUROC in all-period training.}
\label{tab:AUROC_splitting_by_patient_abstract}
\centering
\vspace{-0.5em}
\resizebox{0.9\columnwidth}{!}{%
    \begin{tabular}{lccc}
    \toprule
    Model & SEER (Colon) & CDC & SWPA \\
    \midrule
    \textbf{LR} &      0.867 &        0.837 &         0.914\\
    \textbf{GBDT} &0.871 &       \textbf{0.850} &         \textbf{0.926}\\
    \textbf{MLP} &      \textbf{0.873} &        0.844 &         0.918\\
    \bottomrule
    \end{tabular}%
    }
\end{table}

\begin{figure}[ht]
\centering
  \includegraphics[width=0.9\columnwidth]{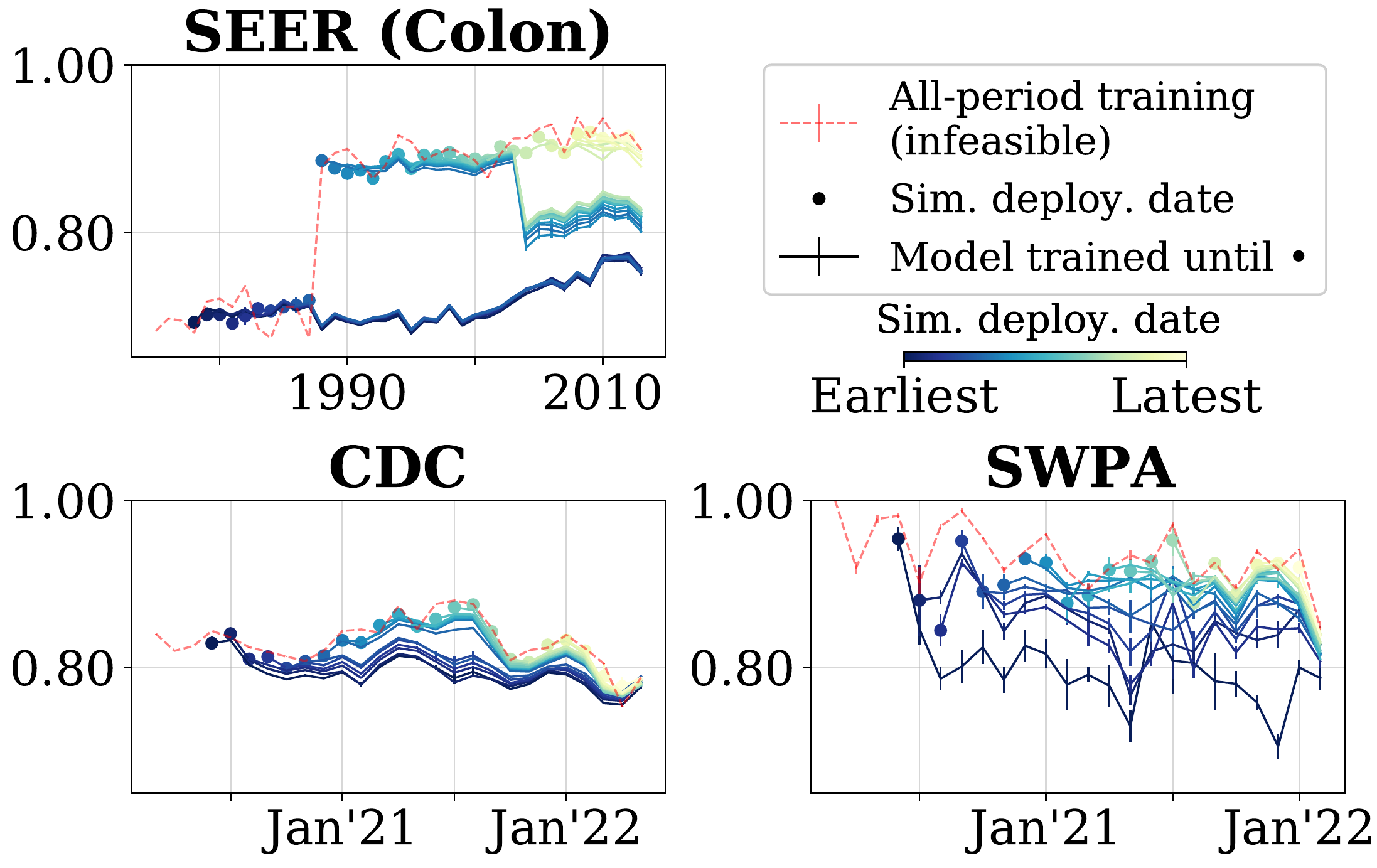}
  \caption{LR average test AUROC vs. time. 
  Solid line 
  = model from a sim. deploy. date (dot), %
  evaluated on future times. 
  Error bars are std. dev. 
  over 5 random splits. %
  Dotted line is infeasible, as it trains on data after sim. deploy. date. 
  }
  \label{fig:absolute_auc_over_time_abstract}
  \vspace{-1em}
\end{figure}

\noindent\textbf{Evaluation Over Time}\hspace{0.5em}~
Figures \ref{fig:absolute_auc_over_time_abstract} and \ref{fig:absolute_auc_over_time} (Appendix) plot AUROC of LR over time using all-historical data. (GBDT, MLP, AUPRC, F1, and accuracy in Appendix \ref{app:alternative_metrics}.)

The test performance of a model from standard all-period training (red dotted line) mostly sits above the performance of any model that could have been realistically been deployed by that date. Thus, all-period training tends to provide an over-optimistic estimate of performance at deployment.

In SEER data, AUROC increases dramatically near 1988, but several of the subsequent models see a large drop
around 2003 (Figure \ref{fig:absolute_auc_over_time_abstract}).
By contrast, in CDC data, model performance 
is relatively smooth over time.
Models trained after December 2020 
have a slight boost in AUROC, 
coinciding with
a surge in cases (and hence sample size, Figure \ref{fig:bar_plot_number_of_samples_pos_outcomes}).
In SWPA COVID-19, there is more variation 
and uncertainty 
in AUROC 
early in the pandemic,
where sample sizes are small. 
In December 2020, sample sizes increase, and models seem to become more robust to changes over time. 
In all data, in-sample test AUROCs 
tend to 
increase over time. 

\begin{figure}[ht]
\centering
  \includegraphics[width=0.98\columnwidth]{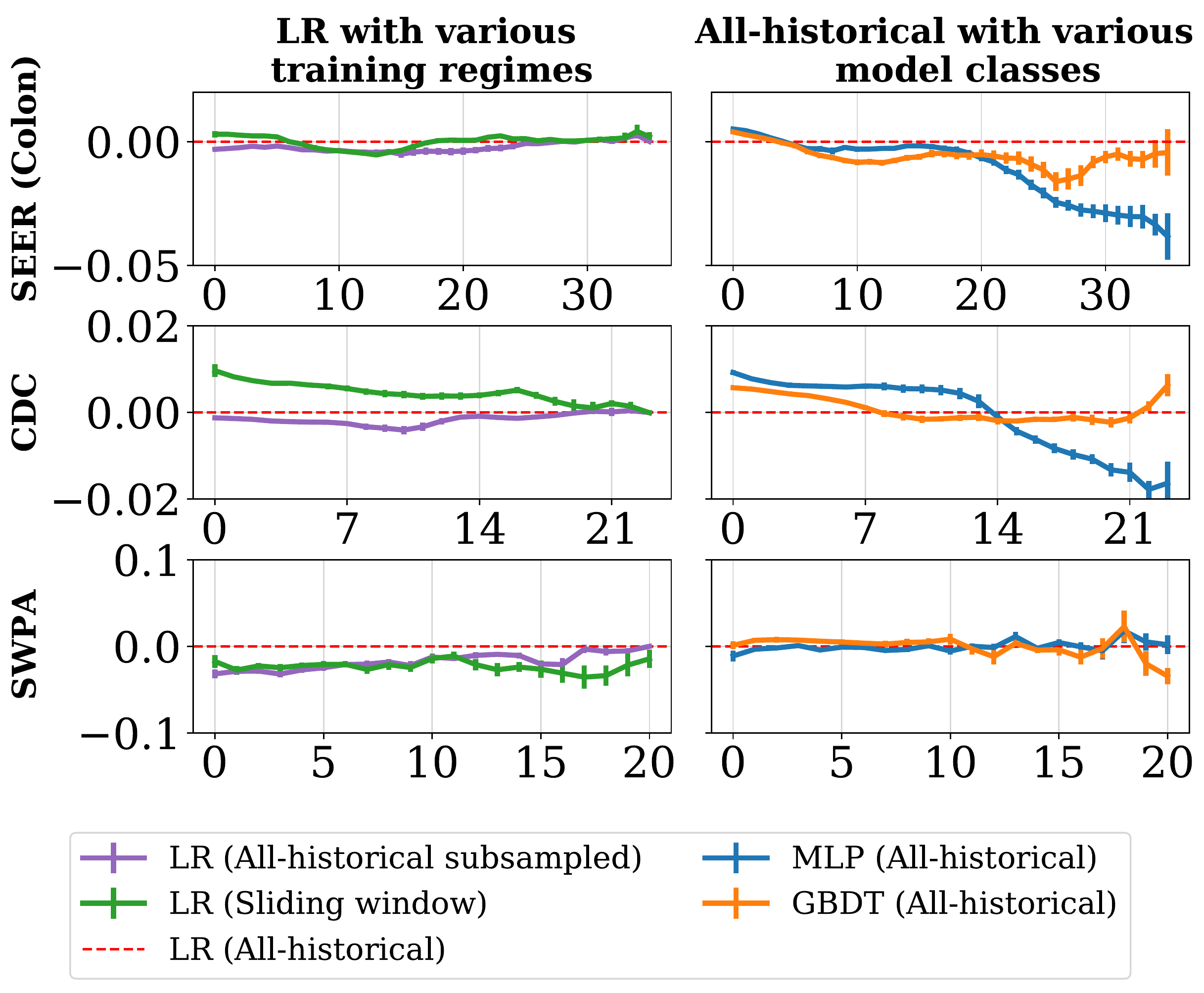}
  \vspace{-0.5em}
  \caption{
  $\text{AUROC} - \text{AUROC}_{\text{LR All-historical}}$ 
  vs. staleness, with various training regimes (left) and model classes (right). Error bars are $\pm$ std. dev.}
  \label{fig:window_model_comparison_abstract}
  \vspace{-1.5em}
\end{figure}

\noindent\textbf{Training Regime Comparison}\hspace{0.5em}~
Training regimes perform differently depending on the dataset (Figure 
\ref{fig:window_model_comparison_abstract} and Appendix Figure \ref{fig:window_model_comparison}, left).
In CDC COVID-19, sliding window outperforms %
all-historical
across all stalenesses.
By contrast, in SWPA COVID-19, which has the least amount of data (Table \ref{tab:dataset_info}), both sliding window and all-historical (subsampled) underperform all-historical.
In SEER (Colon), performance is relatively stable regardless of training regime. 

\noindent\textbf{Model Comparison}\hspace{0.5em}~
In SEER (Colon)
and CDC COVID-19, both GBDT and MLP initially outperform LR when staleness 
is 
$<$~4 years and $<$~7 months (respectively),
but both eventually underperform LR as staleness increases further (Figure 
\ref{fig:window_model_comparison_abstract}, right). 
In SWPA COVID-19, 
all model classes
perform comparably over time. In MIMIC-IV, LR 
remained the best (Appendix Figure \ref{fig:window_model_comparison}, right).

\vspace{-0.5em}

\paragraph{Detecting Possible Sources of Change}
\begin{figure}[ht]
  \includegraphics[width=1\columnwidth]{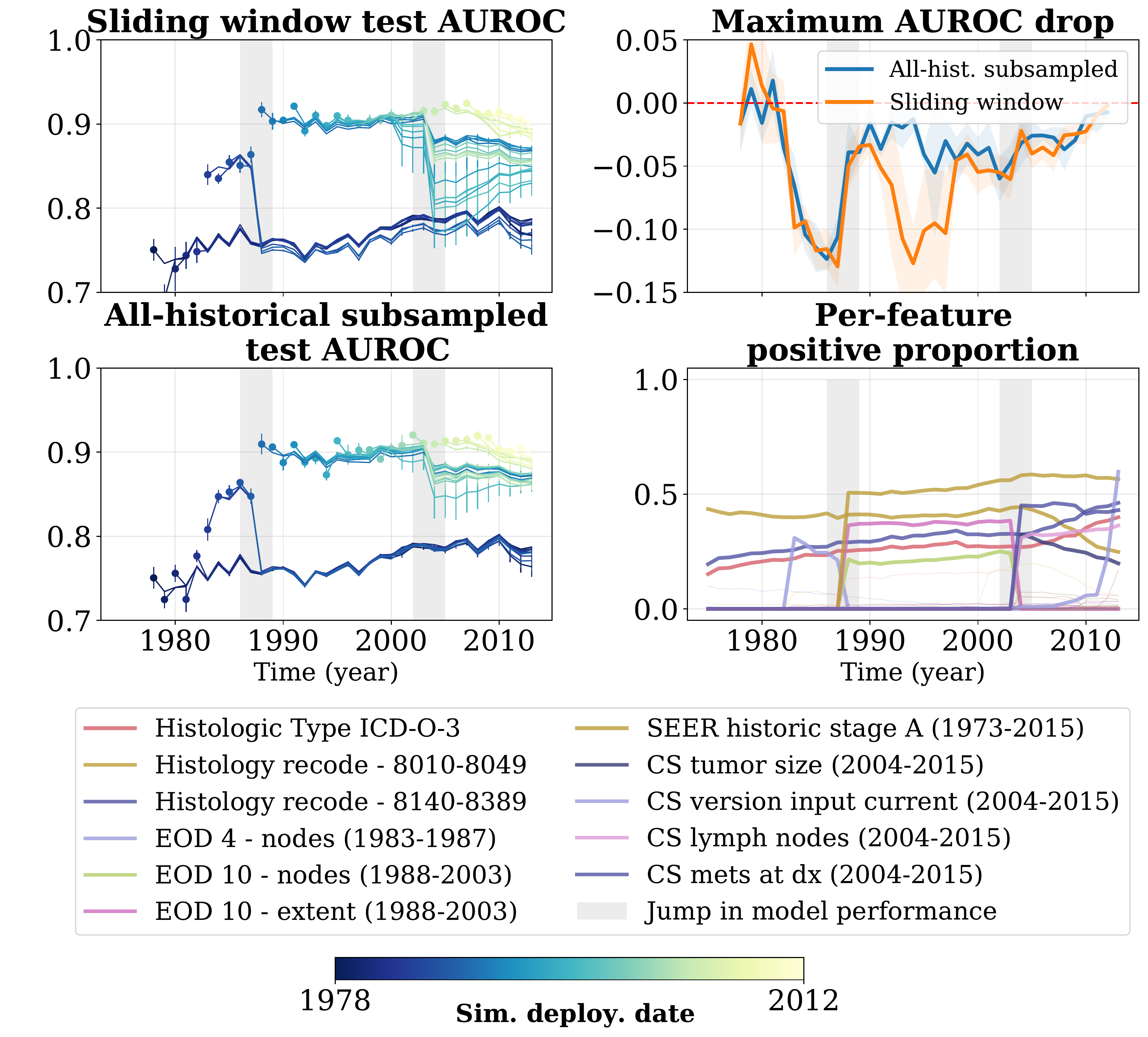}
  \vspace{-1em}
  \caption{SEER (Lung) diagnostic plots. AUROC vs.~time (left), max.~drop in AUROC for each simulated deployment time (top right), and prevalences of the most important features over time (bottom right).}
  \label{fig:diagnositic_plot_seer_lung_main_paper}
  \vspace{-1em}
\end{figure}
Diagnostic plots for all datasets are 
in Appendix \ref{app:diagnostic_plot}. Here, we discuss SEER (Lung) (Figure \ref{fig:diagnositic_plot_seer_lung_main_paper})
in detail.
In 1983, as \verb|EOD 4| features (extent of disease coding schema) are introduced, 
a sudden jump in AUROC occurs.
However, models trained at this time later experience a large %
AUROC drop.
By 1988, \verb|EOD 4| is phased out, and \verb|EOD 10| features are introduced.  
This coincides with another jump in AUROC, 
sustained until 2003 when the \verb|EOD 10| features are removed. %
In this dataset, the 
all-historical training regime seems more robust to changes over time,
as all-historical models trained after 1988 
avoid the %
drop 
that sliding window models undergo once their window excludes pre-1988 data.

\section{Discussion}
As standard all-period training 
evaluates performance on time points already seen in the training set,  
the reported performance
tends to be over-optimistic (Figure \ref{fig:absolute_auc_over_time_abstract}) and 
does not reflect 
degradation that would have occurred in deployment. 
Comparing model classes, in all datasets except MIMIC-IV, GBDT and MLP outperform 
LR %
under standard time-agnostic evaluation
(Appendix Table \ref{tab:AUROC_splitting_by_patient}). However, evaluated across time, 
LR is often comparable and even outperforms more complex models 
once enough time passes after the simulated deployment date (e.g. LR eventually overtaking MLP in SEER Breast, Colon, and Lung; Figure \ref{fig:window_model_comparison_abstract}). 

Sample size can influence which training regime performs better.
In SWPA COVID-19 (smallest dataset), the sliding window and sub-sampled all-historical training regimes perform comparably, both underperforming all-historical (Figure \ref{fig:window_model_comparison_abstract}). 
Here, the benefit of larger sample size in all-historical outweighs the potential issue of stale data clouding the fresh data distribution. 
On the other hand, in CDC COVID-19 (largest dataset), subsampled all-historical performs comparably to all-historical, and sliding window outperforms both across all stalenesses (Figure \ref{fig:window_model_comparison_abstract}). 
Here, the performance of LR may have been saturated even when a sub-sample of all-historical data was used, and the benefit of using more recent data outweighs the larger sample size afforded by all-historical. 
In rapidly evolving environments, the sliding window training regime may be advantageous, as long as there is enough data.

The SEER datasets had dramatic changes in data distribution in both 1988 and 2003, when important features were added and/or removed (Figure \ref{fig:diagnositic_plot_seer_lung_main_paper}). 
One possible reason for the robustness of all-historical models in this dataset is that after 2003, when features like EOD 10 were removed, the model could still rely on features that were introduced prior to the use of EOD 10 in 1988. More broadly, we hypothesize that if a model was trained on a mixture of distributions that occurred throughout the past, it may be better equipped to handle shifts to settings 
similar to those distributions 
in the future.

In conclusion, EMDOT not only 
explores
the suitability of different model classes or training regimes for deployment, but also helps one detect distribution shifts that occurred in the past. Understanding such shifts may help practitioners be prepared for shifts of a similar nature in the future. 

\noindent\textbf{Future Work}\hspace{0.5em}
To alleviate concerns about computation time, we plan to add parallelization 
to EMDOT. 
Another interesting extension is exploring other data modalities (e.g. images). 
More broadly, we hope that others may %
build upon EMDOT to shine new light on how models
fare when evaluated with an eye towards deployment.

\bibliography{refs}

\appendix

\newpage
\onecolumn

\section{EMDOT Python Package}
\label{app:emdot_appendix}

\begin{figure}[H]
  \includegraphics[width=1.05\columnwidth]{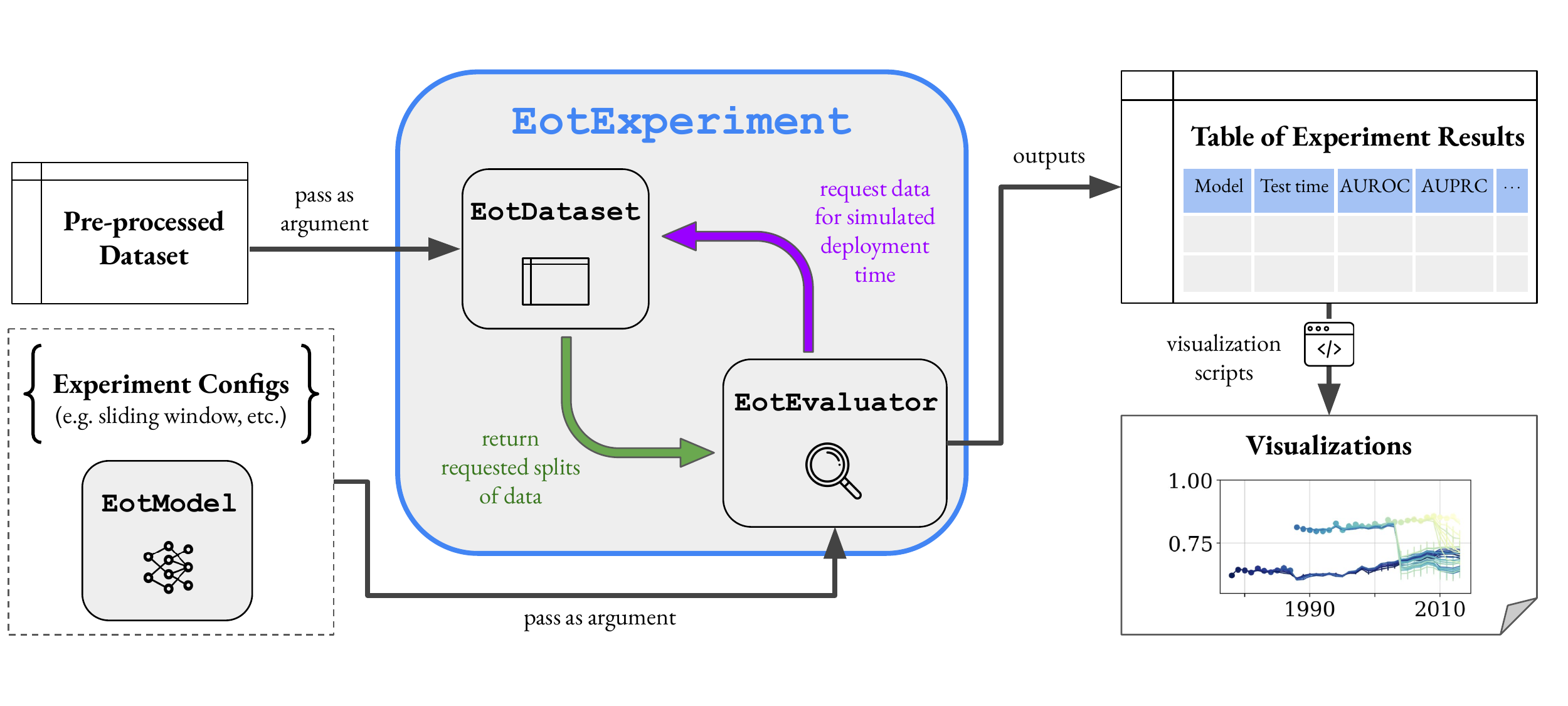}
  \cprotect\caption{
EMDOT Python package workflow diagram. The primary touchpoint of the EMDOT package is the
\verb|EotExperiment|
  object. Users provide a dataframe for their (mostly) preprocessed dataset (EMDOT takes care of normalization based on the relevant training set), their desired experiment configuration (e.g. sliding window), and model class (which should subclass the simple
  \verb|EotModel| abstract class) in order to create an \verb|EotExperiment| object. Running the \verb|run_experiment()| function of the \verb|EotExperiment| returns a dataframe of experiment results that can then be visualized. The diagram also provides insight into some of the internals of the \verb|EotExperiment| object -- there is an \verb|EotDataset| object that handles data splits, and an \verb|EotEvaluator| object that executes the main evaluation loop.
  }
  \label{fig:emdot_diagram}
\end{figure}

Figure \ref{fig:emdot_diagram} illustrates the workflow of the EMDOT Python package.

\newpage
\section{Additional SEER Data Details}\label{app:seer_data}

\begin{figure}[ht]
  \includegraphics[width=1.0\columnwidth]{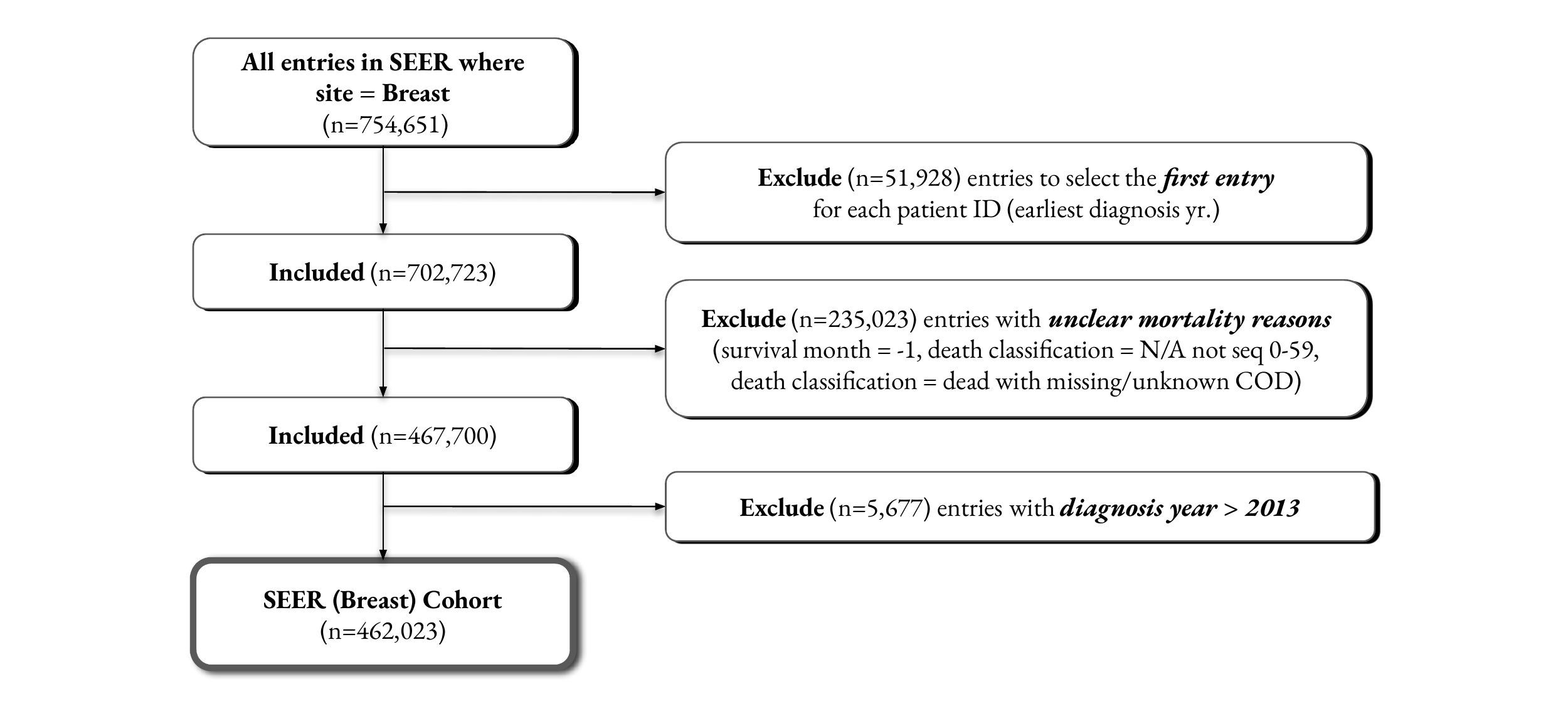}
  \caption{Cohort selection diagram - SEER (Breast)}
  \label{fig:seer_breast_cohort}
\end{figure}

\begin{figure}[ht]
  \includegraphics[width=1.0\columnwidth]{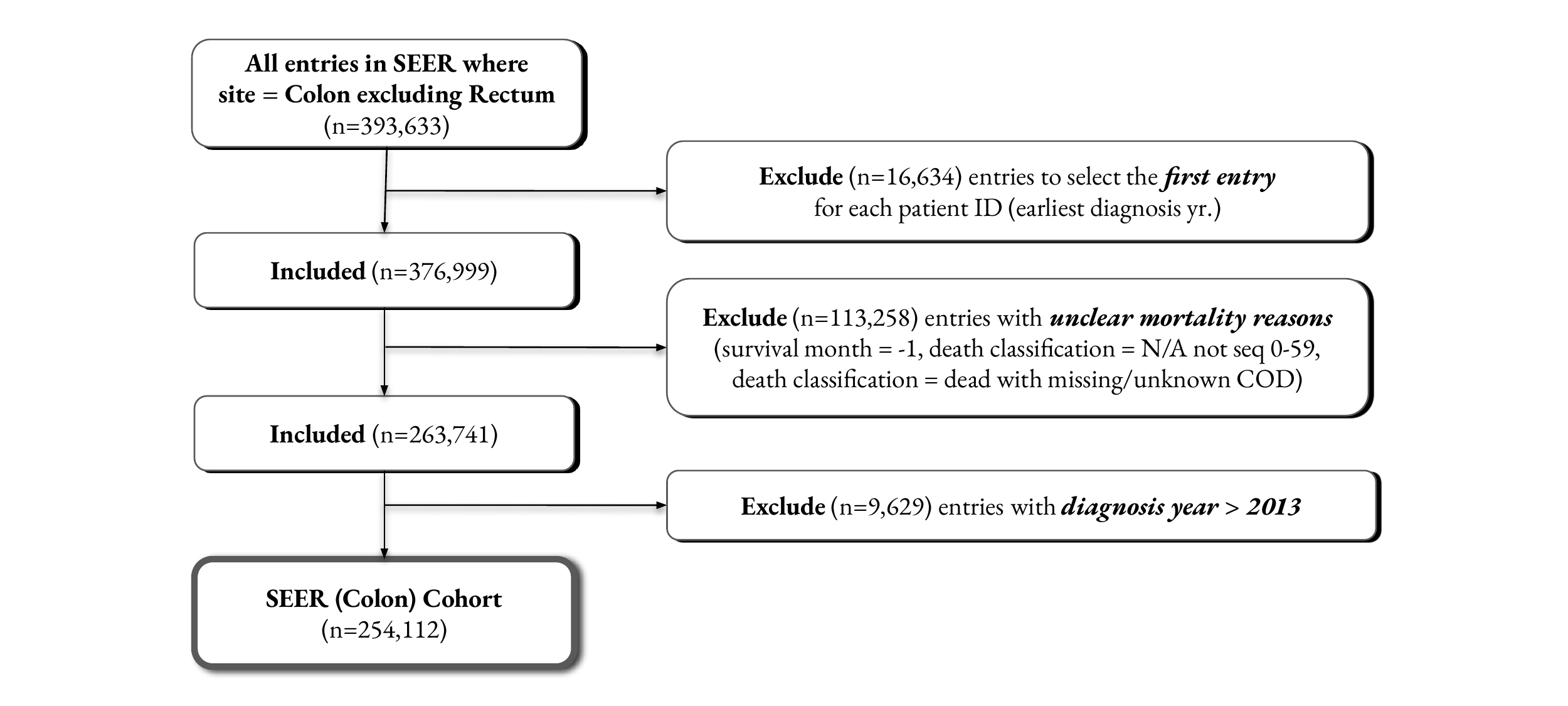}
  \caption{Cohort selection diagram - SEER (Colon)}
  \label{fig:seer_colon_cohort}
\end{figure}

\begin{figure}[ht]
  \includegraphics[width=1.0\columnwidth]{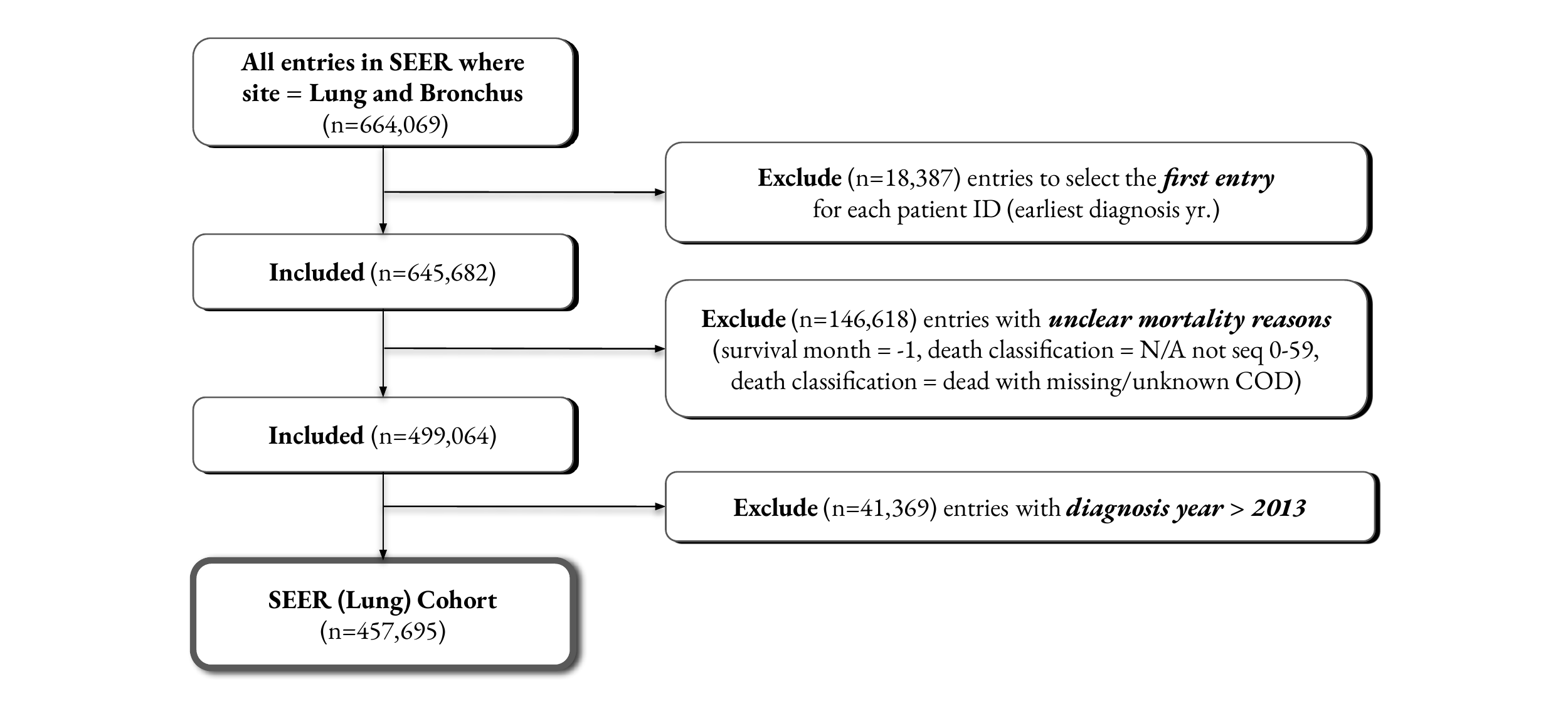}
  \caption{Cohort selection diagram - SEER (Lung)}
  \label{fig:seer_lung_cohort}
\end{figure}

\begin{figure}[ht]
  \includegraphics[width=1.0\columnwidth]{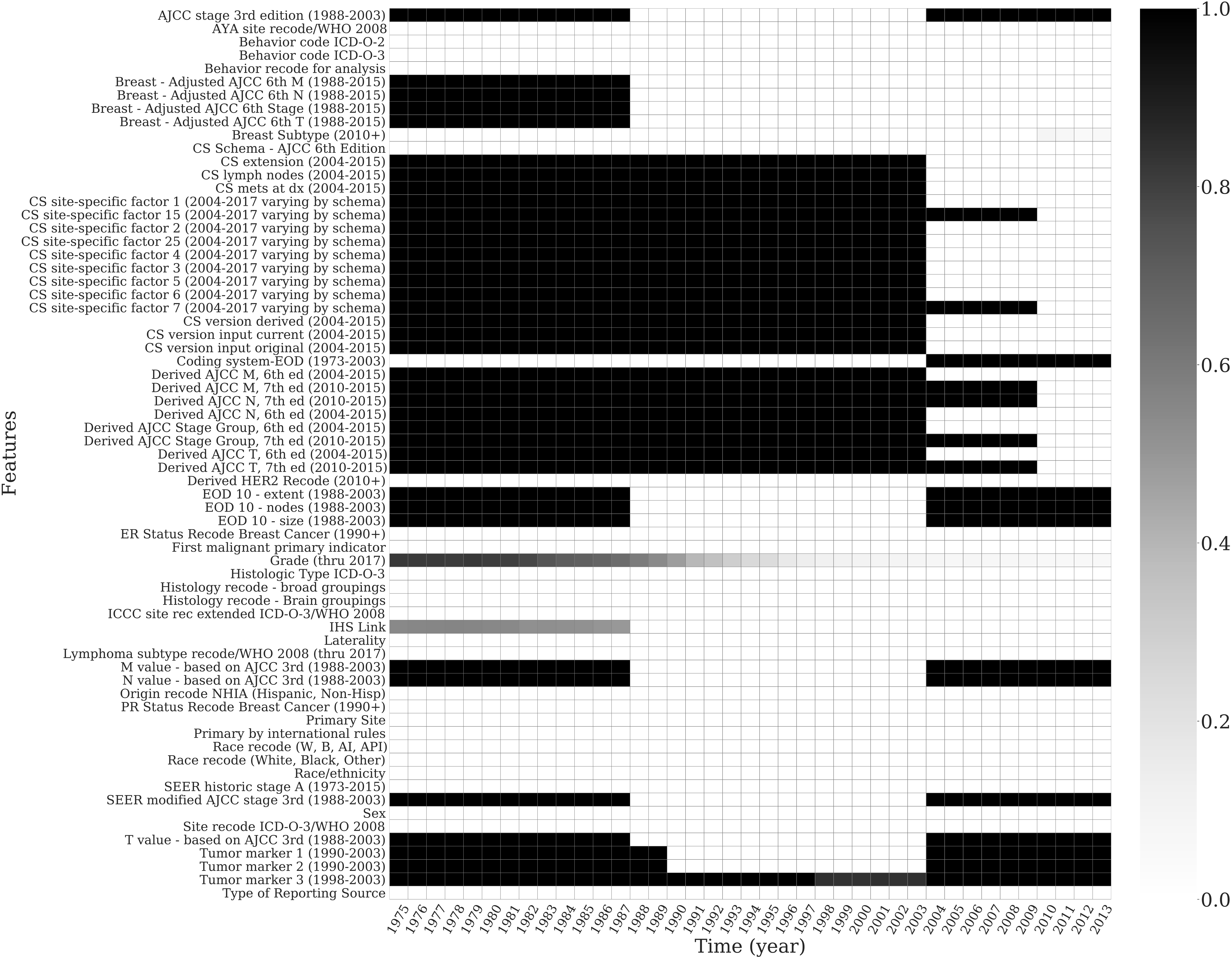}
  \vspace{-2em}
  \caption{Missingness of categorical features in SEER (Breast).}
  \label{fig:heatmap_seer_breast_cate}
\end{figure}

\begin{figure}[ht]
  \includegraphics[width=1.0\columnwidth]{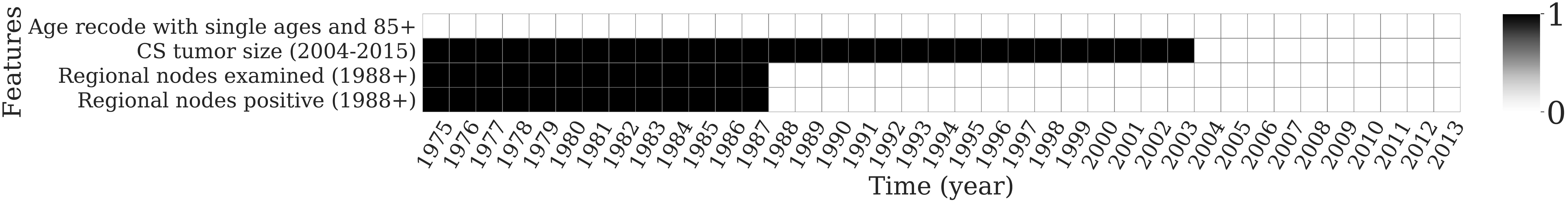}
  \vspace{-2em}
  \caption{Missingness of numerical features in SEER (Breast).}
  \label{fig:heatmap_seer_breast_num}
  \vspace{2em}
\end{figure}

\begin{figure}[ht]
  \includegraphics[width=1.0\columnwidth]{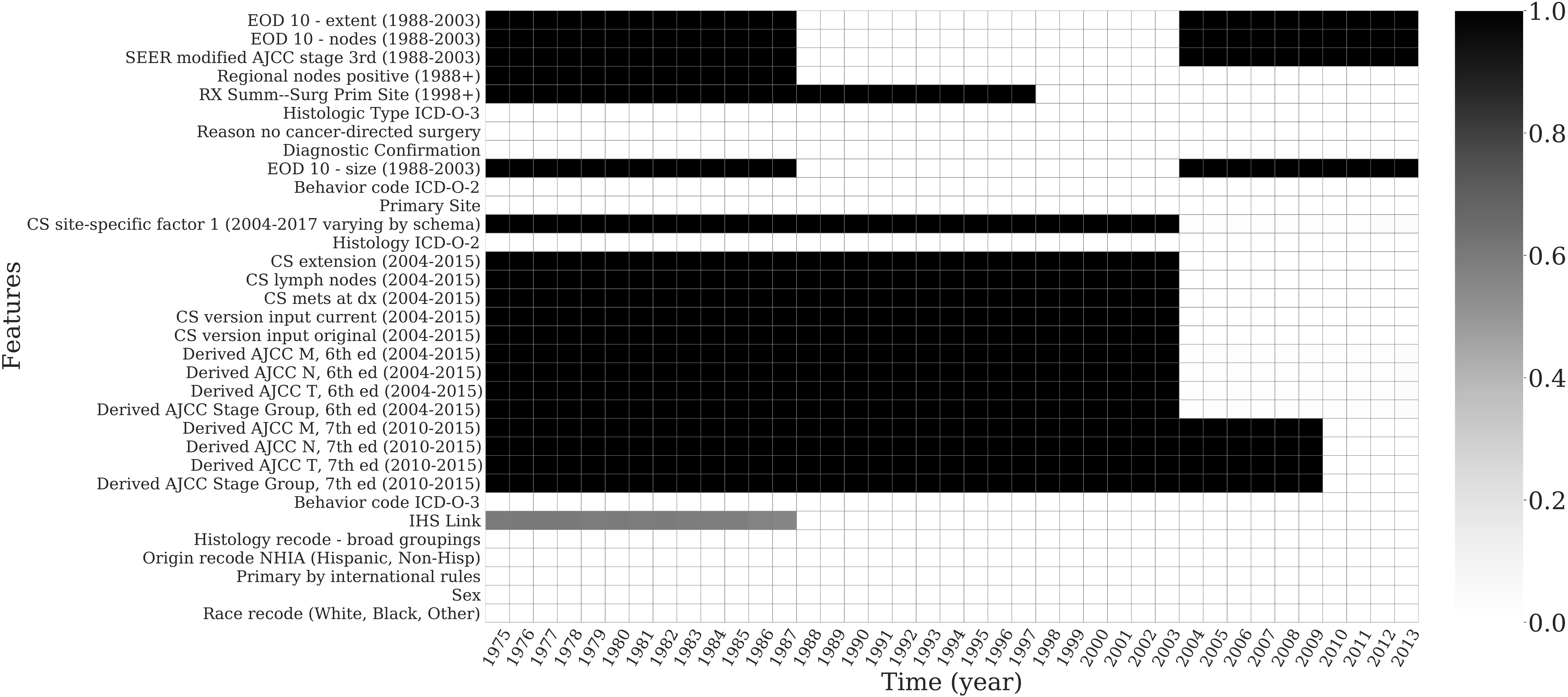}
  \vspace{-2em}
  \caption{Missingness of categorical features in SEER (Colon).}
  \label{fig:heatmap_seer_colon_cate}
\end{figure}

\begin{figure}[ht]
  \includegraphics[width=1.0\columnwidth]{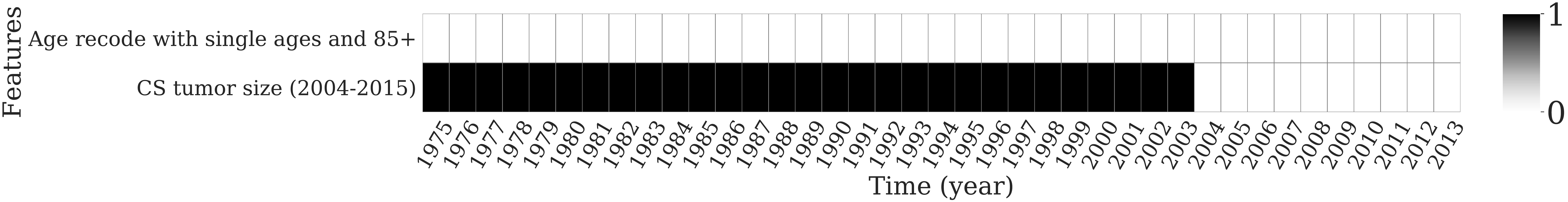}
  \vspace{-2em}
  \caption{Missingness of numerical features in SEER (Colon).}
  \label{fig:heatmap_seer_colon_num}
\end{figure}

\begin{figure}[ht]
  \includegraphics[width=1.0\columnwidth]{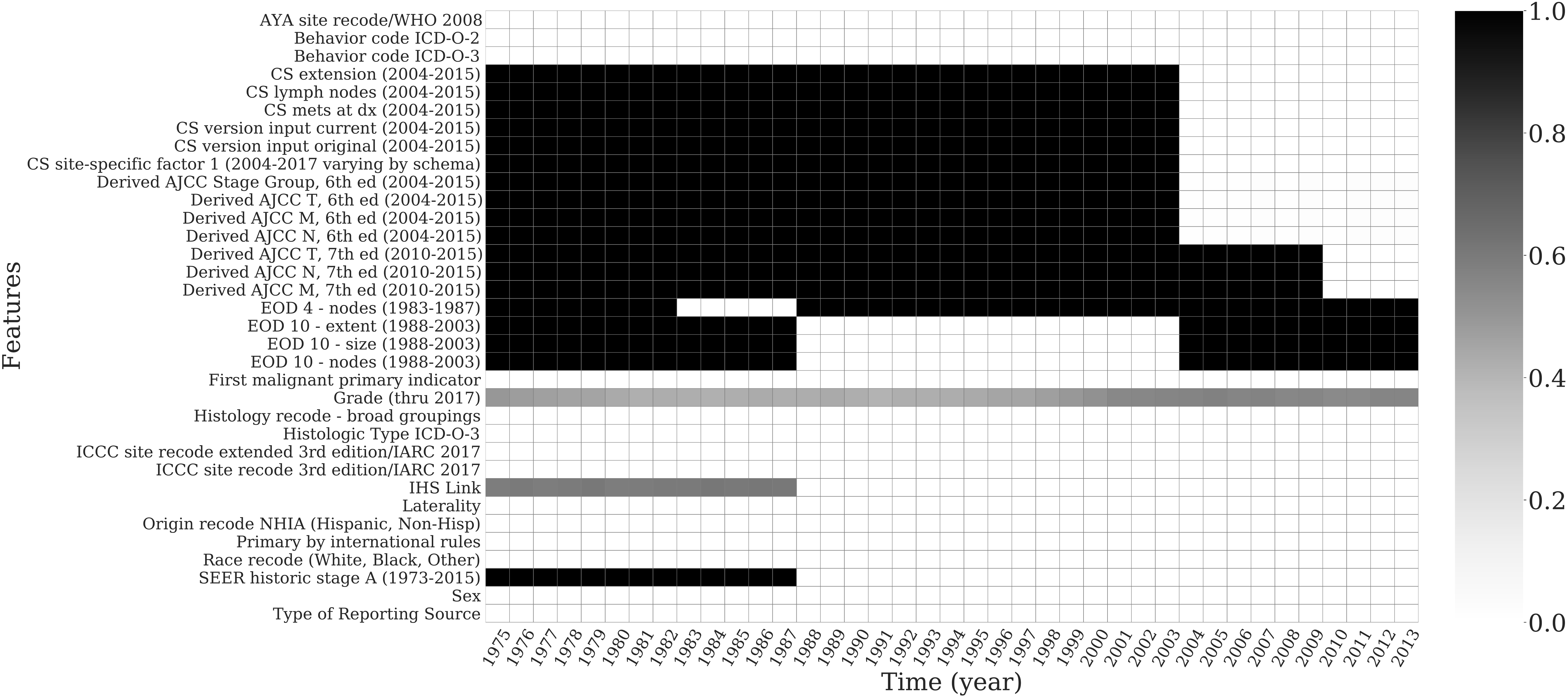}
  \vspace{-2em}
  \caption{Missingness of categorical features in SEER (Lung).}
  \label{fig:heatmap_seer_lung_cate}
\end{figure}

\begin{figure}[ht]
  \includegraphics[width=1.0\columnwidth]{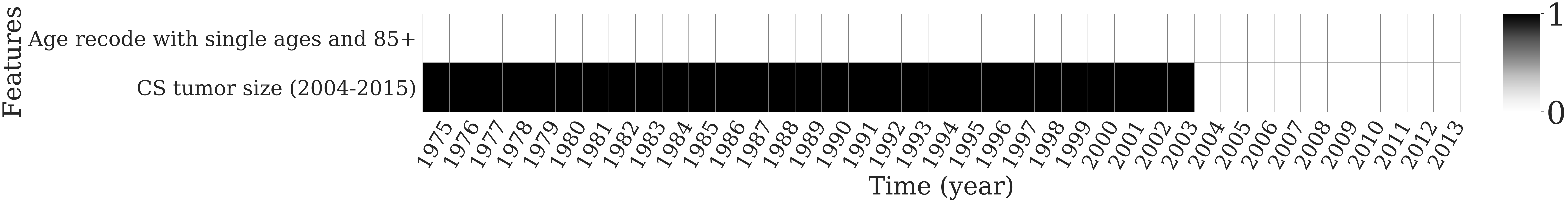}
  \vspace{-2em}
  \caption{Missingness of numerical features in SEER (Lung).}
  \label{fig:heatmap_seer_lung_num}
\end{figure}

The Surveillance, Epidemiology, and End Results (SEER) Program collects cancer incidence data from 
registries
throughout the U.S. 
This data has been used to study survival in several forms of cancer \citep{choi2008conditional,fuller2007conditional,taioli2015determinants,hegselmann2018reproducible}.
Each case includes 
demographics, primary tumor site, tumor morphology, stage and diagnosis, first course of treatment, and survival outcomes
(collected with follow-up) \citep{seerdataset}. 
The performance over time is evaluated on a \emph{yearly} basis. We use the November 2020 version of the SEER database with nine registries (SEER 9), which covers about 9.4\% of the U.S. population. While there are SEER databases that aggregate over more registries and hence cover a greater proportion of the U.S. population, we choose SEER 9 due to the large time range it covers (1975--2018). 
\begin{itemize}
    \item Data access: After filling out a Data Use Agreement and Best Practices Agreement, individuals can easily request access to the SEER dataset.
    \item Cohort selection: Using the SEER$^{*}$Stat software \citep{surveillance2015national}, we define three cohorts of interest: (1) breast cancer, (2) colon cancer, and (3) lung cancer. We primarily follow the cohort selection procedure from \cite{hegselmann2018reproducible}, but we use SEER 9 instead of SEER 18, 
and use data from all available years instead of limiting to 2004--2009. Cohort selection diagrams are given in Figures \ref{fig:seer_breast_cohort}, \ref{fig:seer_colon_cohort}, and \ref{fig:seer_lung_cohort}. If there are multiple samples per patient, we filter to the first entry per patient, which corresponds to when a patient first enters the dataset. This corresponds to a particular interpretation of the prediction: when a patient is first added to a cancer registry, given what we know about that patient, what is their estimated 5-year survival probability? 
    \item Cohort characteristics: Summaries of the SEER (Breast), SEER (Colon), and SEER (Lung) cohort characteristics are in Tables \ref{tab:seer_breast_characteristics}, \ref{tab:seer_colon_characteristics}, and \ref{tab:seer_lung_characteristics}.
    \item Outcome definition: 5-year survival is defined by a confirmation that the patient is alive five years after the year of diagnosis.
    \item Features: We list the features used in the SEER breast, colon, and lung cancer datasets in Section \ref{app:sec_seer_features}.  For all datasets, we convert all categorical variables into dummy features, and apply standard scaling to numerical variables (subtract mean and divide by standard deviation).
    \item Missingness heat maps: are given in Figures 
    \ref{fig:heatmap_seer_breast_cate}, \ref{fig:heatmap_seer_breast_num},
    \ref{fig:heatmap_seer_colon_cate}, \ref{fig:heatmap_seer_colon_num},
    \ref{fig:heatmap_seer_lung_cate}, and \ref{fig:heatmap_seer_lung_num}.
\end{itemize}

\clearpage
\subsection{Cohort Selection and Cohort Characteristics}

\clearpage

\begin{table*}[ht]
\caption{SEER (Breast) cohort characteristics, with count (\%) or median (Q1 -- Q3).}
\vspace{0.5em}
\label{tab:seer_breast_characteristics}
\centering
    \resizebox{1.0\columnwidth}{!}{%
    \begin{tabular}{lccc}
    \toprule
    Characteristic &        & Missingness &       Type \\
    \midrule
    \textbf{Sex} &                         &               &              \\
    \hspace{2em}Female &         459,184 (99.4\%) &          -- &  categorical \\
    \hspace{2em}Male &            2,839 (0.6\%) &          -- &  categorical \\
    \textbf{Age recode with single ages and 85+} &              60 (50-71) &          0.0\% &   continuous \\
    \textbf{Race/ethnicity} &                         &               &              \\
    \hspace{2em}White &         387,247 (83.8\%) &          -- &  categorical \\
    \hspace{2em}Black &           40,217 (8.7\%) &          -- &  categorical \\
    \hspace{2em}Other &           34,559 (7.5\%) &          -- &  categorical \\
    \textbf{Laterality} &                         &               &              \\
    \hspace{2em}Right - origin of primary &         224,777 (48.7\%) &          -- &  categorical \\
    \hspace{2em}Left - origin of primary &         233,549 (50.5\%) &          -- &  categorical \\
    \hspace{2em}Other &            3,697 (0.8\%) &          -- &  categorical \\
    \textbf{Regional nodes positive (1988+)} &                 0 (0-3) &         21.0\% &   continuous \\
    \textbf{T value - based on AJCC 3rd (1988-2003)} &              10 (10-20) &         56.2\% &  categorical \\
    \textbf{Derived AJCC T, 7th ed (2010-2015)} &              13 (13-20) &         85.3\% &  categorical \\
    \textbf{CS site-specific factor 3 (2004-2017 varying by schema)} &                 0 (0-2) &         64.8\% &  categorical \\
    \textbf{Regional nodes examined (1988+)} &                8 (2-15) &         21.0\% &   continuous \\
    \textbf{Coding system-EOD (1973-2003)} &                         &               &              \\
    \hspace{2em}Four-digit EOD (1983-1987) &           44,066 (9.5\%) &          -- &  categorical \\
    \hspace{2em}Ten-digit EOD (1988-2003) &         202,450 (43.8\%) &          -- &  categorical \\
    \hspace{2em}Thirteen-digit (expanded) site specific EOD (1973-1982) &          52,742 (11.4\%) &          -- &  categorical \\
    \hspace{2em}Blank(s) &         162,765 (35.2\%) &          -- &  categorical \\
    \textbf{CS version input original (2004-2015)} &  10,401 (10,300-20,302) &         64.8\% &  categorical \\
    \textbf{CS version input current (2004-2015)} &  20,520 (20,510-20,540) &         64.8\% &  categorical \\
    \textbf{EOD 10 - extent (1988-2003)} &              10 (10-13) &         56.2\% &  categorical \\
    \textbf{Grade (thru 2017)} &                         &               &              \\
    \hspace{2em}Unknown &         130,713 (28.3\%) &          -- &  categorical \\
    \hspace{2em}Moderately differentiated; Grade II &         135,970 (29.4\%) &          -- &  categorical \\
    \hspace{2em}Poorly differentiated; Grade III &         119,900 (26.0\%) &          -- &  categorical \\
    \hspace{2em}Undifferentiated; anaplastic; Grade IV &            8,081 (1.7\%) &          -- &  categorical \\
    \hspace{2em}Well differentiated; Grade I &          67,359 (14.6\%) &          -- &  categorical \\
    \textbf{SEER historic stage A} (1973-2015) &                         &               &              \\
    \hspace{2em}Regional &         136,207 (29.5\%) &          -- &  categorical \\
    \hspace{2em}Localized &         286,927 (62.1\%) &          -- &  categorical \\
    \hspace{2em}Unstaged &            9,242 (2.0\%) &          -- &  categorical \\
    \hspace{2em}Distant &           29,647 (6.4\%) &          -- &  categorical \\
    \textbf{IHS Link} &                         &               &              \\
    \hspace{2em}Record sent for linkage, no IHS match &         409,058 (88.5\%) &          -- &  categorical \\
    \hspace{2em}Record sent for linkage, IHS match &            1,505 (0.3\%) &          -- &  categorical \\
    \hspace{2em}Blank(s) &          51,460 (11.1\%) &          -- &  categorical \\
    \textbf{Histologic Type ICD-O-3} &     8,500 (8,500-8,500) &          0.0\% &  categorical \\
    \textbf{EOD 10 - size (1988-2003)} &              18 (10-30) &         56.2\% &  categorical \\
    \textbf{Type of Reporting Source} &                         &               &              \\
    \hspace{2em}Hospital inpatient/outpatient or clinic &         450,801 (97.6\%) &          -- &  categorical \\
    \hspace{2em}Other &           11,222 (2.4\%) &          -- &  categorical \\
    \textbf{SEER cause-specific death classification} &                         &               &              \\
    \hspace{2em}Alive or dead of other cause &         378,758 (82.0\%) &          -- &  categorical \\
    \hspace{2em}Dead (attributable to this cancer dx) &          83,265 (18.0\%) &          -- &  categorical \\
    \textbf{Survival months} &            135 (74-220) &          0.0\% &  categorical \\
    \textbf{5-year survival} &                         &               &              \\
    \hspace{2em}1 &         378,758 (82.0\%) &          -- &  categorical \\
    \hspace{2em}0 &          83,265 (18.0\%) &          -- &  categorical \\
    \bottomrule
    \end{tabular}%
    }
\end{table*}

\begin{table*}[ht]
\caption{SEER (Colon) cohort characteristics, with count (\%) or median (Q1--Q3).}
\vspace{0.5em}
\label{tab:seer_colon_characteristics}
    \centering
    \resizebox{1.0\columnwidth}{!}{%
    \begin{tabular}{lccc}
    \toprule
    Characteristic &  & Missingness &       Type \\
    \midrule
    \textbf{Sex} &                         &               &              \\
    \hspace{2em}Female &         133,661 (52.6\%) &          -- &  categorical \\
    \hspace{2em}Male &         120,451 (47.4\%) &          -- &  categorical \\
    \textbf{Age recode with single ages and 85+} &              70 (61-79) &          0.0\% &   continuous \\
    \textbf{Race recode (White, Black, Other)} &                         &               &              \\
    \hspace{2em}White &         212,265 (83.5\%) &          -- &  categorical \\
    \hspace{2em}Black &           24,041 (9.5\%) &          -- &  categorical \\
    \hspace{2em}Other &           17,806 (7.0\%) &          -- &  categorical \\
    \textbf{CS version input current (2004-2015)} &  20,510 (20,510-20,540) &         72.8\% &  categorical \\
    \textbf{Derived AJCC T, 6th ed (2004-2015)} &              30 (20-40) &         73.3\% &  categorical \\
    \textbf{Histology ICD-O-2} &     8,140 (8,140-8,210) &          0.0\% &  categorical \\
    \textbf{IHS Link} &                         &               &              \\
    \hspace{2em}Record sent for linkage, no IHS match &         208,802 (82.2\%) &          -- &  categorical \\
    \hspace{2em}Record sent for linkage, IHS match &              744 (0.3\%) &          -- &  categorical \\
    \hspace{2em}Blank(s) &          44,566 (17.5\%) &          -- &  categorical \\
    \textbf{Histology recode - broad groupings} &                         &               &              \\
    \hspace{2em}8140-8389: adenomas and adenocarcinomas &         213,193 (83.9\%) &          -- &  categorical \\
    \hspace{2em}8440-8499: cystic, mucinous and serous neoplasms &          28,257 (11.1\%) &          -- &  categorical \\
    \hspace{2em}8010-8049: epithelial neoplasms, NOS &            8,797 (3.5\%) &          -- &  categorical \\
    \hspace{2em}Other &            3,865 (1.5\%) &          -- &  categorical \\
    \textbf{Regional nodes positive (1988+)} &                1 (0-10) &         29.8\% &   continuous \\
    \textbf{CS mets at dx (2004-2015)} &                0 (0-22) &         72.8\% &   continuous \\
    \textbf{Reason no cancer-directed surgery} &                         &               &              \\
    \hspace{2em}Surgery performed &         223,929 (88.1\%) &          -- &  categorical \\
    \hspace{2em}Not recommended &           13,003 (5.1\%) &          -- &  categorical \\
    \hspace{2em}Other &           17,180 (6.8\%) &          -- &  categorical \\
    \textbf{Derived AJCC T, 6th ed (2004-2015)} &              30 (20-40) &         73.3\% &  categorical \\
    \textbf{CS version input original (2004-2015)} &  10,401 (10,300-20,302) &         72.8\% &  categorical \\
    \textbf{Primary Site} &           184 (182-187) &          0.0\% &  categorical \\
    \textbf{Diagnostic Confirmation} &                         &               &              \\
    \hspace{2em}Positive histology &         244,616 (96.3\%) &          -- &  categorical \\
    \hspace{2em}Radiography without microscopic confirm &            4,822 (1.9\%) &          -- &  categorical \\
    \hspace{2em}Other &            4,674 (1.8\%) &          -- &  categorical \\
    \textbf{EOD 10 - extent (1988-2003)} &              45 (40-85) &         57.0\% &  categorical \\
    \textbf{Histologic Type ICD-O-3} &     8,140 (8,140-8,210) &          0.0\% &  categorical \\
    \textbf{EOD 10 - size (1988-2003)} &             55 (35-999) &         57.0\% &  categorical \\
    \textbf{CS lymph nodes (2004-2015)} &               0 (0-210) &         72.8\% &  categorical \\
    \textbf{SEER cause-specific death classification} &                         &               &              \\
    \hspace{2em}Dead (attributable to this cancer dx) &         119,047 (46.8\%) &          -- &  categorical \\
    \hspace{2em}Alive or dead of other cause &         135,065 (53.2\%) &          -- &  categorical \\
    \textbf{Survival months} &             68 (12-151) &          0.0\% &  categorical \\
    \textbf{5-year survival} &                         &               &              \\
    \hspace{2em}1 &         135,065 (53.2\%) &          -- &  categorical \\
    \hspace{2em}0 &         119,047 (46.8\%) &          -- &  categorical \\
    \bottomrule
    \end{tabular}%
    }
    \vspace{2em}
\end{table*}

\begin{table*}[ht]
\centering
\caption{SEER (Lung) cohort characteristics, with count (\%) or median (Q1 -- Q3).}
\vspace{0.5em}
\label{tab:seer_lung_characteristics}
    \resizebox{1.0\columnwidth}{!}{%
    \begin{tabular}{lccc}
    \toprule
    Characteristic &  & Missingness &       Type \\
    \midrule
    \textbf{Sex} &                         &               &              \\
    \hspace{2em}Female &         187,967 (41.1\%) &          -- &  categorical \\
    \hspace{2em}Male &         269,728 (58.9\%) &          -- &  categorical \\
    \textbf{Age recode with single ages and 85+} &              68 (60-76) &          0.0\% &   continuous \\
    \textbf{Race recode (White, Black, Other)} &                         &               &              \\
    \hspace{2em}White &         384,184 (83.9\%) &          -- &  categorical \\
    \hspace{2em}Black &          47,237 (10.3\%) &          -- &  categorical \\
    \hspace{2em}Other &           26,274 (5.7\%) &          -- &  categorical \\
    \textbf{Histologic Type ICD-O-3} &     8,070 (8,041-8,140) &          0.0\% &  categorical \\
    \textbf{Laterality} &                         &               &              \\
    \hspace{2em}Left - origin of primary &         178,661 (39.0\%) &          -- &  categorical \\
    \hspace{2em}Right - origin of primary &         245,321 (53.6\%) &          -- &  categorical \\
    \hspace{2em}Paired site, but no information concerning laterality &           23,196 (5.1\%) &          -- &  categorical \\
    \hspace{2em}Other &           10,517 (2.3\%) &          -- &  categorical \\
    \textbf{EOD 10 - nodes (1988-2003)} &                 2 (1-9) &         56.3\% &  categorical \\
    \textbf{EOD 4 - nodes (1983-1987)} &                 3 (0-9) &         88.4\% &  categorical \\
    \textbf{Type of Reporting Source} &                         &               &              \\
    \hspace{2em}Hospital inpatient/outpatient or clinic &         445,606 (97.4\%) &          -- &  categorical \\
    \hspace{2em}Other &           12,089 (2.6\%) &          -- &  categorical \\
    \textbf{SEER historic stage A (1973-2015)} &                         &               &              \\
    \hspace{2em}Regional &          79,409 (17.3\%) &          -- &  categorical \\
    \hspace{2em}Distant &         182,467 (39.9\%) &          -- &  categorical \\
    \hspace{2em}Blank(s) &         123,161 (26.9\%) &          -- &  categorical \\
    \hspace{2em}Localized &          50,375 (11.0\%) &          -- &  categorical \\
    \hspace{2em}Unstaged &           22,283 (4.9\%) &          -- &  categorical \\
    \textbf{CS version input current (2004-2015)} &  20,520 (20,510-20,540) &         70.6\% &  categorical \\
    \textbf{CS mets at dx (2004-2015)} &               23 (0-40) &         70.6\% &   continuous \\
    \textbf{CS version input original (2004-2015)} &  10,401 (10,300-20,302) &         70.6\% &  categorical \\
    \textbf{CS tumor size (2004-2015)} &             50 (29-999) &         70.6\% &  categorical \\
    \textbf{EOD 10 - size (1988-2003)} &             80 (35-999) &         56.3\% &  categorical \\
    \textbf{CS lymph nodes (2004-2015)} &             200 (0-200) &         70.6\% &  categorical \\
    \textbf{Histology recode - broad groupings}&                         &               &              \\
    \hspace{2em}8140-8389: adenomas and adenocarcinomas &         147,127 (32.1\%) &          -- &  categorical \\
    \hspace{2em}8010-8049: epithelial neoplasms, NOS &         179,848 (39.3\%) &          -- &  categorical \\
    \hspace{2em}8440-8499: cystic, mucinous and serous neoplasms &            6,266 (1.4\%) &          -- &  categorical \\
    \hspace{2em}Other &         124,454 (27.2\%) &          -- &  categorical \\
    \textbf{EOD 10 - extent (1988-2003)} &              78 (40-85) &         56.3\% &  categorical \\
    \textbf{SEER cause-specific death classification} &                         &               &              \\
    \hspace{2em}Alive or dead of other cause &          49,997 (10.9\%) &          -- &  categorical \\
    \hspace{2em}Dead (attributable to this cancer dx) &         407,698 (89.1\%) &          -- &  categorical \\
    \textbf{Survival months} &                7 (2-19) &          0.0\% &  categorical \\
    \textbf{5-year survival} &                         &               &              \\
    \hspace{2em}1 &          49,997 (10.9\%) &          -- &  categorical \\
    \hspace{2em}0 &         407,698 (89.1\%) &          -- &  categorical \\
    \bottomrule
    \end{tabular}%
    }  
    \vspace{5em}
\end{table*}

\clearpage

\twocolumn
\subsection{Features}\label{app:sec_seer_features}
\textbf{SEER (Breast):}
{\tiny
\begin{verbatim}
AJCC stage 3rd edition (1988-2003)
AYA site recode/WHO 2008
Age recode with single ages and 85+
Behavior code ICD-O-2
Behavior code ICD-O-3
Behavior recode for analysis
Breast - Adjusted AJCC 6th M (1988-2015)
Breast - Adjusted AJCC 6th N (1988-2015)
Breast - Adjusted AJCC 6th Stage (1988-2015)
Breast - Adjusted AJCC 6th T (1988-2015)
Breast Subtype (2010+)
CS Schema - AJCC 6th Edition
CS extension (2004-2015)
CS lymph nodes (2004-2015)
CS mets at dx (2004-2015)
CS site-specific factor 1 (2004-2017 varying by schema)
CS site-specific factor 15 (2004-2017 varying by schema)
CS site-specific factor 2 (2004-2017 varying by schema)
CS site-specific factor 25 (2004-2017 varying by schema)
CS site-specific factor 3 (2004-2017 varying by schema)
CS site-specific factor 4 (2004-2017 varying by schema)
CS site-specific factor 5 (2004-2017 varying by schema)
CS site-specific factor 6 (2004-2017 varying by schema)
CS site-specific factor 7 (2004-2017 varying by schema)
CS tumor size (2004-2015)
CS version derived (2004-2015)
CS version input current (2004-2015)
CS version input original (2004-2015)
Coding system-EOD (1973-2003)
Derived AJCC M, 6th ed (2004-2015)
Derived AJCC M, 7th ed (2010-2015)
Derived AJCC N, 6th ed (2004-2015)
Derived AJCC N, 7th ed (2010-2015)
Derived AJCC Stage Group, 6th ed (2004-2015)
Derived AJCC Stage Group, 7th ed (2010-2015)
Derived AJCC T, 6th ed (2004-2015)
Derived AJCC T, 7th ed (2010-2015)
Derived HER2 Recode (2010+)
EOD 10 - extent (1988-2003)
EOD 10 - nodes (1988-2003)
EOD 10 - size (1988-2003)
ER Status Recode Breast Cancer (1990+)
First malignant primary indicator
Grade (thru 2017)
Histologic Type ICD-O-3
Histology recode - Brain groupings
Histology recode - broad groupings
ICCC site rec extended ICD-O-3/WHO 2008
IHS Link
Laterality
Lymphoma subtype recode/WHO 2008 (thru 2017)
M value - based on AJCC 3rd (1988-2003)
N value - based on AJCC 3rd (1988-2003)
Origin recode NHIA (Hispanic, Non-Hisp)
PR Status Recode Breast Cancer (1990+)
Primary Site
Primary by international rules
Race recode (W, B, AI, API)
Race recode (White, Black, Other)
Race/ethnicity
Regional nodes examined (1988+)
Regional nodes positive (1988+)
SEER historic stage A (1973-2015)
SEER modified AJCC stage 3rd (1988-2003)
Sex
Site recode ICD-O-3/WHO 2008
T value - based on AJCC 3rd (1988-2003)
Tumor marker 1 (1990-2003)
Tumor marker 2 (1990-2003)
Tumor marker 3 (1998-2003)
Type of Reporting Source
\end{verbatim}}
\textbf{SEER (Colon):}
{\tiny
\begin{verbatim}
Age recode with <1 year olds
Age recode with single ages and 85+
Behavior code ICD-O-2
Behavior code ICD-O-3
CS extension (2004-2015)
CS lymph nodes (2004-2015)
CS mets at dx (2004-2015)
CS site-specific factor 1 (2004-2017 varying by schema)
CS tumor size (2004-2015)
CS version input current (2004-2015)
CS version input original (2004-2015)
Derived AJCC M, 6th ed (2004-2015)
Derived AJCC M, 7th ed (2010-2015)
Derived AJCC N, 6th ed (2004-2015)
Derived AJCC N, 7th ed (2010-2015)
Derived AJCC Stage Group, 6th ed (2004-2015)
Derived AJCC Stage Group, 7th ed (2010-2015)
Derived AJCC T, 6th ed (2004-2015)
Derived AJCC T, 7th ed (2010-2015)
Diagnostic Confirmation
EOD 10 - extent (1988-2003)
EOD 10 - nodes (1988-2003)
EOD 10 - size (1988-2003)
Histologic Type ICD-O-3
Histology ICD-O-2
Histology recode - broad groupings
IHS Link
Origin recode NHIA (Hispanic, Non-Hisp)
Primary Site
Primary by international rules
RX Summ--Surg Prim Site (1998+)
Race recode (White, Black, Other)
Reason no cancer-directed surgery
Regional nodes positive (1988+)
SEER modified AJCC stage 3rd (1988-2003)
Sex
\end{verbatim}}
\textbf{SEER (Lung):}
{\tiny
\begin{verbatim}
AYA site recode/WHO 2008
Age recode with <1 year olds
Age recode with single ages and 85+
Behavior code ICD-O-2
Behavior code ICD-O-3
CS extension (2004-2015)
CS lymph nodes (2004-2015)
CS mets at dx (2004-2015)
CS site-specific factor 1 (2004-2017 varying by schema)
CS tumor size (2004-2015)
CS version input current (2004-2015)
CS version input original (2004-2015)
Derived AJCC M, 6th ed (2004-2015)
Derived AJCC M, 7th ed (2010-2015)
Derived AJCC N, 6th ed (2004-2015)
Derived AJCC N, 7th ed (2010-2015)
Derived AJCC Stage Group, 6th ed (2004-2015)
Derived AJCC T, 6th ed (2004-2015)
Derived AJCC T, 7th ed (2010-2015)
EOD 10 - extent (1988-2003)
EOD 10 - nodes (1988-2003)
EOD 10 - size (1988-2003)
EOD 4 - nodes (1983-1987)
First malignant primary indicator
Grade (thru 2017)
Histologic Type ICD-O-3
Histology recode - broad groupings
ICCC site recode 3rd edition/IARC 2017
ICCC site recode extended 3rd edition/IARC 2017
IHS Link
Laterality
Origin recode NHIA (Hispanic, Non-Hisp)
Primary by international rules
Race recode (White, Black, Other)
SEER historic stage A (1973-2015)
Sex
Type of Reporting Source
\end{verbatim}}
\onecolumn

\subsection{Missingness heatmaps}

This section plots missingness heatmaps of categorical and numerical features in each SEER dataset over time. Darker color means larger proportion of missing data.

\vspace{-3em}
\vspace{-3em}
\vspace{-3em}

\clearpage

\section{Additional CDC COVID-19 Data Details}\label{app:cdc_data}
The COVID-19 Case Surveillance Detailed Data \citep{cdc_data} is a national, publicly available dataset provided by the CDC. 
It contains 33 elements, with patient-level data including symptoms, demographics, and state of residence. The performance over time is evaluated on a \emph{monthly} basis. We use the version the released on June 6th, 2022.

\begin{figure}[!h]
  \includegraphics[width=\columnwidth]{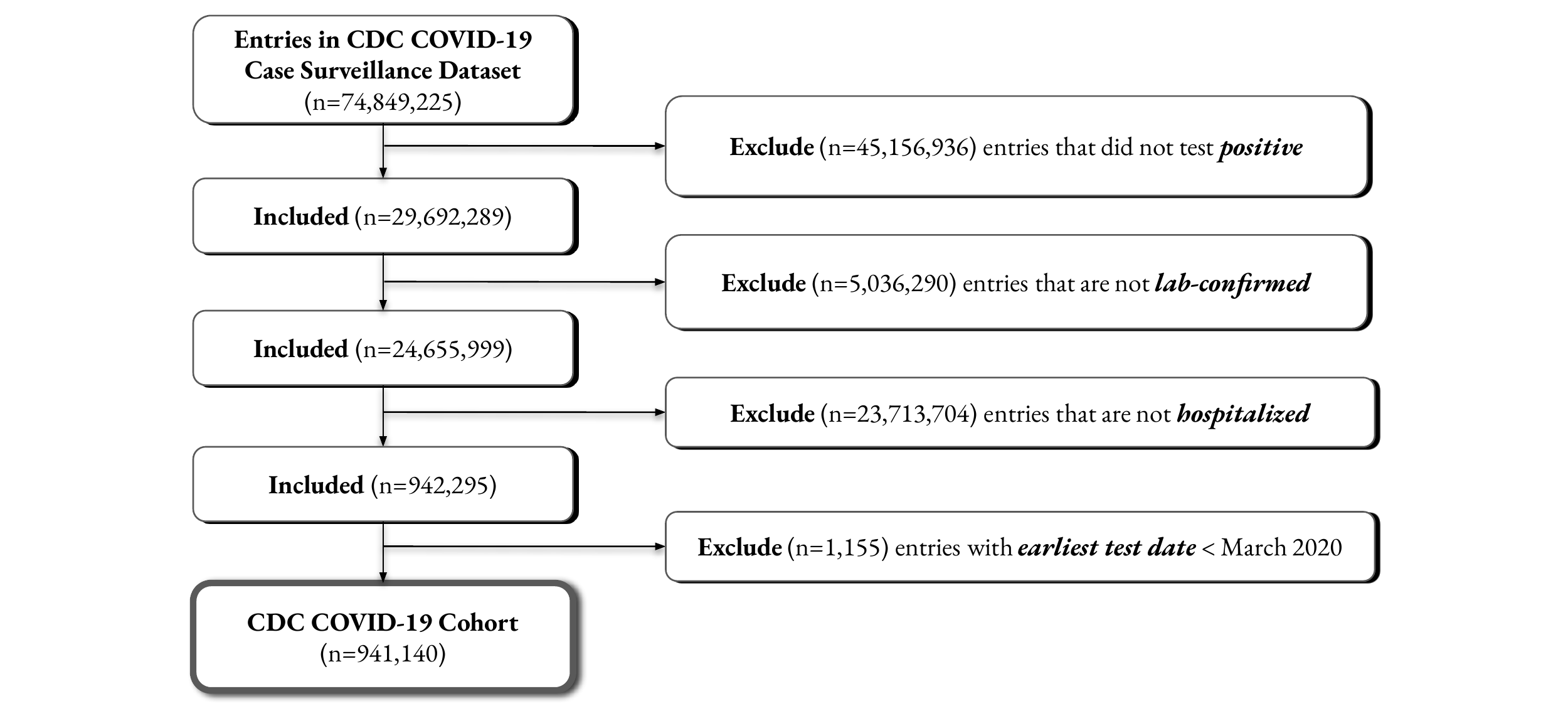}
  \caption{Cohort selection diagram - CDC COVID-19}
  \label{fig:cdc_covid_cohort}
\end{figure}

\begin{figure}[!h]
  \includegraphics[width=1.0\columnwidth]{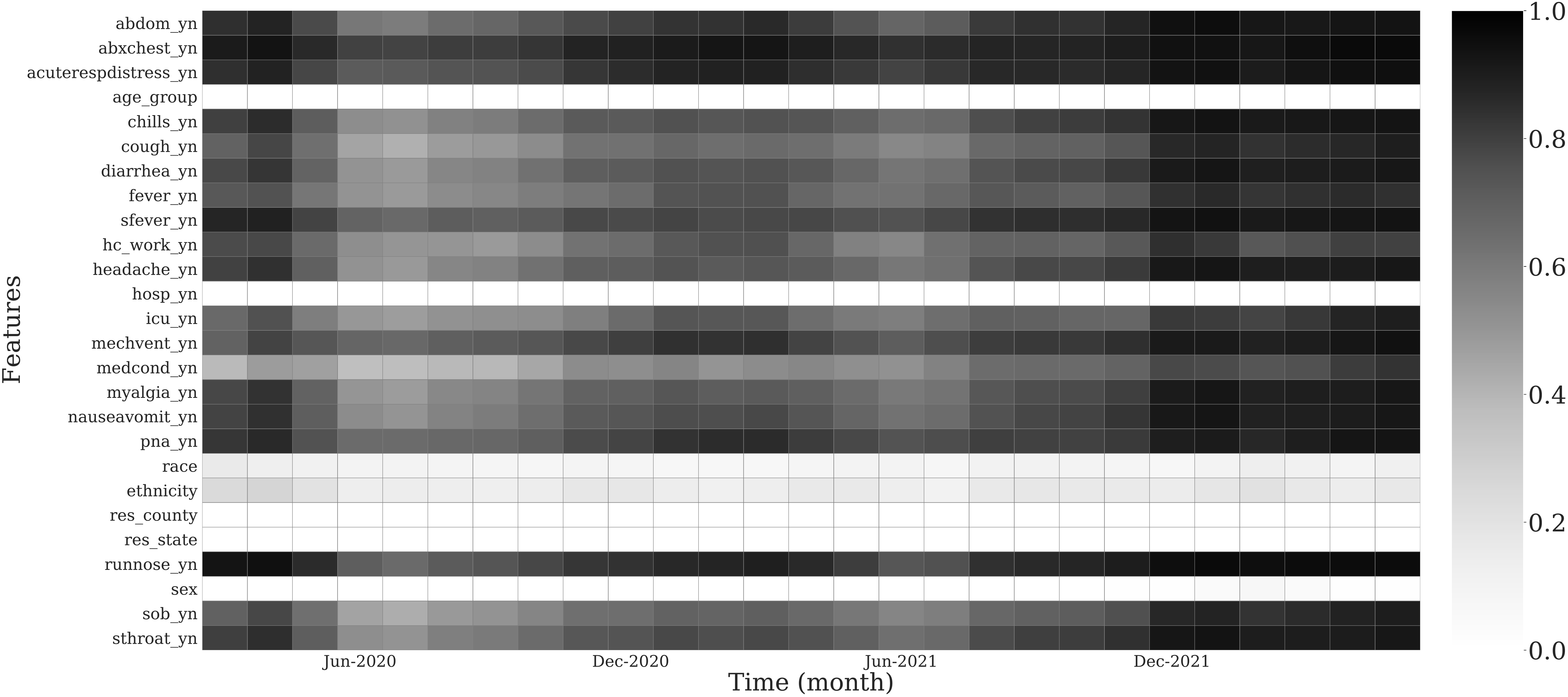}
  \caption{Missingness over time for features in CDC COVID-19 dataset after cohort selection. The darker the color, the larger the proportion of missing data.}
  \label{fig:heatmap_cdc_covid}
\end{figure}

\begin{figure}[H]
\centering
  {%
    \subfigure[By age group]{\label{fig:stack_age_group}%
      \includegraphics[width=0.5\columnwidth]{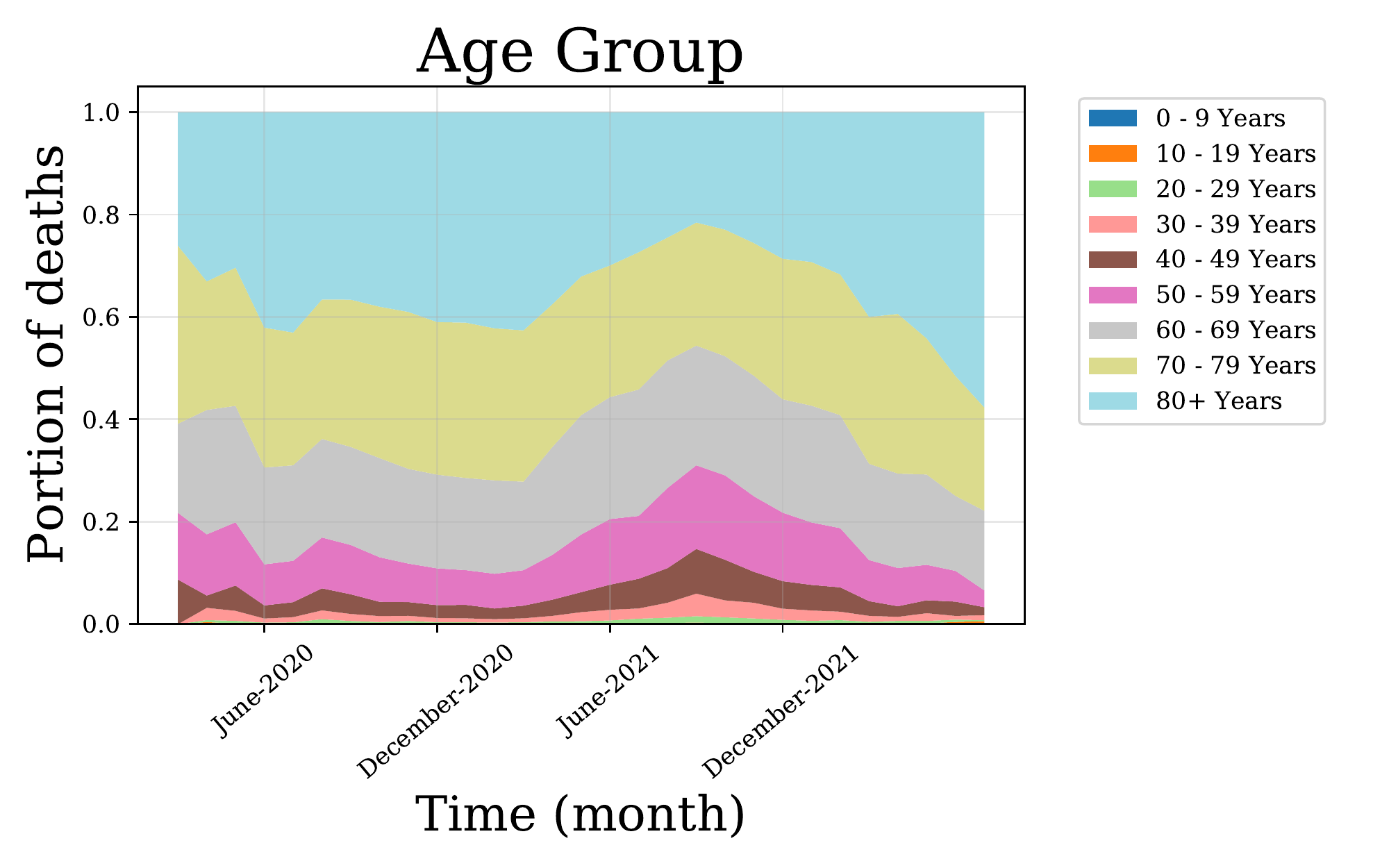}}%
    \subfigure[By state of residence]{\label{fig:stack_state_residence}%
      \includegraphics[width=0.5\columnwidth]{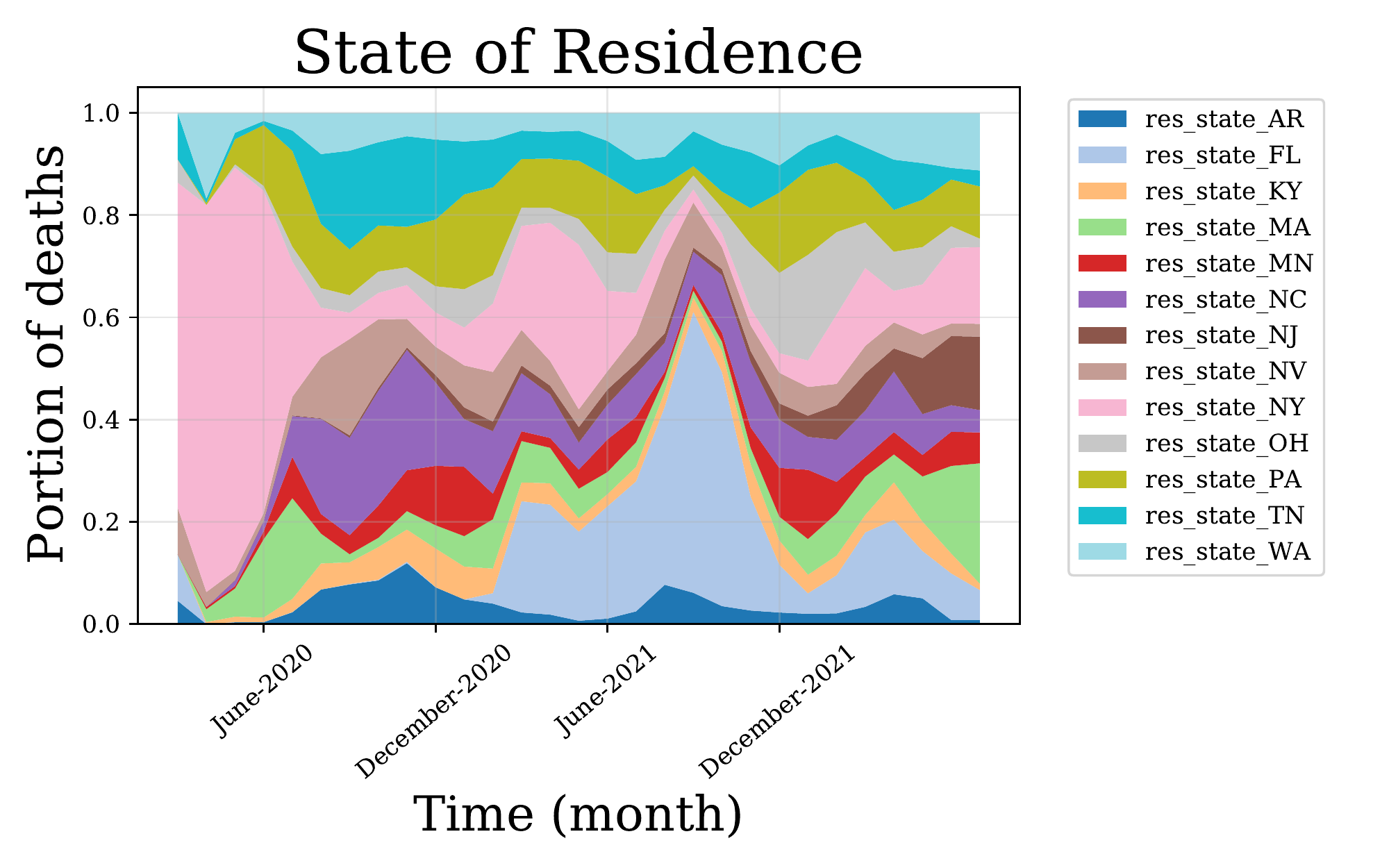}}%
  }
  \caption{Proportion of deaths over time for each age group and state of residence.}
  \label{fig:cdc_stack_area_plot}
\end{figure}

\begin{itemize}
    \item First, a disclaimer: ``The CDC does not take responsibility for the scientific validity or accuracy of methodology, results, statistical analyses, or conclusions presented.''
    \item Data access: To access the data, users must complete a registration information and data use restrictions agreement (RIDURA).
    \item Cohort selection: The cohort consists of all patients who were lab-confirmed positive for COVID-19, had a non-null positive specimen date, and were hospitalized (\verb|hosp_yn = Yes|). Cohort selection diagrams are given in Figures \ref{fig:cdc_covid_cohort}
    \item Cohort characteristics: Cohort characteristics are given in Table \ref{tab:cdc_covid_characteristics}.
    \item Outcome definition: mortality, defined by \verb|death_yn = Yes|
    \item Features: We list the features used in the CDC COVID-19 datasets in Section \ref{app:sec_cdc_features}.
    We convert all categorical variables into dummy features, and apply standard scaling to numerical variables (subtract mean and divide by standard deviation).
    \item Missingness heat map: is given in Figure \ref{fig:heatmap_cdc_covid}. 
    \item Additionally, we provide stacked area plots showing how the distribution of ages (Figure \ref{fig:stack_age_group} and states \ref{fig:stack_state_residence} shifts over time.
\end{itemize}
\clearpage

\subsection{Cohort Selection and Cohort Characteristics}

\clearpage

\begin{table*}[htb]
\centering
\caption{CDC COVID-19 cohort characteristics, with count (\%) or median (Q1--Q3).}
\vspace{0.5em}
\label{tab:cdc_covid_characteristics}
    \resizebox{0.75\columnwidth}{!}{%
    \begin{tabular}{lccc}
    \toprule
    Characteristic &  & Missingness &       Type \\
    \midrule
    \textbf{Sex} &                   &               &              \\
    \hspace{2em}Female &   455,376 (48.4\%) &          -- &  categorical \\
    \hspace{2em}Male &   475,223 (50.5\%) &          -- &  categorical \\
    \hspace{2em}Unknown/Missing &      10,541 (1.1\%) &          -- &  categorical \\
    \textbf{Age Group} &                   &               &              \\
    \hspace{2em}0 - 9  &     16,373 (1.7\%) &          -- &  categorical \\
    \hspace{2em}10 - 19  &     17,252 (1.8\%) &          -- &  categorical \\
    \hspace{2em}20 - 29  &     48,505 (5.2\%) &          -- &  categorical \\
    \hspace{2em}30 - 39  &     71,776 (7.6\%) &          -- &  categorical \\
    \hspace{2em}40 - 49  &     88,531 (9.4\%) &          -- &  categorical \\
    \hspace{2em}50 - 59  &   141,805 (15.1\%) &          -- &  categorical \\
    \hspace{2em}60 - 69  &   189,354 (20.1\%) &          -- &  categorical \\
    \hspace{2em}70 - 79  &   189,018 (20.1\%) &          -- &  categorical \\
    \hspace{2em}80+  &   177,765 (18.9\%) &          -- &  categorical \\
    \hspace{2em}Missing &        761 (0.1\%) &          -- &  categorical \\
    \textbf{Race} &                   &               &              \\
    \hspace{2em}White &   544,199 (57.8\%) &          -- &  categorical \\
    \hspace{2em}Black &   173,847 (18.5\%) &          -- &  categorical \\
    \hspace{2em}Other &   205,547 (21.8\%) &          -- &  categorical \\
    \textbf{State of Residence} &                   &               &              \\
    \hspace{2em}NY &   189,684 (20.2\%) &          -- &  categorical \\
    \hspace{2em}OH &     70,097 (7.4\%) &          -- &  categorical \\
    \hspace{2em}FL &     35,679 (3.8\%) &          -- &  categorical \\
    \hspace{2em}WA &     58,854 (6.3\%) &          -- &  categorical \\
    \hspace{2em}MA &     31,441 (3.3\%) &          -- &  categorical \\
    \hspace{2em}Other &   555,353 (59.0\%) &          -- &  categorical \\
    \textbf{Mechanical Ventilation} &                   &               &              \\
    \hspace{2em}Yes &     38,009 (4.0\%) &          -- &  categorical \\
    \hspace{2em}No &   138,331 (14.7\%) &          -- &  categorical \\
    \hspace{2em}Unknown/Missing &   764,800 (81.2\%) &          -- &  categorical \\
    \textbf{Mortality} &                   &               &              \\
    \hspace{2em}1 &   190,786 (20.3\%) &          -- &  categorical \\
    \hspace{2em}0 &   750,354 (79.7\%) &          -- &  categorical \\
    \bottomrule
    \end{tabular}%
    }
\end{table*}
\clearpage

\subsection{Features}\label{app:sec_cdc_features}
{\small \begin{verbatim} abdom_yn, abxchest_yn, acuterespdistress_yn, age_group, chills_yn, cough_yn, diarrhea_yn, 
ethnicity, fever_yn, hc_work_yn, headache_yn, hosp_yn, icu_yn, mechvent_yn, medcond_yn, month, 
myalgia_yn, nauseavomit_yn, pna_yn, race, relative_month, res_county, res_state, runnose_yn, 
sex, sfever_yn, sob_yn, sthroat_yn, \end{verbatim}}

\subsection{Missingness heatmaps}

\subsection{Additional Figures}
\clearpage

\section{Additional SWPA COVID-19 Data Details}\label{app:swpa_data}
The Southwestern Pennsylvania (SWPA) COVID-19 dataset consists of EHR data from patients tested for COVID-19. It was collected by a major healthcare provider in SWPA \citep{ahn_clinical_concepts}, and includes
patient demographics, labs, problem histories, medications, inpatient vs. outpatient status, and other information collected in the patient encounter. The performance over time is evaluated on a \emph{monthly} basis.

\begin{figure}[!h]
  \includegraphics[width=1.0\columnwidth]{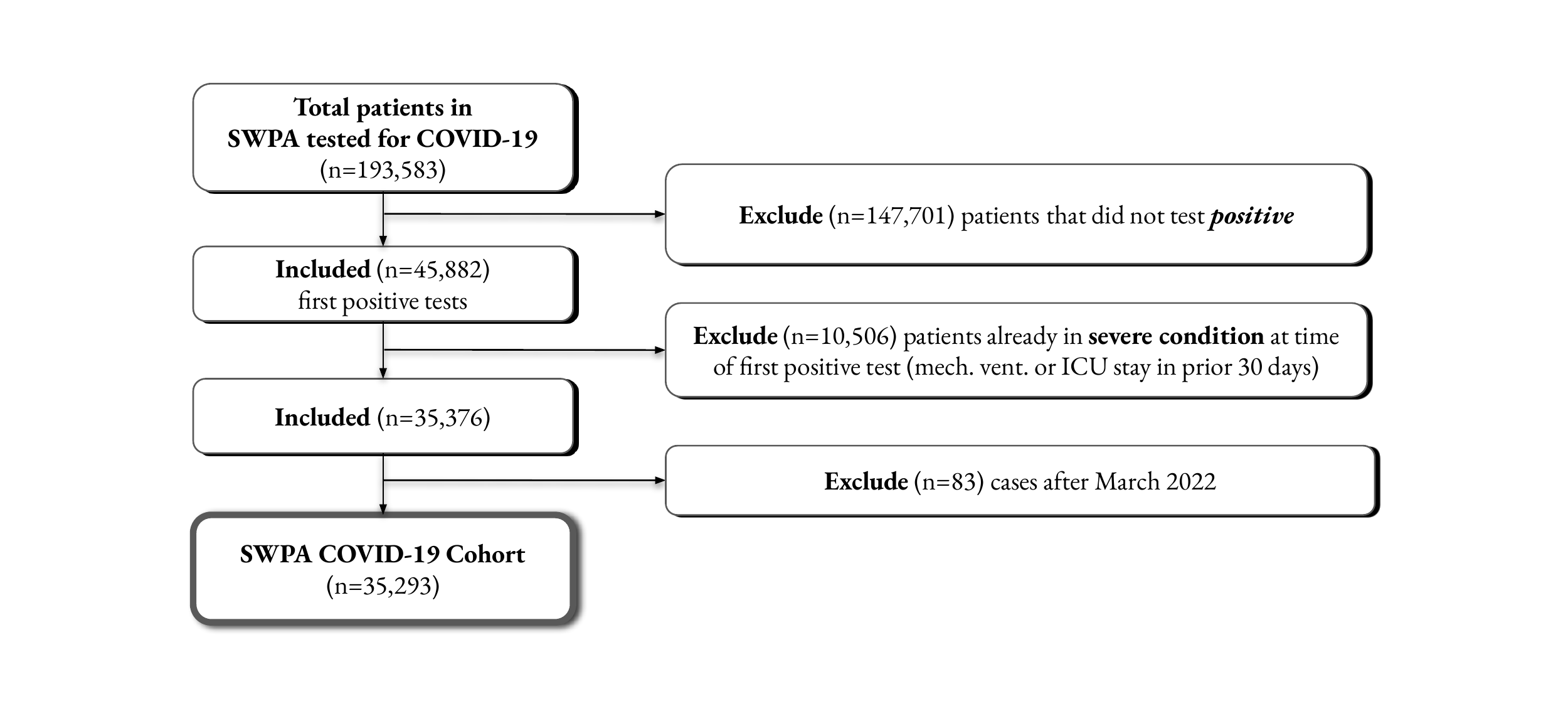}
  \caption{Cohort selection diagram - SWPA COVID-19}
  \label{fig:swpa_covid_cohort}
\end{figure}

\begin{figure}[H]
    \includegraphics[width=1.0\columnwidth]{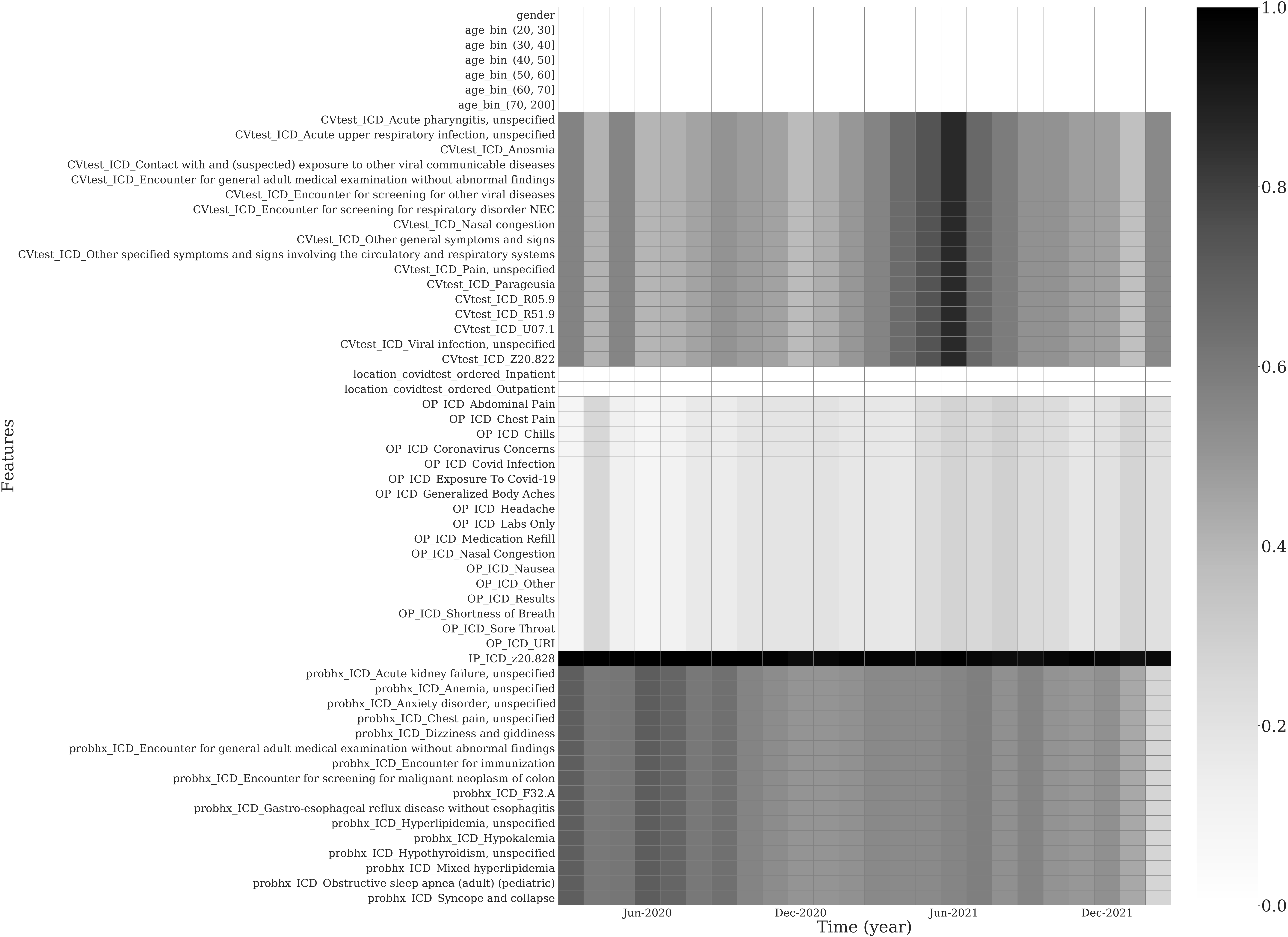}
  \caption{Missingness of categorical features in SWPA COVID-19 dataset (part 1).}
  \label{fig:heatmap_pa_covid_cate_1}
\end{figure}

\begin{figure}[H]
  \includegraphics[width=1.0\columnwidth]{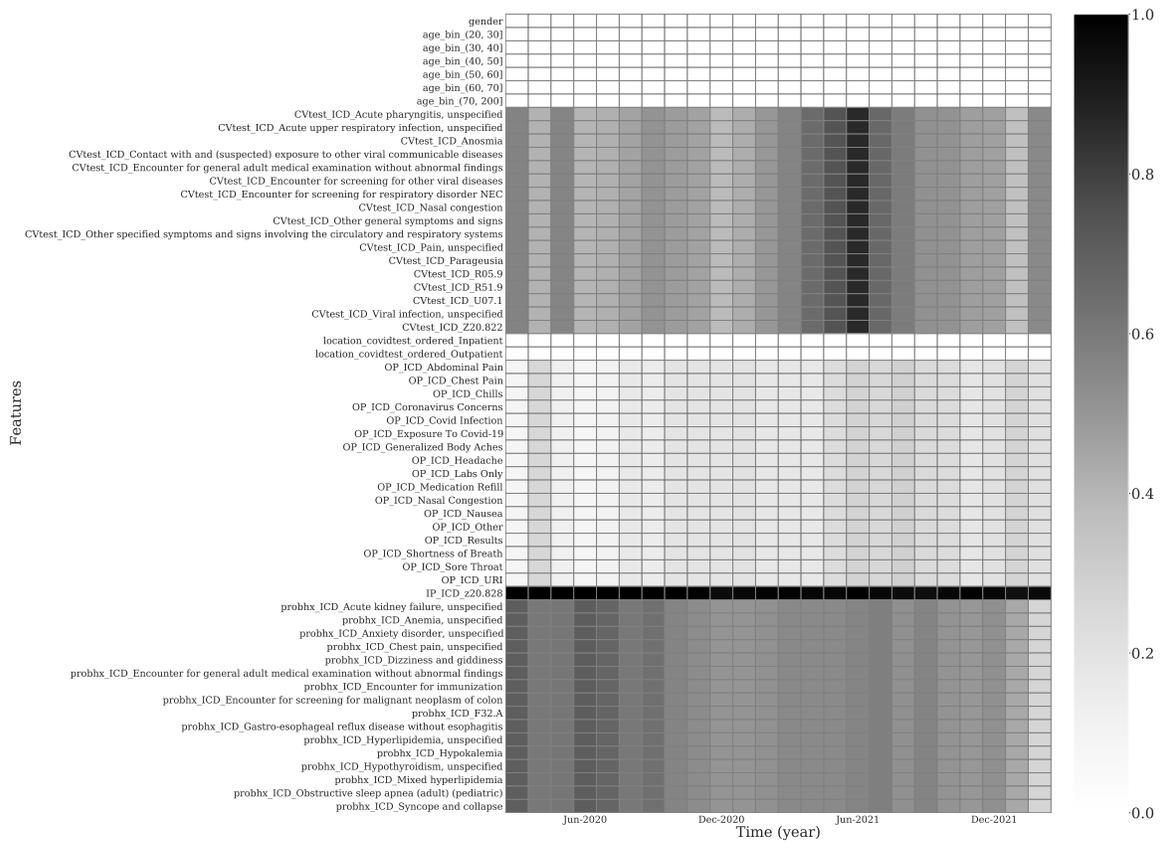}
  \caption{Missingness of categorical features in SWPA COVID-19 dataset (part 2).}
  \label{fig:heatmap_pa_covid_cate_2}
\end{figure}

\begin{figure}[H]
  \includegraphics[width=1.0\columnwidth]{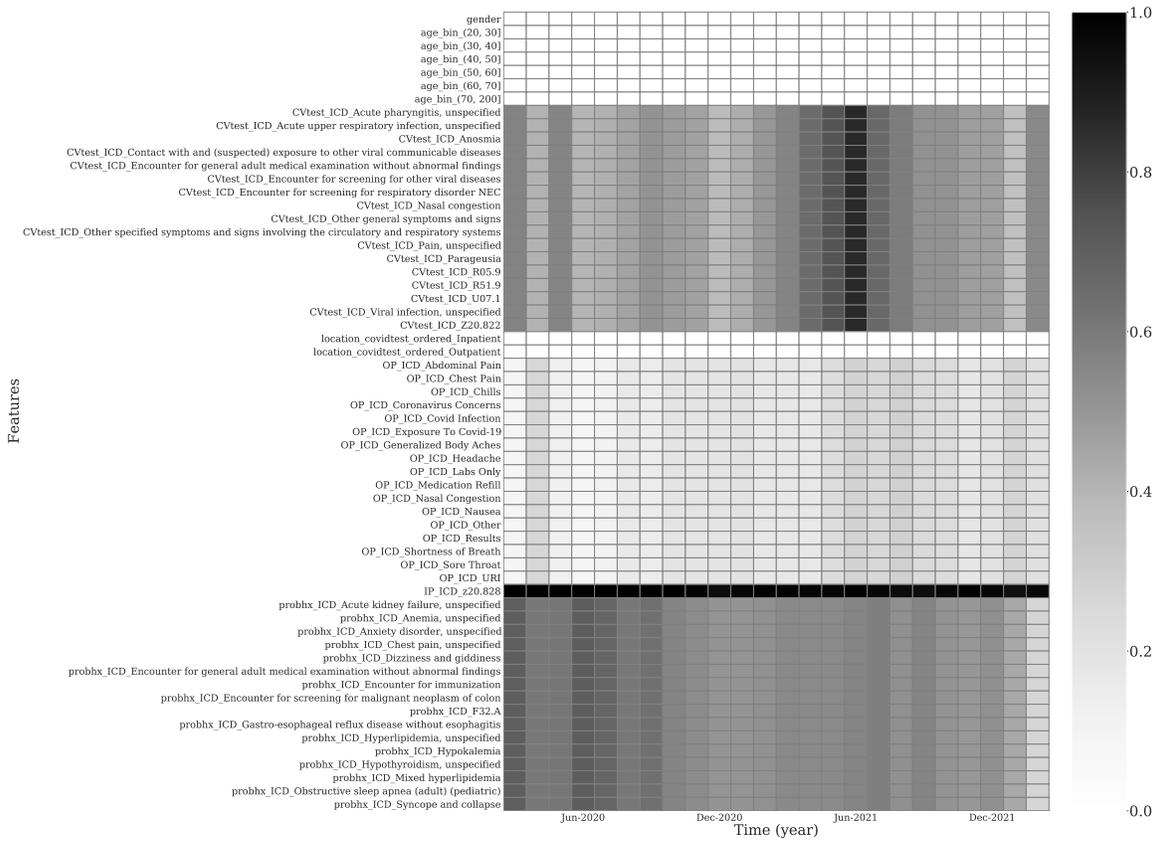}
  \caption{Missingness of categorical features in SWPA COVID-19 dataset (part 3).}
  \label{fig:heatmap_pa_covid_cate_3}
\end{figure}

\begin{figure}[H]
  \includegraphics[width=1.0\columnwidth]{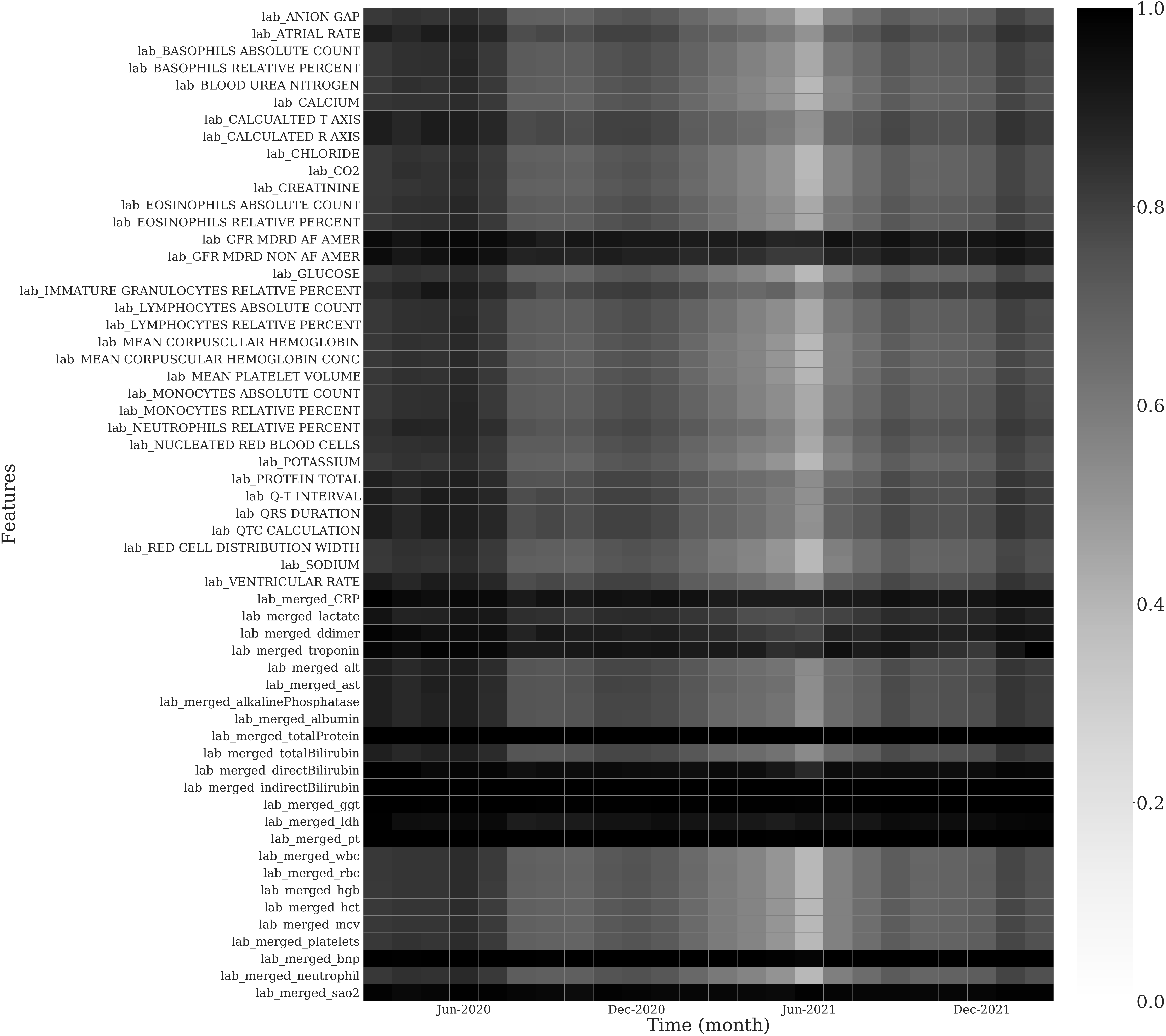}
  \caption{Missingness of numerical features in SWPA COVID-19.}
  \label{fig:heatmap_pa_covid_num}
\end{figure}

\begin{itemize}
    \item Data access: This is a private dataset.
    \item Cohort selection: The cohort consists of COVID-19 patients who tested positive for COVID-19 and were not already in the ICU or mechanically ventilated. We filter for the first positive test, and define features and outcomes relative to that time.  Cohort selection diagrams are given in Figures \ref{fig:swpa_covid_cohort}. If there are multiple samples per patient, we filter to the first entry per patient, which corresponds to when a patient first enters the dataset. This corresponds to a particular interpretation of the prediction: when a patient is first tests positive, given what we know about that patient, what is their estimated risk of 90-day mortality?
    \item Cohort characteristics: Cohort characteristics are given in Table \ref{tab:swpa_cohort_characteristics}.
    \item Outcome definition: 90-day mortality by comparing the death date and test date
    \item Features: We list the features used in the SWPA COVID-19 datasets in Section \ref{app:sec_swpa_features}. We convert all categorical variables into dummy features, and apply standard scaling to numerical variables (subtract mean and divide by standard deviation). To create a fixed length feature vector, where applicable we take the most recent value of each feature (e.g. most recent lab values).
    \item Missingness heat maps: are given in Figures \ref{fig:heatmap_pa_covid_cate_1}, \ref{fig:heatmap_pa_covid_cate_2}, \ref{fig:heatmap_pa_covid_cate_3}, and
    \ref{fig:heatmap_pa_covid_num},
\end{itemize}
\clearpage

\subsection{Cohort Selection and Cohort Characteristics}

\begin{table*}[ht]
\centering
\caption{SWPA COVID-19 cohort characteristics, with count (\%) or median (Q1--Q3).}
\vspace{0.5em}
\label{tab:swpa_cohort_characteristics}
\resizebox{0.6\columnwidth}{!}{%
    \begin{tabular}{lccc}
    \toprule
    Characteristic &  & Missingness &       Type \\
    \midrule
    \textbf{Gender} &                   &               &              \\
    \hspace{2em}Female &    20,283 (57.5\%) &          -- &  categorical \\
    \hspace{2em}Male &    15,003 (42.5\%) &          -- &  categorical \\
    \hspace{2em}Unknown &          7 (0.0\%) &          --&  categorical \\
    \textbf{Age} &                   &               &              \\
    \hspace{2em}Under 20 &                   3,210 (9.1\%)&               --&              categorical\\
    \hspace{2em}20 -- 30 &                   4,349 (12.3\%)&               --&              categorical\\
    \hspace{2em}30 -- 40 &                   4,667 (13.2\%)&               --&              categorical\\
    \hspace{2em}40 -- 50 &                   4,653 (13.2\%)&               --&              categorical\\
    \hspace{2em}50 -- 60 &                   6,111 (17.3\%)&               --&              categorical\\
    \hspace{2em}60 -- 70 &                   5,700 (16.2\%)&               --&              categorical\\
    \hspace{2em}70+ &                   6,603 (18.7\%)&               --&              categorical\\
    \textbf{Location of test} & & & \\
    \hspace{2em}Inpatient &                   14,911 (42.2\%)&               --&              categorical\\
    \hspace{2em}Outpatient &                   17,661 (50.0\%)&               --&              categorical\\
    \hspace{2em}Unknown &                   2,721 (7.7\%)&               --&              categorical\\
    \textbf{90-day mortality} &                   &               &              \\
    \hspace{2em}True &      1,516 (4.3\%) &          -- &  categorical \\
    \hspace{2em}False &    33,777 (95.7\%) &          -- &  categorical \\
    \bottomrule
    \end{tabular}%
    }
\end{table*}

\clearpage
\twocolumn
\subsection{Features}\label{app:sec_swpa_features}
{\tiny
\begin{verbatim}
Asthma
CAD
CHF
CKD
COPD
CRP
CVtest_ICD_Acute pharyngitis, unspecified
CVtest_ICD_Acute upper respiratory infection, unspecified
CVtest_ICD_Anosmia
CVtest_ICD_Contact with and (suspected) exposure to other viral 
    communicable diseases
CVtest_ICD_Encounter for general adult medical 
    examination without 
    abnormal findings
CVtest_ICD_Encounter for screening for other viral diseases
CVtest_ICD_Encounter for screening for respiratory disorder NEC
CVtest_ICD_Nasal congestion
CVtest_ICD_Other general symptoms and signs
CVtest_ICD_Other specified symptoms and signs involving the 
    circulatory and respiratory systems
CVtest_ICD_Pain, unspecified
CVtest_ICD_Parageusia
CVtest_ICD_R05.9
CVtest_ICD_R51.9
CVtest_ICD_U07.1
CVtest_ICD_Viral infection, unspecified
CVtest_ICD_Z20.822
ESLD
Hypertension
IP_ICD_z20.828
Immunocompromised
Interstitial Lung disease
OP_ICD_Abdominal Pain
OP_ICD_Chest Pain
OP_ICD_Chills
OP_ICD_Coronavirus Concerns
OP_ICD_Covid Infection
OP_ICD_Exposure To Covid-19
OP_ICD_Generalized Body Aches
OP_ICD_Headache
OP_ICD_Labs Only
OP_ICD_Medication Refill
OP_ICD_Nasal Congestion
OP_ICD_Nausea
OP_ICD_Other
OP_ICD_Results
OP_ICD_Shortness of Breath
OP_ICD_Sore Throat
OP_ICD_URI
age_bin_(20, 30]
age_bin_(30, 40]
age_bin_(40, 50]
age_bin_(50, 60]
age_bin_(60, 70]
age_bin_(70, 200]
bmi
cancer
cough
covid_vaccination_given
diabetes
fatigue
fever
gender
hyperglycemia
lab_ANION GAP
lab_ATRIAL RATE
lab_BASOPHILS ABSOLUTE COUNT
lab_BASOPHILS RELATIVE PERCENT
lab_BLOOD UREA NITROGEN
lab_CALCIUM
lab_CALCUALTED T AXIS
lab_CALCULATED R AXIS
lab_CHLORIDE
lab_CO2
lab_CREATININE
lab_EOSINOPHILS ABSOLUTE COUNT
lab_EOSINOPHILS RELATIVE PERCENT
lab_GFR MDRD AF AMER
lab_GFR MDRD NON AF AMER
lab_GLUCOSE
lab_IMMATURE GRANULOCYTES RELATIVE PERCENT
lab_LYMPHOCYTES ABSOLUTE COUNT
lab_LYMPHOCYTES RELATIVE PERCENT
lab_MEAN CORPUSCULAR HEMOGLOBIN
lab_MEAN CORPUSCULAR HEMOGLOBIN CONC
lab_MEAN PLATELET VOLUME
lab_MONOCYTES ABSOLUTE COUNT
lab_MONOCYTES RELATIVE PERCENT
lab_NEUTROPHILS RELATIVE PERCENT
lab_NUCLEATED RED BLOOD CELLS
lab_POTASSIUM
lab_PROTEIN TOTAL
lab_Q-T INTERVAL
lab_QRS DURATION
lab_QTC CALCULATION
lab_RED CELL DISTRIBUTION WIDTH
lab_SODIUM
lab_VENTRICULAR RATE
lab_merged_CRP
lab_merged_albumin
lab_merged_alkalinePhosphatase
lab_merged_alt
lab_merged_ast
lab_merged_bnp
lab_merged_ddimer
lab_merged_directBilirubin
lab_merged_ggt
lab_merged_hct
lab_merged_hgb
lab_merged_indirectBilirubin
lab_merged_lactate
lab_merged_ldh
lab_merged_mcv
lab_merged_neutrophil
lab_merged_platelets
lab_merged_pt
lab_merged_rbc
lab_merged_sao2
lab_merged_totalBilirubin
lab_merged_totalProtein
lab_merged_troponin
lab_merged_wbc
labs_ICD_Acute pharyngitis, unspecified
labs_ICD_Acute upper respiratory infection, unspecified
labs_ICD_Chest pain, unspecified
labs_ICD_Contact with and (suspected) exposure to other 
    viral communicable diseases
labs_ICD_Dyspnea, unspecified
labs_ICD_Encounter for other preprocedural examination
labs_ICD_Essential (primary) hypertension
labs_ICD_Fever, unspecified
labs_ICD_Heart failure, unspecified
labs_ICD_Other general symptoms and signs
labs_ICD_Other pulmonary embolism without acute cor pulmonale
labs_ICD_Other specified abnormalities of plasma proteins
labs_ICD_R05.9
labs_ICD_Shortness of breath
labs_ICD_Syncope and collapse
labs_ICD_U07.1
labs_ICD_Unspecified atrial fibrillation
labs_ICD_Viral infection, unspecified
labs_ICD_Z20.822
liver disease
location_covidtest_ordered_Inpatient
location_covidtest_ordered_Outpatient
lung disease
med_dx_Acquired hypothyroidism
med_dx_Anxiety
med_dx_COVID-19
med_dx_Encounter for antineoplastic chemotherapy
med_dx_Encounter for antineoplastic chemotherapy and immunotherapy
med_dx_Encounter for antineoplastic immunotherapy
med_dx_Encounter for immunization
med_dx_Gastroesophageal reflux disease without esophagitis
med_dx_Gastroesophageal reflux disease, esophagitis presence 
    not specified
med_dx_Generalized anxiety disorder
med_dx_Hyperlipidemia, unspecified hyperlipidemia type
med_dx_Hypomagnesemia
med_dx_Hypothyroidism, unspecified type
med_dx_Iron deficiency anemia, unspecified iron deficiency anemia type
med_dx_Mixed hyperlipidemia
med_dx_Primary osteoarthritis of right knee
medication_ACETAMINOPHEN 325 MG TABLET
medication_ALBUTEROL SULFATE 2.5 MG/3 ML (0.083 %
    FOR NEBULIZATION
medication_ALBUTEROL SULFATE HFA 90 MCG/ACTUATION AEROSOL INHALER
medication_ASPIRIN 81 MG TABLET,DELAYED RELEASE
medication_DEXAMETHASONE SODIUM PHOSPHATE 4 MG/ML INJECTION SOLUTION
medication_DIPHENHYDRAMINE 50 MG/ML INJECTION (WRAPPER)
medication_EPINEPHRINE 0.3 MG/0.3 ML INJECTION, AUTO-INJECTOR
medication_FENTANYL (PF) 50 MCG/ML INJECTION SOLUTION
medication_HYDROCODONE 5 MG-ACETAMINOPHEN 325 MG TABLET
medication_HYDROCORTISONE SOD SUCCINATE (PF) 100 MG/2 ML SOLUTION 
    FOR INJECTION
medication_IOPAMIDOL 76 %
medication_LACTATED RINGERS INTRAVENOUS SOLUTION
medication_MIDAZOLAM 1 MG/ML INJECTION SOLUTION
medication_NALOXONE 0.4 MG/ML INJECTION SOLUTION
medication_ONDANSETRON HCL (PF) 4 MG/2 ML INJECTION SOLUTION
medication_OXYCODONE 5 MG TABLET
medication_PANTOPRAZOLE 40 MG TABLET,DELAYED RELEASE
medication_PROPOFOL 10 MG/ML INTRAVENOUS BOLUS (20 ML)
medication_SODIUM CHLORIDE 0.9 %
medication_SODIUM CHLORIDE 0.9 %
myalgia
obesity
past7Dprobhx_ICD_Acute kidney failure, unspecified
past7Dprobhx_ICD_Anemia, unspecified
past7Dprobhx_ICD_Anxiety disorder, unspecified
past7Dprobhx_ICD_Chest pain, unspecified
past7Dprobhx_ICD_Dizziness and giddiness
past7Dprobhx_ICD_Encounter for general adult medical examination 
    without abnormal findings
past7Dprobhx_ICD_Encounter for immunization
past7Dprobhx_ICD_Encounter for screening for malignant 
    neoplasm of colon
past7Dprobhx_ICD_F32.A
past7Dprobhx_ICD_Gastro-esophageal reflux disease 
    without esophagitis
past7Dprobhx_ICD_Hyperlipidemia, unspecified
past7Dprobhx_ICD_Hypokalemia
past7Dprobhx_ICD_Hypothyroidism, unspecified
past7Dprobhx_ICD_Mixed hyperlipidemia
past7Dprobhx_ICD_Obstructive sleep apnea (adult) (pediatric)
past7Dprobhx_ICD_Syncope and collapse
past7Dprobhx_ICD_Type 2 diabetes mellitus without complications
past7Dprobhx_ICD_Unspecified atrial fibrillation
probhx_ICD_Acute kidney failure, unspecified
probhx_ICD_Anemia, unspecified
probhx_ICD_Anxiety disorder, unspecified
probhx_ICD_Chest pain, unspecified
probhx_ICD_Dizziness and giddiness
probhx_ICD_Encounter for general adult medical examination without 
    abnormal findings
probhx_ICD_Encounter for immunization
probhx_ICD_Encounter for screening for malignant neoplasm of colon
probhx_ICD_F32.A
probhx_ICD_Gastro-esophageal reflux disease without esophagitis
probhx_ICD_Hyperlipidemia, unspecified
probhx_ICD_Hypokalemia
probhx_ICD_Hypothyroidism, unspecified
probhx_ICD_Mixed hyperlipidemia
probhx_ICD_Obstructive sleep apnea (adult) (pediatric)
probhx_ICD_Syncope and collapse
probhx_ICD_Type 2 diabetes mellitus without complications
probhx_ICD_Unspecified atrial fibrillation
transplant
troponin
vaccine_COVID-19 RS-AD26 (PF) Vaccine (Janssen)
vaccine_COVID-19 Vaccine, Unspecified
vaccine_COVID-19 mRNA (PF) Vaccine (Moderna)
vaccine_COVID-19 mRNA (PF) Vaccine (Pfizer)
vaccine_Flu Whole
vaccine_INFLUENZA, CCIV4
vaccine_Influenza
vaccine_Influenza High PF
vaccine_Influenza ID PF
vaccine_Influenza PF
vaccine_Influenza Vaccine, Quadrivalent, Adjuvanted
vaccine_Influenza, High-dose, Quadrivalent
vaccine_Influenza, Quadrivalent
vaccine_Influenza, Recombinant (RIV4)
vaccine_Influenza, Recombinant (Riv3)
vaccine_Influenza, Trivalent, Adjuvanted
vaccine_LAIV3
vaccine_Pneumococcal
vaccine_Pneumococcal Conjugate 13-valent
vaccine_Pneumococcal Polysaccharide
vaccine_TIVA
\end{verbatim}}
\onecolumn

\subsection{Missingness heatmaps}

This section plots missingness heatmaps of categorical and numerical features over time. Darker color means larger proportion of missing data.

\clearpage

\clearpage

\clearpage

\clearpage

\section{Additional MIMIC-IV Data Details}\label{app:mimic_data}
The Medical Information Mart for Intensive Care (MIMIC)-IV \citep{mimiciv_v1} database contains EHR data from patients admitted to critical care units 
from 2008--2019. MIMIC-IV is an update to MIMIC-III, adding time annotations placing each sample into a three-year time range, and removing elements from the old CareVue EHR system (before 2008). We approximate the year of each sample by taking the midpoint of its time range. The performance over time is evaluated on a \emph{yearly} basis. Our study uses MIMIC-IV-1.0. 

\begin{figure}[!h]
  \includegraphics[width=1.0\columnwidth]{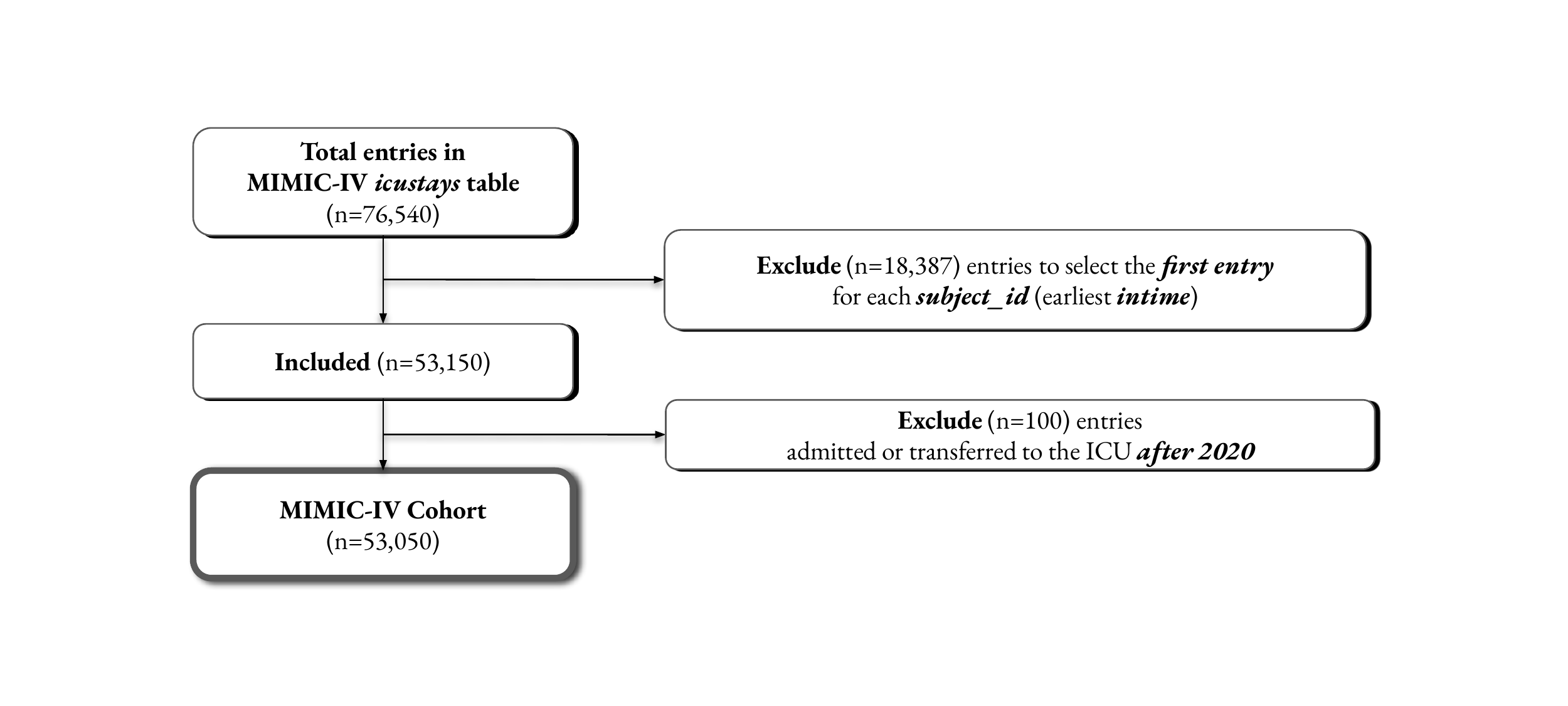}
  \caption{Cohort selection diagram - MIMIC-IV}
  \label{fig:mimic_iv_cohort}
\end{figure}

\begin{figure}[H]
  \includegraphics[width=1.0\columnwidth]{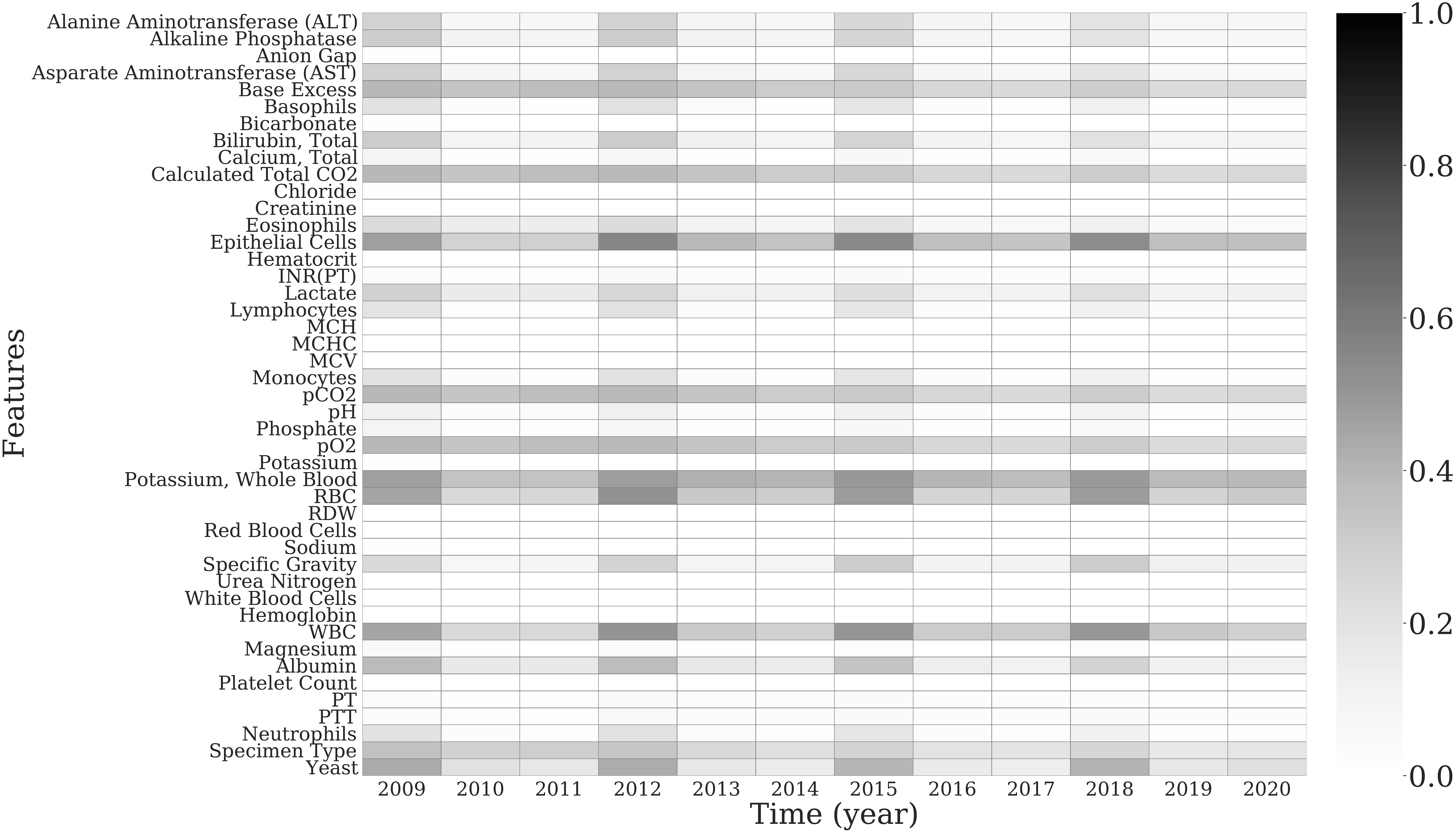}
  \caption{Missingness over time for labevents features in MIMIC-IV dataset after cohort selection. The darker the color, the larger the proportion of missing data.}
  \label{fig:heatmap_mimic_lab}
\end{figure}

\begin{figure}[H]
  \includegraphics[width=1.0\columnwidth]{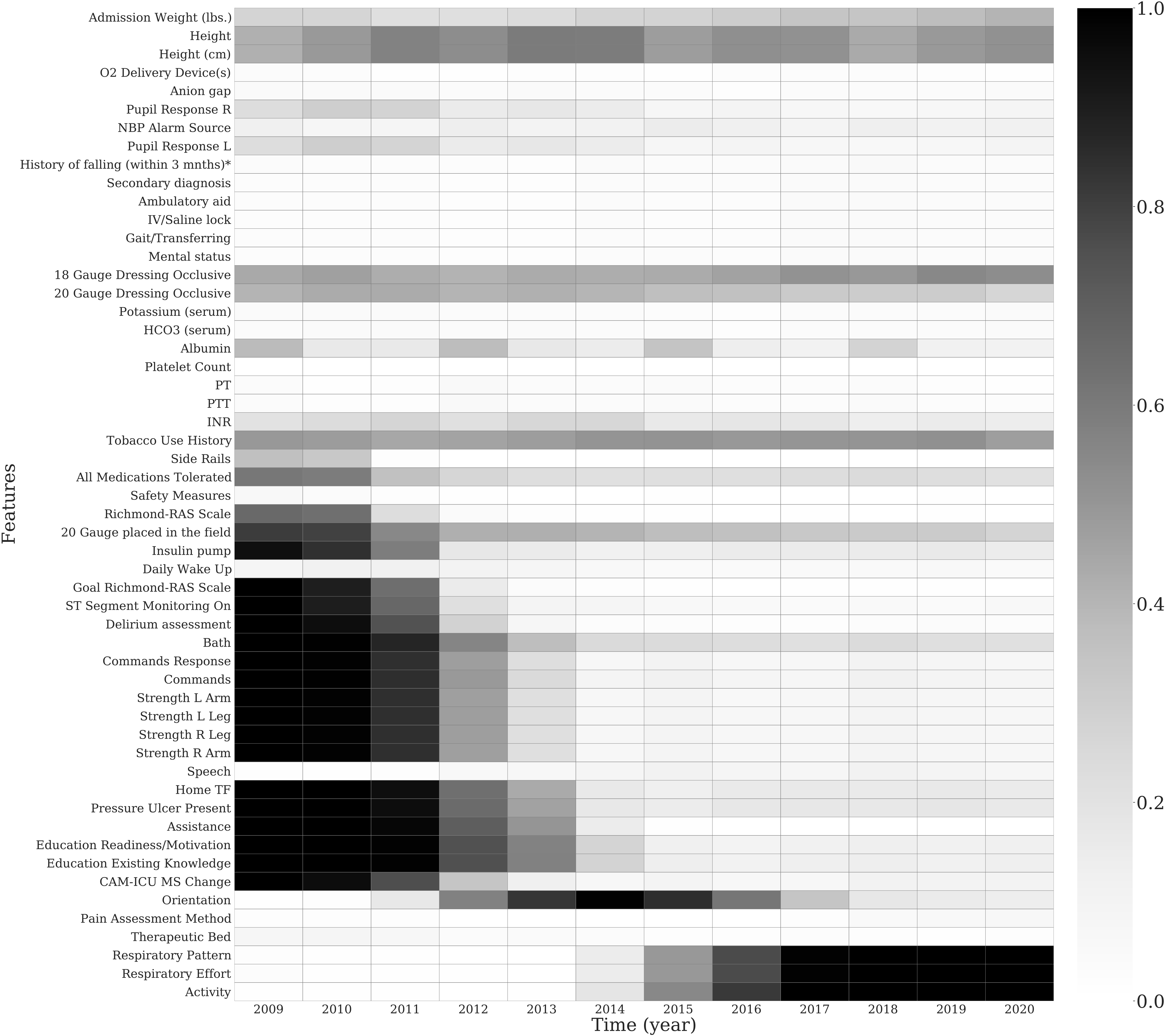}
  \caption{Missingness over time for chartevents features in MIMIC-IV dataset after cohort selection. The darker the color, the larger the proportion of missing data. (part 1)}
  \label{fig:heatmap_mimic_chart_1}
\end{figure}

\begin{figure}[H]
  \includegraphics[width=1.0\columnwidth]{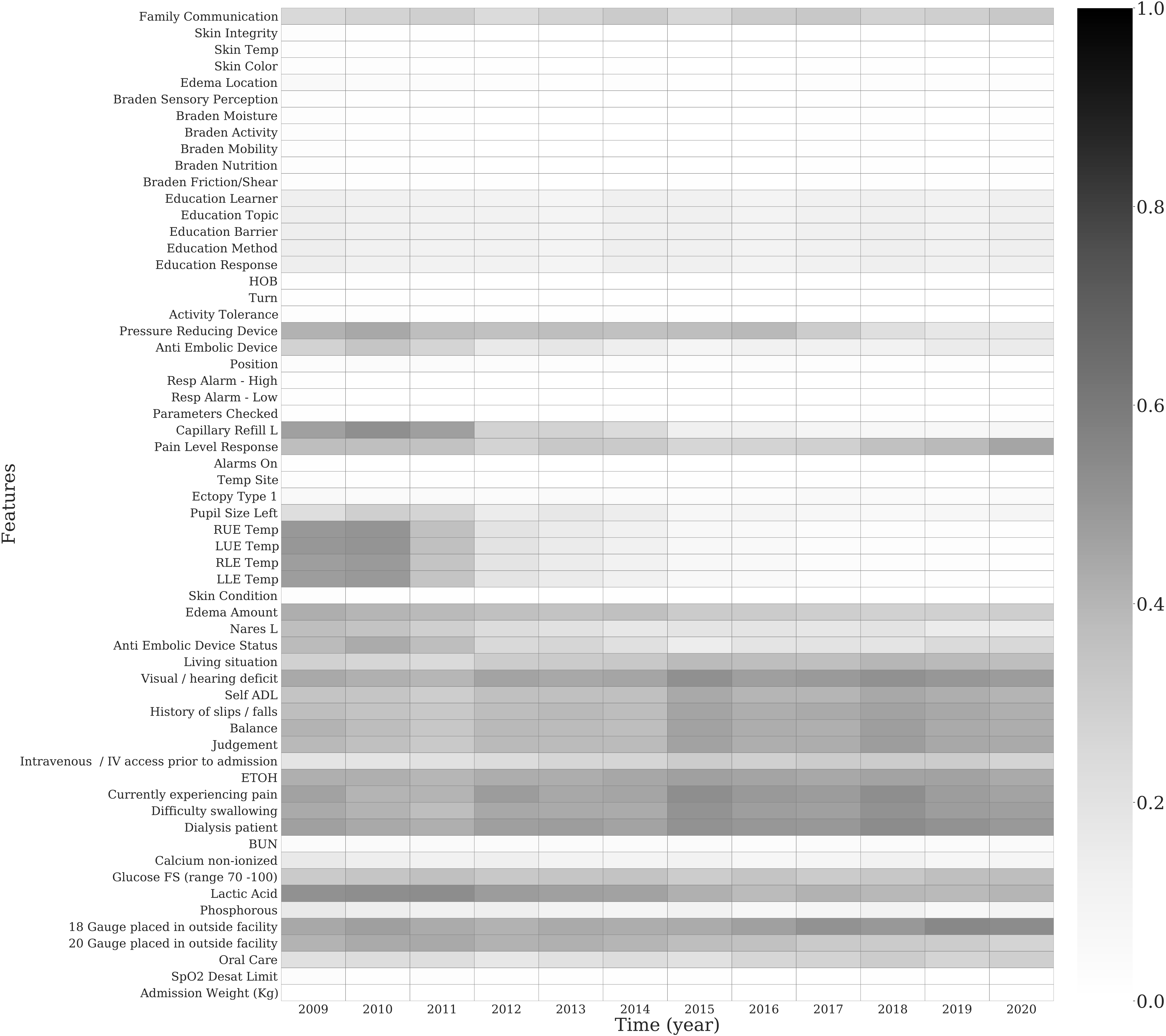}
  \caption{Missingness over time for chartevents features in MIMIC-IV dataset after cohort selection. The darker the color, the larger the proportion of missing data. (part 2)}
  \label{fig:heatmap_mimic_chart_2}
\end{figure}

\begin{figure}[H]
  \includegraphics[width=1.0\columnwidth]{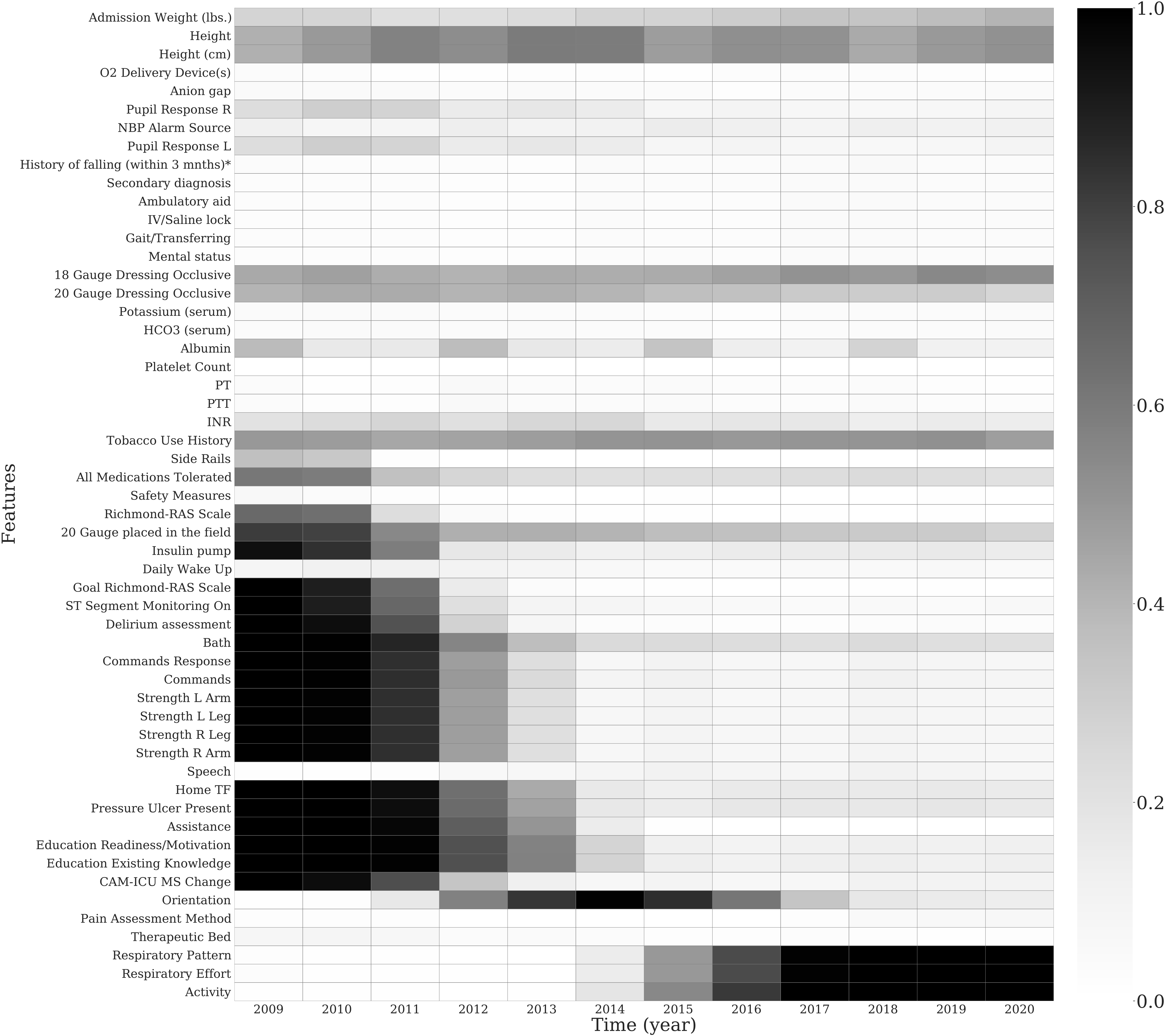}
  \caption{Missingness over time for chartevents features in MIMIC-IV dataset after cohort selection. The darker the color, the larger the proportion of missing data. (part 3)}
  \label{fig:heatmap_mimic_chart_3}
\end{figure}

\begin{itemize} 
    \item Data access: Users must create a Physionet account, become credentialed, and sign a data use agreement (DUA).
    \item Cohort selection: We select all patients in the \verb|icustays| table, filtering for their first encounter (minimum \verb|intime|), and defining a feature vector only using information available by the first 24 hrs of their first encounter. (Selection diagram in Figure \ref{fig:mimic_iv_cohort}). If there are multiple samples per patient, we filter to the first entry per patient, which corresponds to when a patient first enters the dataset. This corresponds to a particular interpretation of the prediction: when a patient first visits the ICU, given what we know about that patient, what is their estimated risk of in-ICU mortality?
    \item Outcome definition: The outcome of interest is in-ICU mortality, defined by comparing the \verb|outtime| of the patient's ICU visit with the patient's \verb|dod| (date of death, in the \verb|patients| table). As noted in the documentation, out-of-hospital mortality is not recorded. 
    \item Cohort characteristics: Cohort characteristics are given in Table \ref{tab:mimic_iv_characteristics}.
    \item Features: We list the features used in the MIMIC-IV datasets in Section \ref{app:sec_mimic_features}. We convert all categorical variables into dummy features, and apply standard scaling to numerical variables (subtract mean and divide by standard deviation). To create a fixed length feature vector, we take the most recent value of any patient history data available (e.g. most recent lab values).
    \item Missingness heat maps: are given in Figures \ref{fig:heatmap_mimic_lab}, \ref{fig:heatmap_mimic_chart_1}, \ref{fig:heatmap_mimic_chart_2},
    \ref{fig:heatmap_mimic_chart_3}.
\end{itemize}

\newpage
\subsection{Cohort Selection and Cohort Characteristics}

\begin{table*}[htb]
\caption{MIMIC-IV cohort characteristics, with count (\%) or median (Q1--Q3).}
\vspace{0.5em}
\label{tab:mimic_iv_characteristics}
    \centering
    \resizebox{0.9\columnwidth}{!}{%
    \begin{tabular}{lcccc}
    \toprule
    Characteristic &  & Missingness &       Type \\
    \midrule
    \textbf{Gender} &                   &               &              \\
    \hspace{2em}Female &    23,313 (43.9\%) &            -- &  categorical \\
    \hspace{2em}Male &    29,737 (56.1\%) &            -- &  categorical \\
    \textbf{Age at Admission} &        66 (54-78) &          0.0\% &   continuous \\
    \textbf{O2 Delivery Device(s)} &                   &               &              \\
    \hspace{2em}Use device &    33,359 (62.9\%) &            -- &  categorical \\
    \hspace{2em}None &    18,549 (35.0\%) &            -- &  categorical \\
    \hspace{2em}Missing &      1,142 (2.2\%) &            -- &  categorical \\
    \textbf{Pupil Response R} &                   &               &              \\
    \hspace{2em}Brisk  &    39,708 (74.9\%) &            -- &  categorical \\
    \hspace{2em}Sluggish &      4,603 (8.7\%) &            -- &  categorical \\
    \hspace{2em}Non-reactive &      1,812 (3.4\%) &            -- &  categorical \\
    \hspace{2em}Missing &     6,927 (13.1\%) &            -- &  categorical \\
    \textbf{first\_careunit} &                   &               &              \\
    \hspace{2em}Medical Intensive Care Unit (MICU) &    10,213 (19.3\%) &            -- &  categorical \\
    \hspace{2em}Surgical Intensive Care Unit (SICU) &     8,241 (15.5\%) &            -- &  categorical \\
    \hspace{2em}Medical/Surgical Intensive Care Unit (MICU/S... &     8,808 (16.6\%) &            -- &  categorical \\
    \hspace{2em}Cardiac Vascular Intensive Care Unit (CVICU) &     9,437 (17.8\%) &            -- &  categorical \\
    \hspace{2em}Coronary Care Unit (CCU) &     6,098 (11.5\%) &            -- &  categorical \\
    \hspace{2em}Trauma SICU (TSICU) &     6,947 (13.1\%) &            -- &  categorical \\
    \hspace{2em}Other &      3,306 (6.2\%) &            -- &  categorical \\
    \textbf{Anion Gap} &        13 (11-16) &          0.5\% &   continuous \\
    \textbf{Heart Rhythm} &                   &               &              \\
    \hspace{2em}SR (Sinus Rhythm) &    34,004 (64.1\%) &            -- &  categorical \\
    \hspace{2em}Abnormal heart rhythm &    18,657 (35.2\%) &            -- &  categorical \\
    \hspace{2em}Missing &        389 (0.7\%) &            -- &  categorical \\
    \textbf{Glucose FS (range 70 -100)} &     131 (110-164) &         32.7\% &   continuous \\
    \textbf{Eye Opening} &                   &               &              \\
    \hspace{2em}Spontaneously &    39,216 (73.9\%) &            -- &  categorical \\
    \hspace{2em}To Speech &     7,387 (13.9\%) &            -- &  categorical \\
    \hspace{2em}None &      4,538 (8.6\%) &            -- &  categorical \\
    \hspace{2em}To Pain &      1,702 (3.2\%) &            -- &  categorical \\
    \hspace{2em}Missing &        207 (0.4\%) &            -- &  categorical \\
    \textbf{Lactate} &           2 (1-2) &         22.0\% &   continuous \\
    \textbf{Motor Response} &                   &               &              \\
    \hspace{2em}Obeys Commands &    44,409 (83.7\%) &            -- &  categorical \\
    \hspace{2em}Localizes Pain &      3,419 (6.4\%) &            -- &  categorical \\
    \hspace{2em}Flex-withdraws &      1,673 (3.2\%) &            -- &  categorical \\
    \hspace{2em}No response &      2,930 (5.5\%) &            -- &  categorical \\
    \hspace{2em}Abnormal extension &        157 (0.3\%) &            -- &  categorical \\
    \hspace{2em}Abnormal Flexion &        238 (0.4\%) &            -- &  categorical \\
    \hspace{2em}Missing &        224 (0.4\%) &          --   &  categorical \\
    \textbf{Respiratory Pattern} &                   &               &              \\
    \hspace{2em}Regular &    29,373 (55.4\%) &          --   &  categorical \\
    \hspace{2em}Not regular &      1,739 (3.3\%) &        --     &  categorical \\
    \hspace{2em}Missing &    21,938 (41.4\%) &            -- &  categorical \\
    \textbf{Richmond-RAS Scale} &          0 (-1-0) &         15.4\% &  categorical \\
    \textbf{in-icu mortality} &                   &               &              \\
    \hspace{2em}0 &    49,716 (93.7\%) &         --    &  categorical \\
    \hspace{2em}1 &      3,334 (6.3\%) &          --   &  categorical \\
    \bottomrule
    \end{tabular}%
    }   
\end{table*}
\clearpage

\twocolumn
\subsection{Features}\label{app:sec_mimic_features}

{\tiny
\begin{verbatim}
18 Gauge Dressing Occlusive
18 Gauge placed in outside facility
20 Gauge Dressing Occlusive
20 Gauge placed in outside facility
20 Gauge placed in the field
Abdominal Assessment
Activity
Activity Tolerance
Admission Weight (Kg)
Admission Weight (lbs.)
Alanine Aminotransferase (ALT)
Alarms On
Albumin
Alkaline Phosphatase
All Medications Tolerated
Ambulatory aid
Anion Gap
Anion gap
Anti Embolic Device
Anti Embolic Device Status
Asparate Aminotransferase (AST)
Assistance
BUN
Balance
Base Excess
Basophils
Bath
Bicarbonate
Bilirubin, Total
Bowel Sounds
Braden Activity
Braden Friction/Shear
Braden Mobility
Braden Moisture
Braden Nutrition
Braden Sensory Perception
CAM-ICU MS Change
Calcium non-ionized
Calcium, Total
Calculated Total CO2
Capillary Refill L
Capillary Refill R
Chloride
Chloride (serum)
Commands
Commands Response
Cough Effort
Cough Type
Creatinine
Creatinine (serum)
Currently experiencing pain
Daily Wake Up
Delirium assessment
Dialysis patient
Diet Type
Difficulty swallowing
Dorsal PedPulse L
Dorsal PedPulse R
ETOH
Ectopy Type 1
Edema Amount
Edema Location
Education Barrier
Education Existing Knowledge
Education Learner
Education Method
Education Readiness/Motivation
Education Response
Education Topic
Eosinophils
Epithelial Cells
Eye Opening
Family Communication
Flatus
GU Catheter Size
Gait/Transferring
Glucose (serum)
Glucose FS (range 70 -100)
Goal Richmond-RAS Scale
HCO3 (serum)
HOB
HR
HR Alarm - High
HR Alarm - Low
Heart Rhythm
Height
Height (cm)
Hematocrit
Hematocrit (serum)
Hemoglobin
History of falling (within 3 mnths)*
History of slips / falls
Home TF
INR
INR(PT)
IV/Saline lock
Insulin pump
Intravenous  / IV access prior to admission
Judgement
LLE Color
LLE Temp
LLL Lung Sounds
LUE Color
LUE Temp
LUL Lung Sounds
Lactate
Lactic Acid
Living situation
Lymphocytes
MCH
MCHC
MCV
Magnesium
Mental status
Monocytes
Motor Response
NBP Alarm - High
NBP Alarm - Low
NBP Alarm Source
NBPd
NBPm
NBPs
Nares L
Nares R
Neutrophils
O2 Delivery Device(s)
Oral Care
Oral Cavity
Orientation
PT
PTT
Pain Assessment Method
Pain Cause
Pain Level
Pain Level Acceptable
Pain Level Response
Pain Location
Pain Management
Pain Present
Pain Type
Parameters Checked
Phosphate
Phosphorous
Platelet Count
Position
PostTib Pulses L
PostTib Pulses R
Potassium
Potassium (serum)
Potassium, Whole Blood
Pressure Reducing Device
Pressure Ulcer Present
Pupil Response L
Pupil Response R
Pupil Size Left
Pupil Size Right
RBC
RDW
RLE Color
RLE Temp
RLL Lung Sounds
RR
RUE Color
RUE Temp
RUL Lung Sounds
Radial Pulse L
Radial Pulse R
Red Blood Cells
Resp Alarm - High
Resp Alarm - Low
Respiratory Effort
Respiratory Pattern
Richmond-RAS Scale
ST Segment Monitoring On
Safety Measures
Secondary diagnosis
Self ADL
Side Rails
Skin Color
Skin Condition
Skin Integrity
Skin Temp
Sodium
Sodium (serum)
SpO2
SpO2 Alarm - High
SpO2 Alarm - Low
SpO2 Desat Limit
Specific Gravity
Specimen Type
Speech
Strength L Arm
Strength L Leg
Strength R Arm
Strength R Leg
Support Systems
Temp Site
Temperature F
Therapeutic Bed
Tobacco Use History
Turn
Untoward Effect
Urea Nitrogen
Urine Source
Verbal Response
Visual / hearing deficit
WBC
White Blood Cells
Yeast
admit_age
gender
pCO2
pH
pO2
\end{verbatim}}
\onecolumn

\subsection{Missingness heatmaps}

\clearpage

\clearpage

\clearpage

\section{Additional OPTN (Liver) Data Details}\label{app:optn_data}
The Organ Procurement and Transplantation Network (OPTN) database tracks organ donation and transplant events in the U.S. Our study uses data from candidates on the liver transplant wait list. The performance over time is evaluated on a \emph{yearly} basis.

\begin{figure}[ht]
  \includegraphics[width=1.0\textwidth]{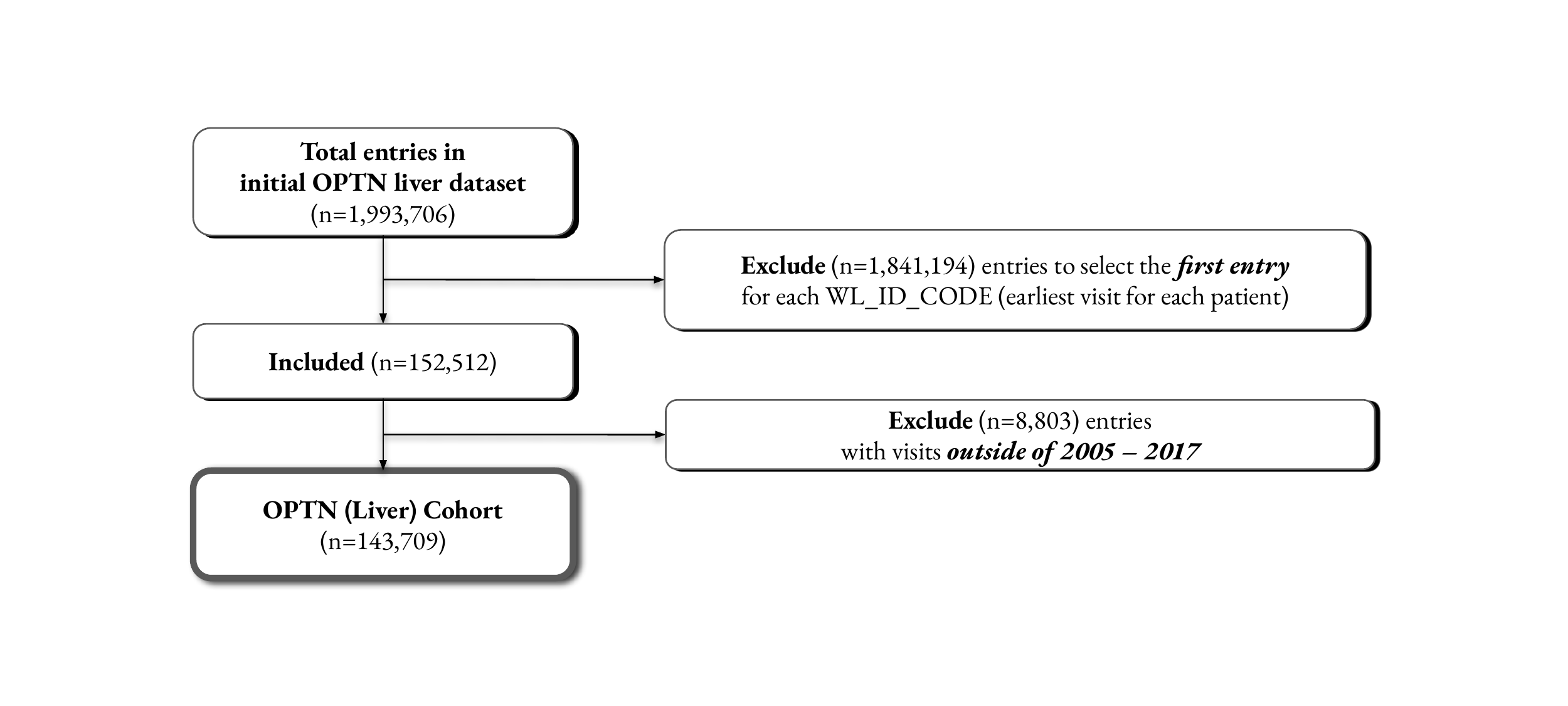}
  \vspace{-3em}
  \caption{Cohort selection diagram - OPTN (Liver)}
  \label{fig:optn_liver_cohort}
\end{figure}

\begin{figure}[H]
  \includegraphics[width=1.0\columnwidth]{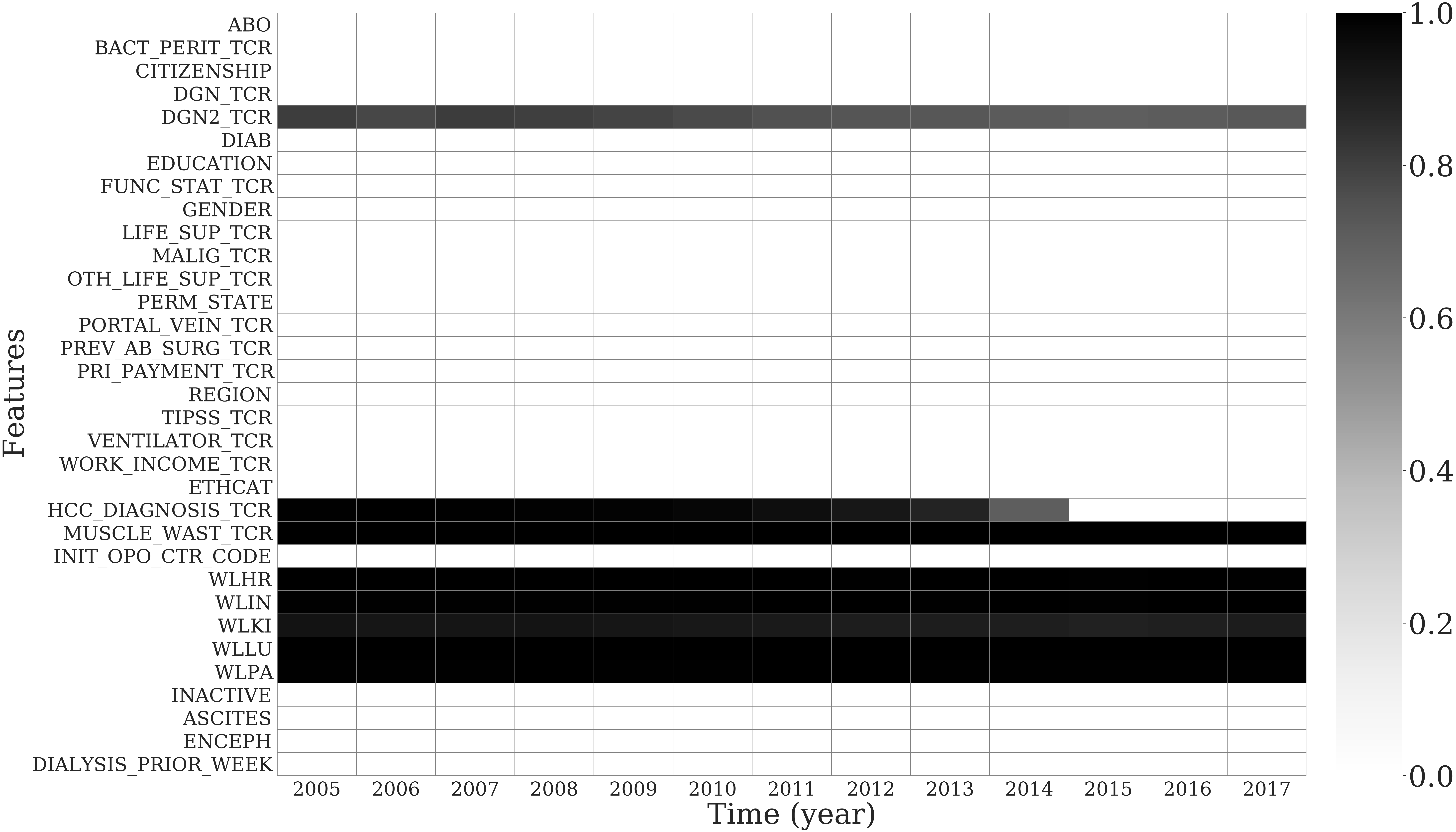}
  \caption{Missingness over time for categorical features in OPTN (Liver) dataset after cohort selection. The darker the color, the larger the proportion of missing data.}
  \label{fig:heatmap_optn_liver_cate}
\end{figure}

\begin{figure}[H]
  \includegraphics[width=1.0\columnwidth]{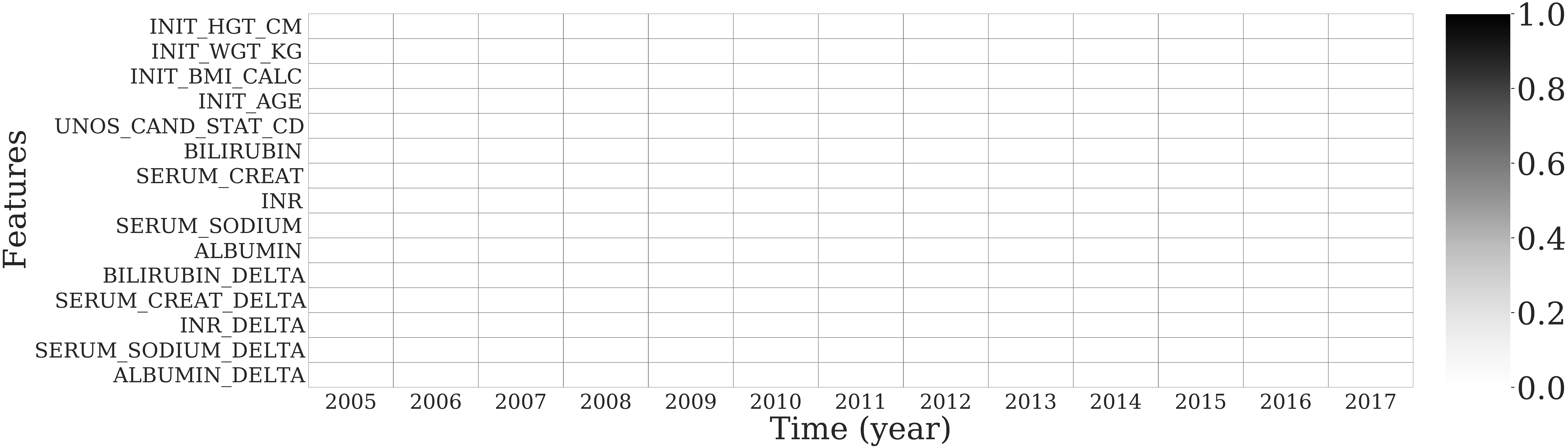}
  \caption{Missingness over time for numerical features in OPTN (Liver) dataset after cohort selection. The darker the color, the larger the proportion of missing data. (Near-zero missingness here.)}
  \label{fig:heatmap_optn_liver_num}
\end{figure}

\begin{itemize}
    \item First, we provide the disclaimer: 
``The data reported here have been supplied by the United Network for Organ Sharing as the contractor for the Organ Procurement and transplantation Network. The interpretation and reporting of these data are the responsibility of the author(s) and in no way should be seen as an official policy of or interpretation by the OPTN or the U.S. Government''.
    \item Data access: After signing the Data Use Agreement - I from Organ Procedurement And Transplantation network, users can access the OPTN (Liver) dataset.
    \item Cohort selection: The cohort consists of liver transplant candidates on the waiting list (2005-2017). We follow the same pipeline as \citet{byrd2021predicting} to extract the data, except that we select the first record for each patient. Cohort selection diagrams are given in Figures \ref{fig:optn_liver_cohort}. This corresponds to a particular interpretation of the prediction: when a patient is first added to the transplant list, given what we know about that patient, what is their estimated risk of 180-day mortality?
    \item Outcome definition: 180-day mortality from when the patient was first added to the list
    \item Cohort characteristics: Cohort characteristics are given in Table \ref{tab:optn_liver_characteristics}.
    \item Features: We list the features used in the OPTN liver dataset in Section \ref{app:sec_optn_features}. We convert all categorical variables into dummy features, and apply standard scaling to numerical variables (subtract mean and divide by standard deviation).
    \item Missingness heat maps: are given in Figures \ref{fig:heatmap_optn_liver_cate} and \ref{fig:heatmap_optn_liver_num}.
\end{itemize}

\clearpage
\subsection{Cohort Selection and Cohort Characteristics}

\begin{table*}[!htb]
\caption{OPTN (Liver) cohort characteristics, with count (\%) or median (Q1 -- Q3).}
\vspace{0.5em}
\label{tab:optn_liver_characteristics}
    \centering
    \resizebox{0.8\columnwidth}{!}{%
    \begin{tabular}{lccc}
    \toprule
    Feature name (value) &   & Empty (ratio) &       Type \\
    \midrule
    \textbf{Gender} &                        &               &              \\
    \hspace{2em}Male &         92,560 (64.4\%) &            -- &  categorical \\
    \hspace{2em}Female &         51,149 (35.6\%) &            -- &  categorical \\
    \textbf{INIT\_AGE} &             56 (49-62) &          0.0\% &   continuous \\
    \textbf{FUNC\_STAT\_TCR} &    2,070 (2,050-2,080) &          0.0\% &  categorical \\
    \textbf{INIT\_OPO\_CTR\_CODE} &  11,036 (3,782-19,282) &          0.0\% &  categorical \\
    \textbf{ALBUMIN} &                3 (3-4) &          0.0\% &   continuous \\
    \textbf{HCC\_DIAGNOSIS\_TCR} &                        &               &              \\
    \hspace{2em}No &         31,390 (21.8\%) &            -- &  categorical \\
    \hspace{2em}Yes &          11,312 (7.9\%) &            -- &  categorical \\
    \hspace{2em}Missing &        101,007 (70.3\%) &            -- &  categorical \\
    \textbf{PERM\_STATE} &                        &               &              \\
    \hspace{2em}CA &         19,645 (13.7\%) &            -- &  categorical \\
    \hspace{2em}TX &         14,692 (10.2\%) &            -- &  categorical \\
    \hspace{2em}NY &           9,976 (6.9\%) &            -- &  categorical \\
    \hspace{2em}GA &           4,052 (2.8\%) &            -- &  categorical \\
    \hspace{2em}MD &           4,050 (2.8\%) &            -- &  categorical \\
    \hspace{2em}FL &           7,602 (5.3\%) &            -- &  categorical \\
    \hspace{2em}PA &           8,013 (5.6\%) &            -- &  categorical \\
    \hspace{2em}MI &           3,989 (2.8\%) &            -- &  categorical \\
    \hspace{2em}Other &         71,007 (49.4\%) &            -- &  categorical \\
    \textbf{EDUCATION} &                4 (3-5) &          0.0\% &  categorical \\
    \textbf{ASCITES} &                2 (1-2) &          0.0\% &  categorical \\
    \textbf{MORTALITY\_180D} &                        &               &              \\
    \hspace{2em}1 &           4,635 (3.2\%) &            -- &  categorical \\
    \hspace{2em}0 &        139,074 (96.8\%) &            -- &  categorical \\
    \bottomrule
    \end{tabular}%
    }
\end{table*}
\clearpage

\twocolumn
\subsection{Features}\label{app:sec_optn_features}
{\small
\begin{verbatim}
ABO
BACT_PERIT_TCR
CITIZENSHIP
DGN_TCR
DGN2_TCR
DIAB
EDUCATION
FUNC_STAT_TCR
GENDER
LIFE_SUP_TCR
MALIG_TCR
OTH_LIFE_SUP_TCR
PERM_STATE
PORTAL_VEIN_TCR
PREV_AB_SURG_TCR
PRI_PAYMENT_TCR
REGION
TIPSS_TCR
VENTILATOR_TCR
WORK_INCOME_TCR
ETHCAT
HCC_DIAGNOSIS_TCR
MUSCLE_WAST_TCR
INIT_OPO_CTR_CODE
WLHR
WLIN
WLKI
WLLU
WLPA
INACTIVE
ASCITES
ENCEPH
DIALYSIS_PRIOR_WEEK
INIT_HGT_CM
INIT_WGT_KG
INIT_BMI_CALC
INIT_AGE
UNOS_CAND_STAT_CD
BILIRUBIN
SERUM_CREAT
INR
SERUM_SODIUM
ALBUMIN
BILIRUBIN_DELTA
SERUM_CREAT_DELTA
INR_DELTA
SERUM_SODIUM_DELTA
ALBUMIN_DELTA
\end{verbatim}}
\onecolumn

\subsection{Missingness heatmaps}

\clearpage

\section{Logistic Regression Coefficients from Splitting by Patient}

To help with intuition in important features for the predictive task on each dataset, here we have the coefficients of logistic regression models trained from splitting by patient.

\begin{table*}[ht]
\caption{SEER (Breast) top 10 important features for LR models, all-period training.}
\vspace{0.5em}
\label{tab:seer_breast_coef}
\centering
    \begin{tabular}{lc}
    \toprule
    Feature &  Coefficient \\
    \midrule
    SEER historic stage A (1973-2015)\_Distant &    -2.113944 \\
    SEER historic stage A (1973-2015)\_Localized &     1.676493 \\
    Regional nodes examined (1988+)\_95.0 &    -1.167844 \\
    CS lymph nodes (2004-2015)\_750 &     1.100824 \\
    CS lymph nodes (2004-2015)\_755 &     1.023753 \\
    Histologic Type ICD-O-3\_8530 &    -0.913494 \\
    Histologic Type ICD-O-3\_8543 &     0.902798 \\
    Breast - Adjusted AJCC 6th T (1988-2015)\_T4d &     0.899491 \\
    Histologic Type ICD-O-3\_8211 &     0.877848 \\
    EOD 10 - extent (1988-2003)\_85 &    -0.791136 \\
    \bottomrule
    \end{tabular}%
\end{table*}

\begin{table*}[ht]
\caption{SEER (Colon) top 10 important features for LR models, all-period training.}
\vspace{0.5em}
\label{tab:seer_colon_coef}
\centering
    \begin{tabular}{lc}
    \toprule
    Feature &  Coefficient \\
    \midrule
    Reason no cancer-directed surgery\_Surgery performed &     2.360161 \\
    Regional nodes positive (1988+)\_00 &     1.897706 \\
    Regional nodes positive (1988+)\_01 &     1.872008 \\
    modified AJCC stage 3rd (1988-2003)\_40 &    -1.787481 \\
    EOD 10 - extent (1988-2003)\_13 &     1.766066 \\
    Reason no cancer-directed surgery\_Not recommended,  &    -1.752474 \\
    contraindicated due to other cond; autopsy only (1973-2002)&\\
    EOD 10 - extent (1988-2003)\_85 &    -1.732619 \\
    EOD 10 - extent (1988-2003)\_70 &    -1.704333 \\
    CS mets at dx (2004-2015)\_99 &     1.619905 \\
    CS mets at dx (2004-2015)\_00 &     1.609454 \\
    \bottomrule
    \end{tabular}%
\end{table*}

\begin{table*}[ht]
\caption{SEER (Lung) top 10 important features for LR models, all-period training.}
\vspace{0.5em}
\label{tab:seer_lung_coef}
\centering
    \begin{tabular}{lc}
    \toprule
    Feature &  Coefficient \\
    \midrule
    Histologic Type ICD-O-3\_8240 &     2.514539 \\
    EOD 4 - nodes (1983-1987)\_0 &     2.074730 \\
    EOD 4 - nodes (1983-1987)\_7 &    -1.777530 \\
    EOD 10 - size (1988-2003)\_140 &    -1.587893 \\
    Histologic Type ICD-O-3\_8141 &    -1.546566 \\
    CS tumor size (2004-2015)\_998.0 &    -1.515856 \\
    EOD 4 - nodes (1983-1987)\_6 &    -1.497022 \\
    Type of Reporting Source\_Nursing/convalescent home/hospice &    -1.338998 \\
    CS mets at dx (2004-2015)\_51 &    -1.326595 \\
    EOD 10 - size (1988-2003)\_150 &    -1.326196 \\
    \bottomrule
    \end{tabular}%
\end{table*}

\begin{table*}[ht]
\caption{CDC COVID-19 top 10 important features for LR models, all-period training.}
\vspace{0.5em}
\label{tab:cdc_covid_coef}
\centering
    \begin{tabular}{lc}
    \toprule
    Feature &  Coefficient \\
    \midrule
    res\_state\_DE &     2.202055 \\
    age\_group\_0 - 9 Years &    -2.114818 \\
    age\_group\_80+ Years &     1.965279 \\
    age\_group\_10 - 19 Years &    -1.681099 \\
    res\_state\_GA &     1.391469 \\
    age\_group\_70 - 79 Years &     1.379589 \\
    res\_county\_WICHITA &     1.290644 \\
    age\_group\_20 - 29 Years &    -1.189734 \\
    res\_county\_SUMNER &    -1.135073 \\
    mechvent\_yn\_Yes &     1.117372 \\
    \bottomrule
    \end{tabular}%
\end{table*}

\begin{table*}[ht]
\caption{SWPA COVID-19 top 10 important features for LR models according to experiments splitting by patient.}
\vspace{0.5em}
\label{tab:pa_covid_coef}
\centering
    \resizebox{1\columnwidth}{!}{%
    \begin{tabular}{lc}
    \toprule
    Feature &  Coefficient \\
    \midrule
    age\_bin\_(70, 200]\_0 &    -0.781337 \\
    age\_bin\_(70, 200]\_1 &     0.780673 \\
    medication\_FENTANYL (PF) 50 MCG/ML INJECTION SOLUTION\_0.0 &     0.651419 \\
    medication\_EPINEPHRINE 0.3 MG/0.3 ML INJECTION, AUTO-INJECTOR\_nan &    -0.627565 \\
    medication\_HYDROCORTISONE SOD SUCCINATE (PF) 100 MG/2 ML SOLUTION FOR INJECTION\_0.0 &     0.544222 \\
    medication\_HYDROCODONE 5 MG-ACETAMINOPHEN 325 MG TABLET\_nan &    -0.520368 \\
    medication\_DEXAMETHASONE SODIUM PHOSPHATE 4 MG/ML INJECTION SOLUTION\_0.0 &     0.502954 \\
    medication\_ASPIRIN 81 MG TABLET,DELAYED RELEASE\_nan &    -0.479100 \\
    bmi\_nan &    -0.427569 \\
    age\_bin\_(60, 70]\_0 &    -0.380688 \\
    \bottomrule
    \end{tabular}%
    }
\end{table*}

\begin{table*}[ht]
\caption{MIMIC-IV top 10 important features for LR models, all-period training.}
\vspace{0.5em}
\label{tab:mimic_iv_coef}
\centering
    \resizebox{0.6\columnwidth}{!}{%
    \begin{tabular}{lc}
    \toprule
    Feature &  Coefficient \\
    \midrule
    O2 Delivery Device(s)\_None &    -0.307334 \\
    Eye Opening\_None &     0.301737 \\
    admit\_age &     0.299712 \\
    O2 Delivery Device(s)\_Nasal cannula &    -0.248463 \\
    Motor Response\_Obeys Commands &    -0.230931 \\
    Pupil Response L\_Non-reactive &     0.223776 \\
    Richmond-RAS Scale\_ 0  Alert and calm &    -0.205476 \\
    Temp Site\_Blood &    -0.204514 \\
    HR\_0.0 &     0.197299 \\
    Diet Type\_NPO &     0.195156 \\
    \bottomrule
    \end{tabular}%
    }
\end{table*}

\begin{table*}[ht]
\caption{OPTN (Liver) top 10 important features for LR models, all-period training.}
\vspace{0.5em}
\label{tab:optn_liver_coef}
\centering
    \resizebox{0.5\columnwidth}{!}{%
    \begin{tabular}{lc}
    \toprule
    Feature &  Coefficient \\
    \midrule
    SERUM\_CREAT\_DELTA &     0.660589 \\
    FUNC\_STAT\_TCR\_2020.0 &     0.241507 \\
    FUNC\_STAT\_TCR\_2080.0 &    -0.236288 \\
    DGNC\_4110.0 &    -0.234680 \\
    REGION\_5.0 &     0.223940 \\
    EDUCATION\_998.0 &     0.218549 \\
    ASCITES\_3.0 &     0.218329 \\
    ASCITES\_1.0 &    -0.214076 \\
    INIT\_OPO\_CTR\_CODE\_1054 &    -0.209265 \\
    INIT\_OPO\_CTR\_CODE\_4743 &    -0.207778 \\
    \bottomrule
    \end{tabular}%
    }
\end{table*}
\clearpage

\section{Diagnostic plots}\label{app:diagnostic_plot}
We select important features to highlight based on high positive proportions and high relative feature importance. 

\subsection{SEER (Breast)}
\begin{figure}[H]
  \includegraphics[width=1.0\columnwidth]{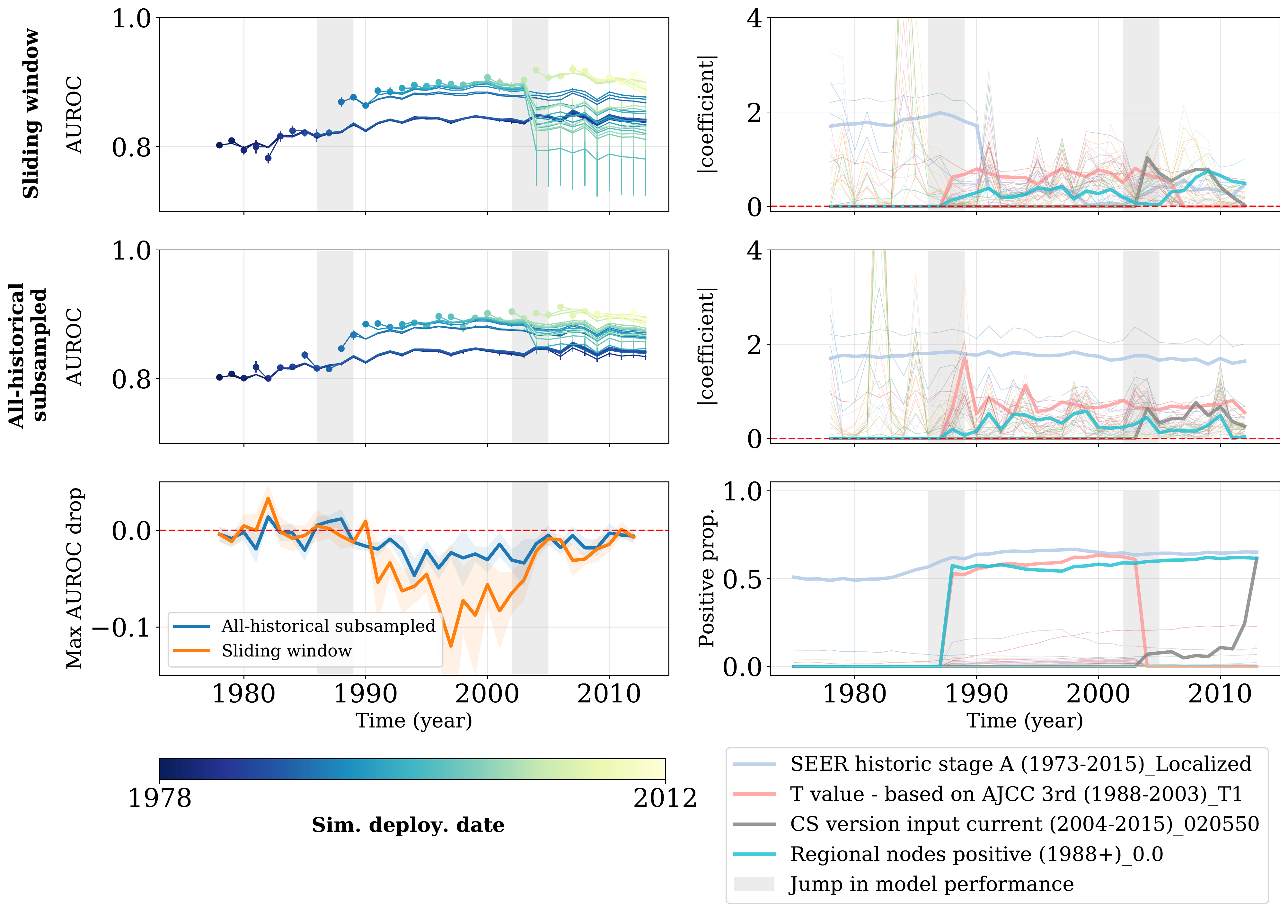}
  \caption{Diagnostic plot of SEER (Breast) dataset. The left column includes absolute AUROC versus time for both sliding window and all-historical subsampled, and the maximum AUROC drop for each trained model. The right column provides the absolute coefficients of each trained model from both regimes, and positive proportion of the significant features over time. As shown in the gray highlighted region, there are jumps in performance around 1988 and 2003, which coincides with the introducing and removal of several features (e.g. T value - based on AJCC 3rd (1988-2003)\_T1). The latency of jumps in coefficients are caused by length of sliding window.}
  \label{fig:diagnositic_plot_seer_breast}
\end{figure}

\clearpage

\subsection{SEER (Colon)}
\begin{figure}[H]
  \includegraphics[width=1.0\columnwidth]{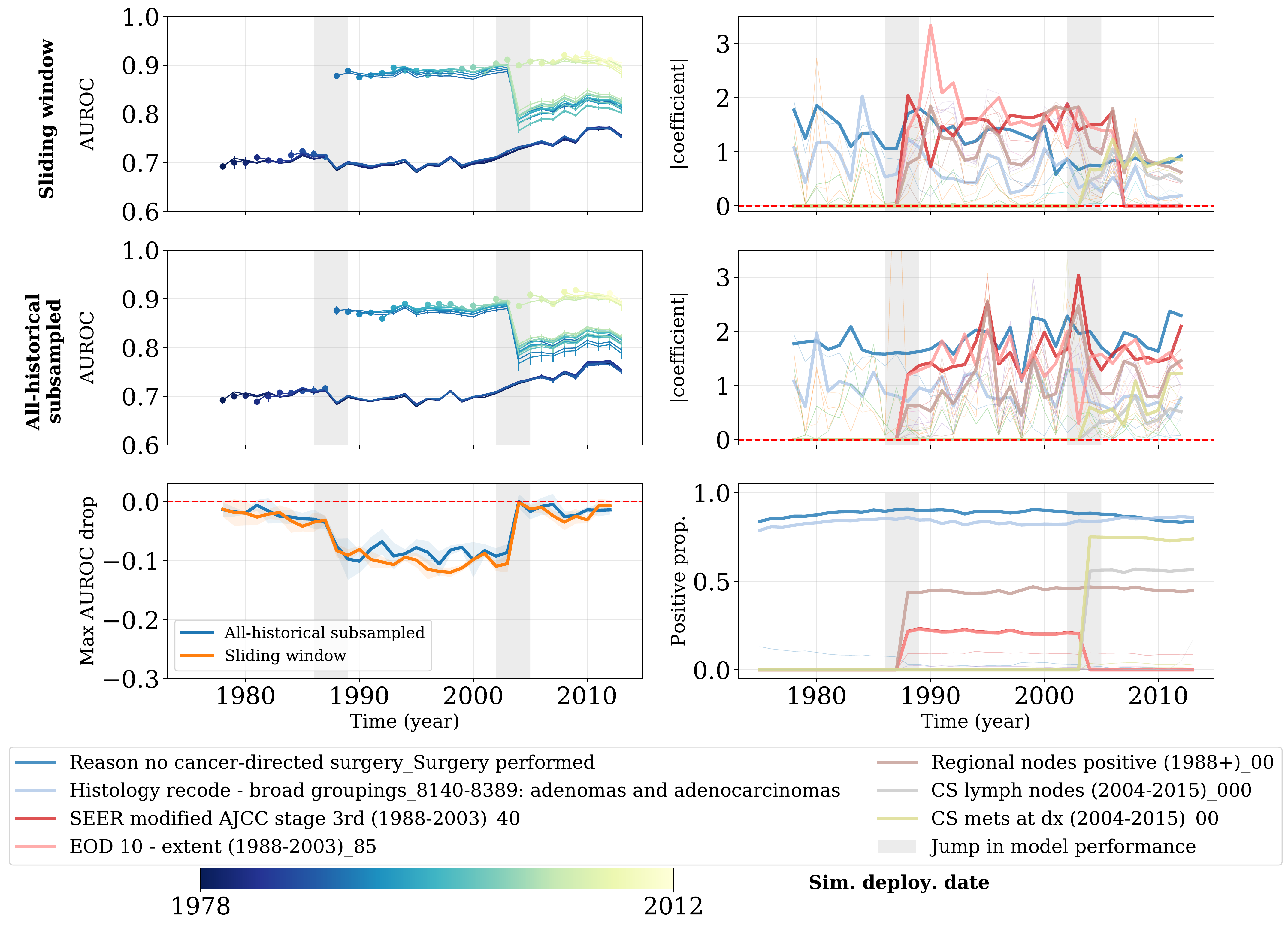}
  \caption{Diagnostic plot of SEER (Colon) dataset. The left column includes absolute AUROC versus time for both sliding window and all-historical subsampled, and the maximum AUROC drop for each trained model. The right column provides the absolute coefficients of each trained model from both regimes, and positive proportion of the significant features over time. As shown in the gray highlighted region, there are jumps in performance around 1988 and 2003, which coincides with the introducing and removal of several features (e.g. SEER modified AJCC stage 3rd (1988-2003)\_40). The latency of jumps in coefficients are caused by length of sliding window.}
  \label{fig:diagnositic_plot_seer_colon}
\end{figure}

\clearpage

\subsection{SEER (Lung)}
\begin{figure}[H]
  \includegraphics[width=1.0\columnwidth]{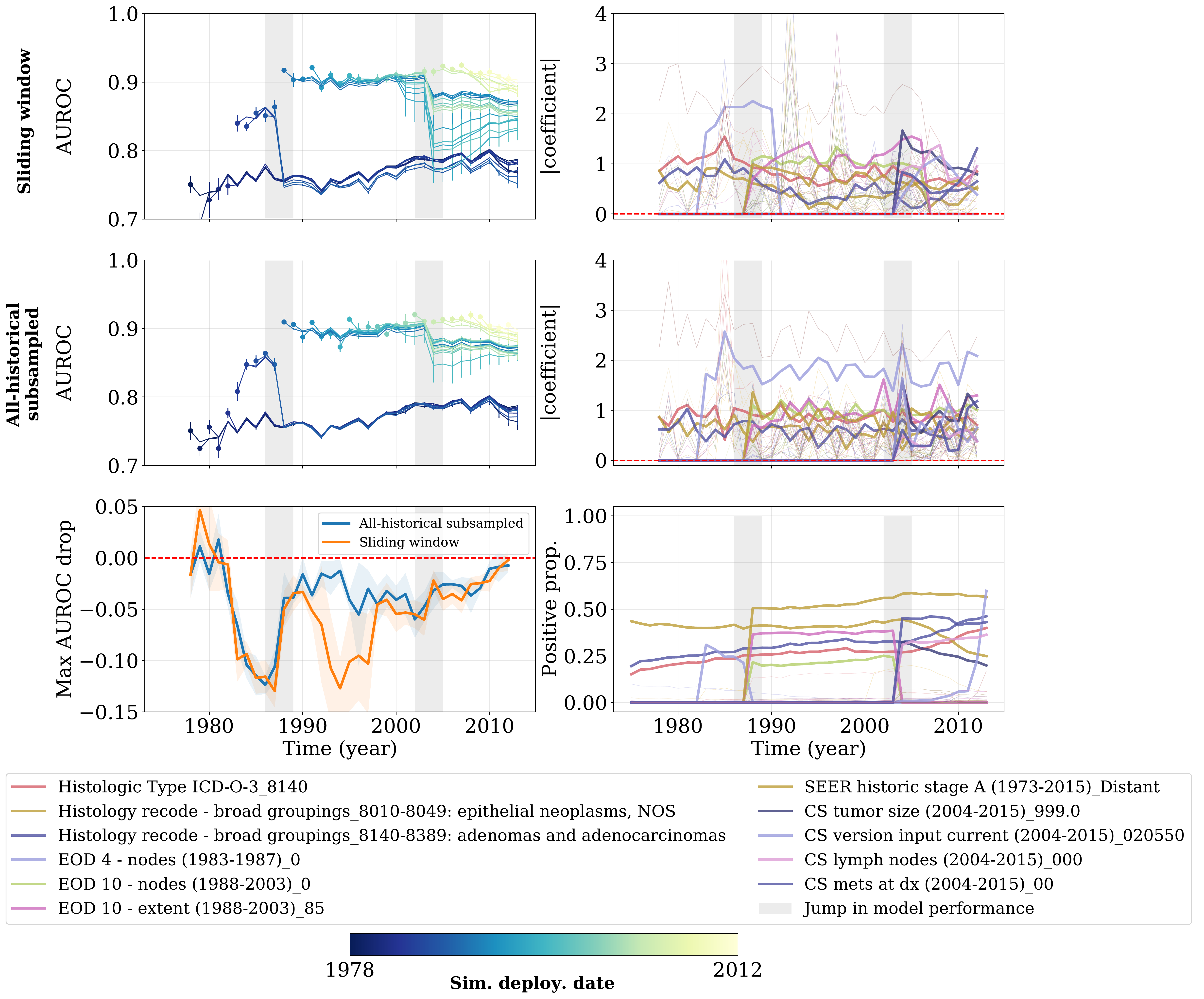}
  \caption{Diagnostic plot of SEER (Lung) dataset. The left column includes absolute AUROC versus time for both sliding window and all-historical subsampled, and the maximum AUROC drop for each trained model. The right column provides the absolute coefficients of each trained model from both regimes, and positive proportion of the significant features over time. As shown in the gray highlighted region, there are jumps in performance around 1988 and 2003, which coincides with the introducing and removal of several features (e.g. EOD 10 - nodes (1988-2013)\_0 \& EOD 10 - extent (1988-2003)\_85). The latency of jumps in coefficients are caused by length of sliding window.}
  \label{fig:diagnositic_plot_seer_lung}
\end{figure}

\clearpage

\subsection{CDC COVID-19}
\begin{figure}[H]
  \includegraphics[width=1.0\columnwidth]{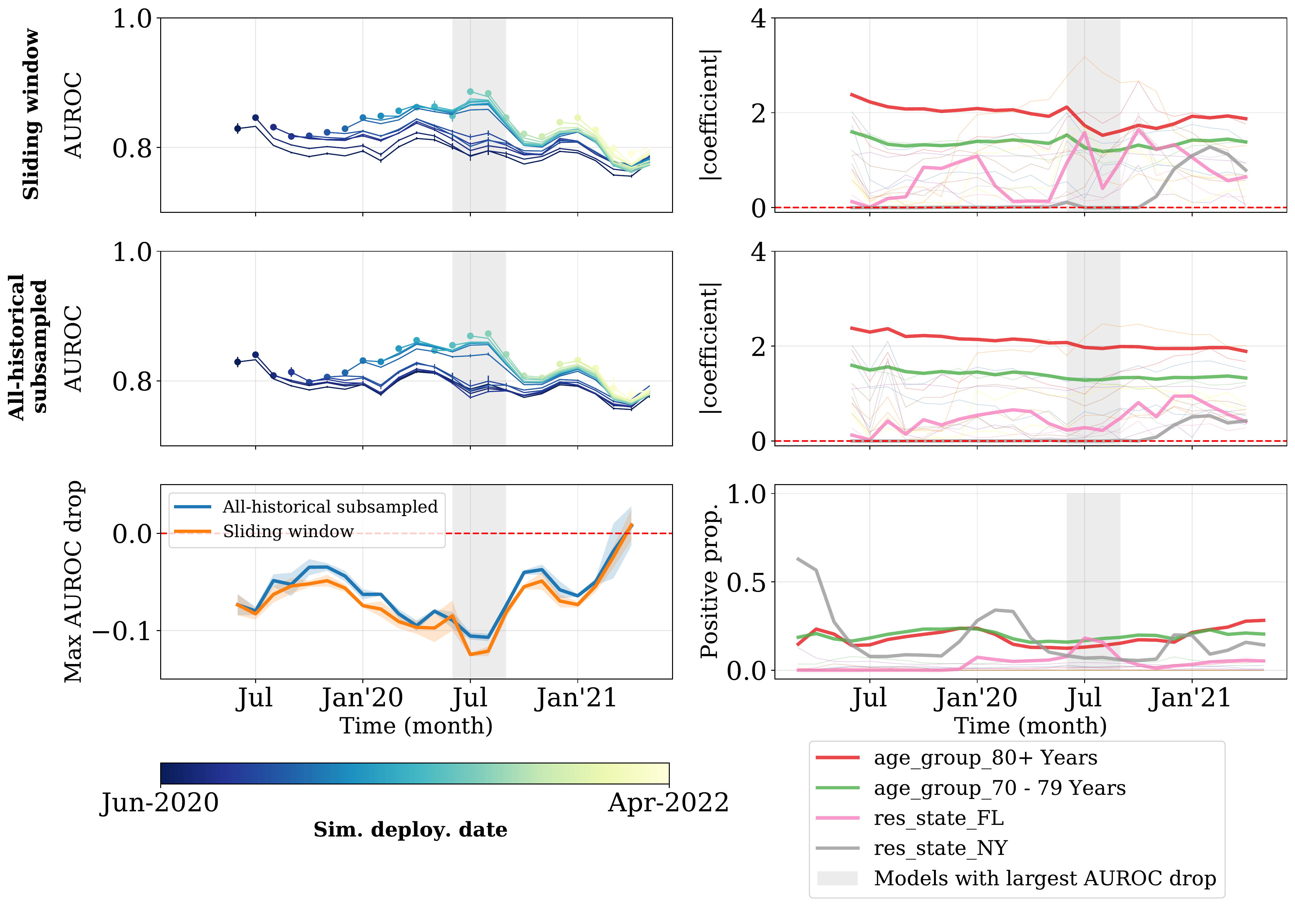}
  \caption{Diagnostic plot of CDC COVID-19. The left column includes absolute AUROC versus time for both sliding window and all-historical subsampled, and the maximum AUROC drop for each trained model. The right column provides the absolute coefficients of each trained model from both regimes, and positive proportion of the significant features over time. As shown in the gray highlighted region, the models trained around June 2021 suffer the largest maximum AUROC drop, coinciding with a shift in distribution of ages (Figure \ref{fig:stack_age_group}) and states (Figure \ref{fig:stack_state_residence}). The latency of jumps in coefficients are caused by length of sliding window.}
  \label{fig:diagnositic_plot_cdc_covid}
\end{figure}

\clearpage

\subsection{SWPA COVID-19}
\begin{figure}[H]
  \includegraphics[width=1.0\columnwidth]{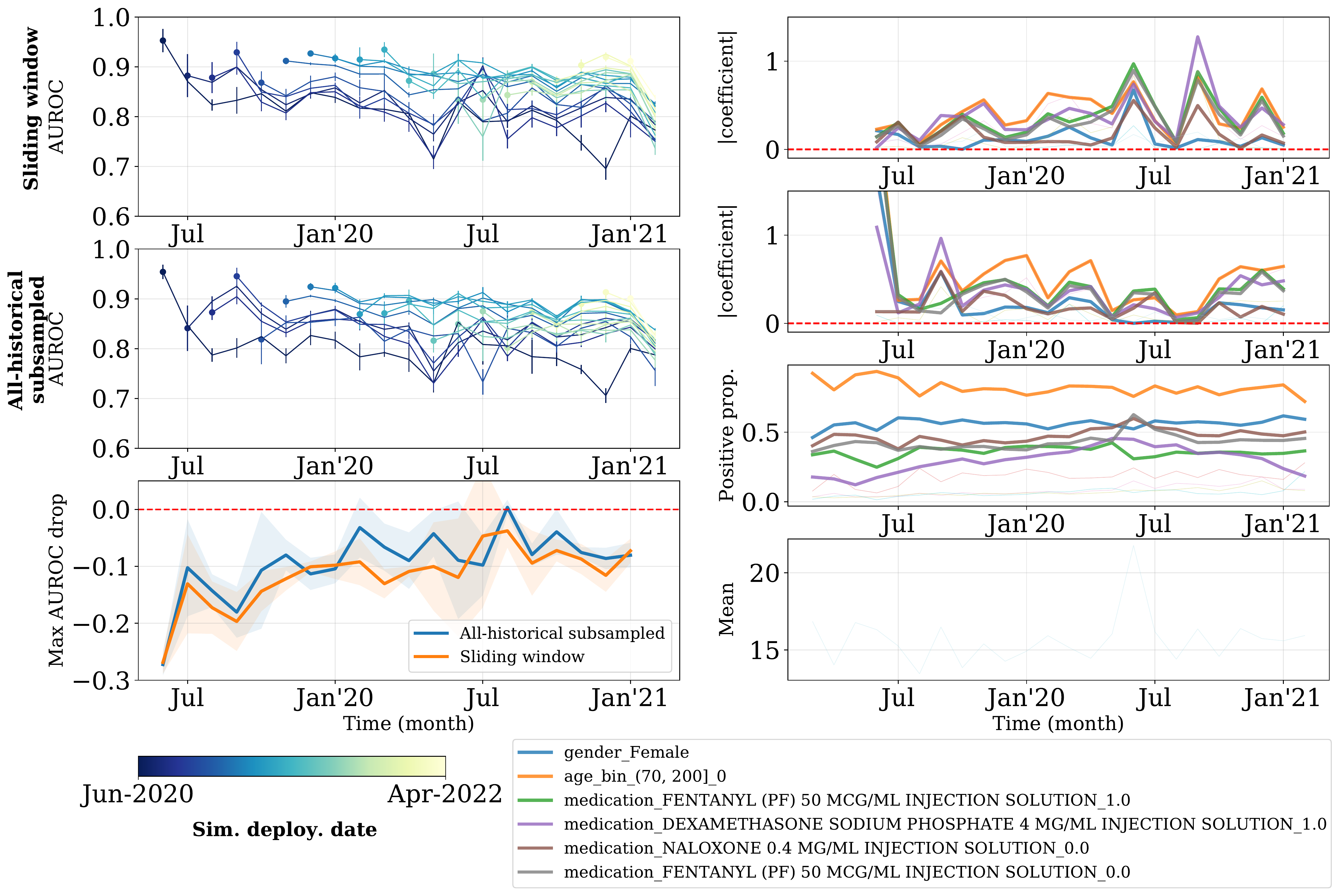}
  \caption{Diagnostic plot of SWPA COVID-19. The left column includes absolute AUROC versus time for both sliding window and all-historical subsampled, and the maximum AUROC drop for each trained model. The right column provides the absolute coefficients of each trained model from both regimes, and positive proportion of the significant features over time. One of the hypotheses for relatively large uncertainty is smaller sample size.}
  \label{fig:diagnositic_plot_pa_covid}
\end{figure}

\clearpage

\subsection{MIMIC-IV}
\begin{figure}[H]
  \includegraphics[width=1.0\columnwidth]{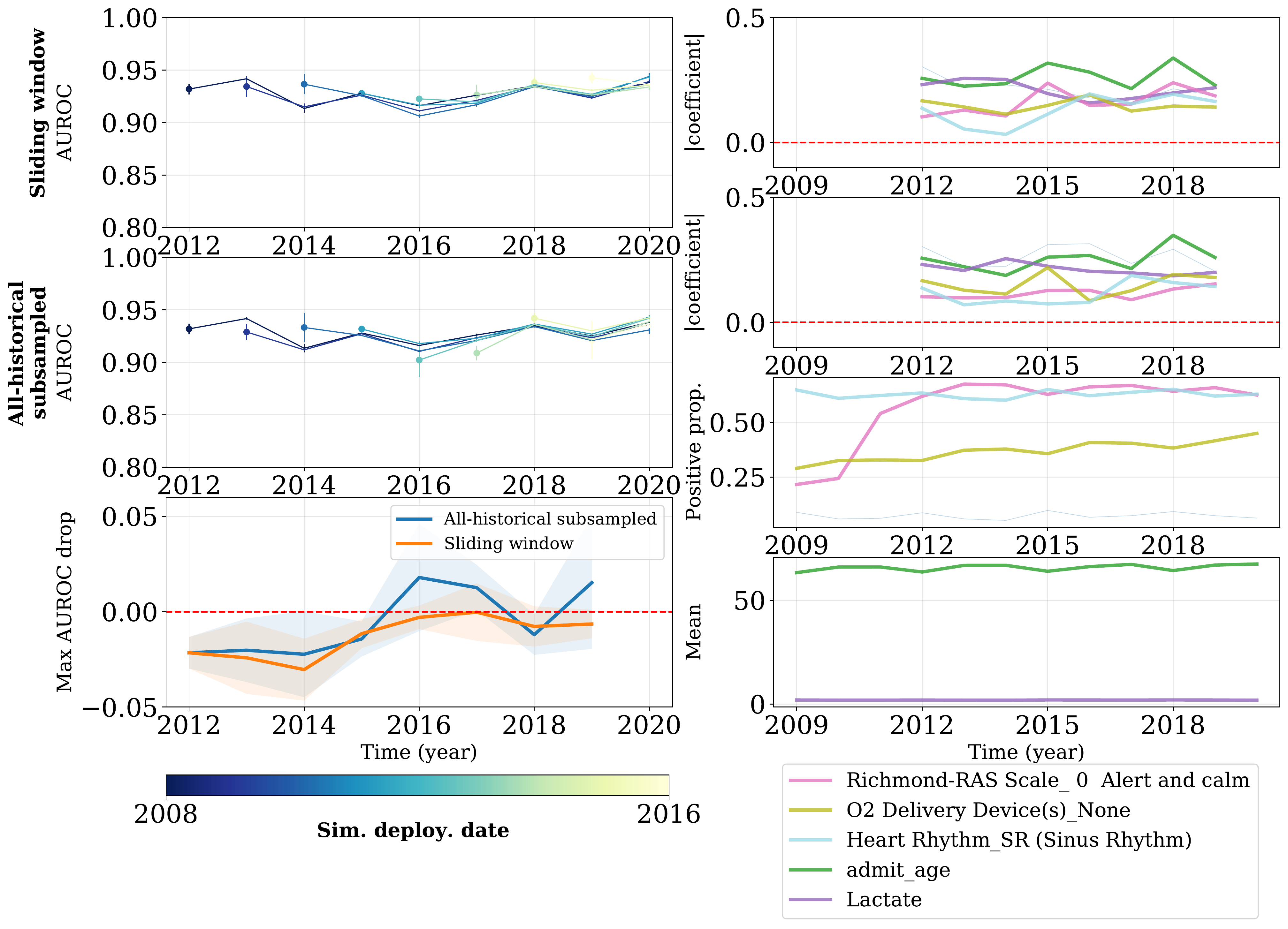}
  \caption{Diagnostic plot of MIMIC-IV. The left column includes absolute AUROC versus time for both sliding window and all-historical subsampled, and the maximum AUROC drop for each trained model. The right column provides the absolute coefficients of each trained model from both regimes, and positive proportion of the significant features over time. The model performance is relatively stable, coinciding with relatively stable distributions of a majority of important features.}
  \label{fig:diagnositic_plot_mimic_iv}
\end{figure}

\clearpage

\subsection{OPTN (Liver)}
\vspace{-2em}
\begin{figure}[H]
  \includegraphics[width=1.0\columnwidth]{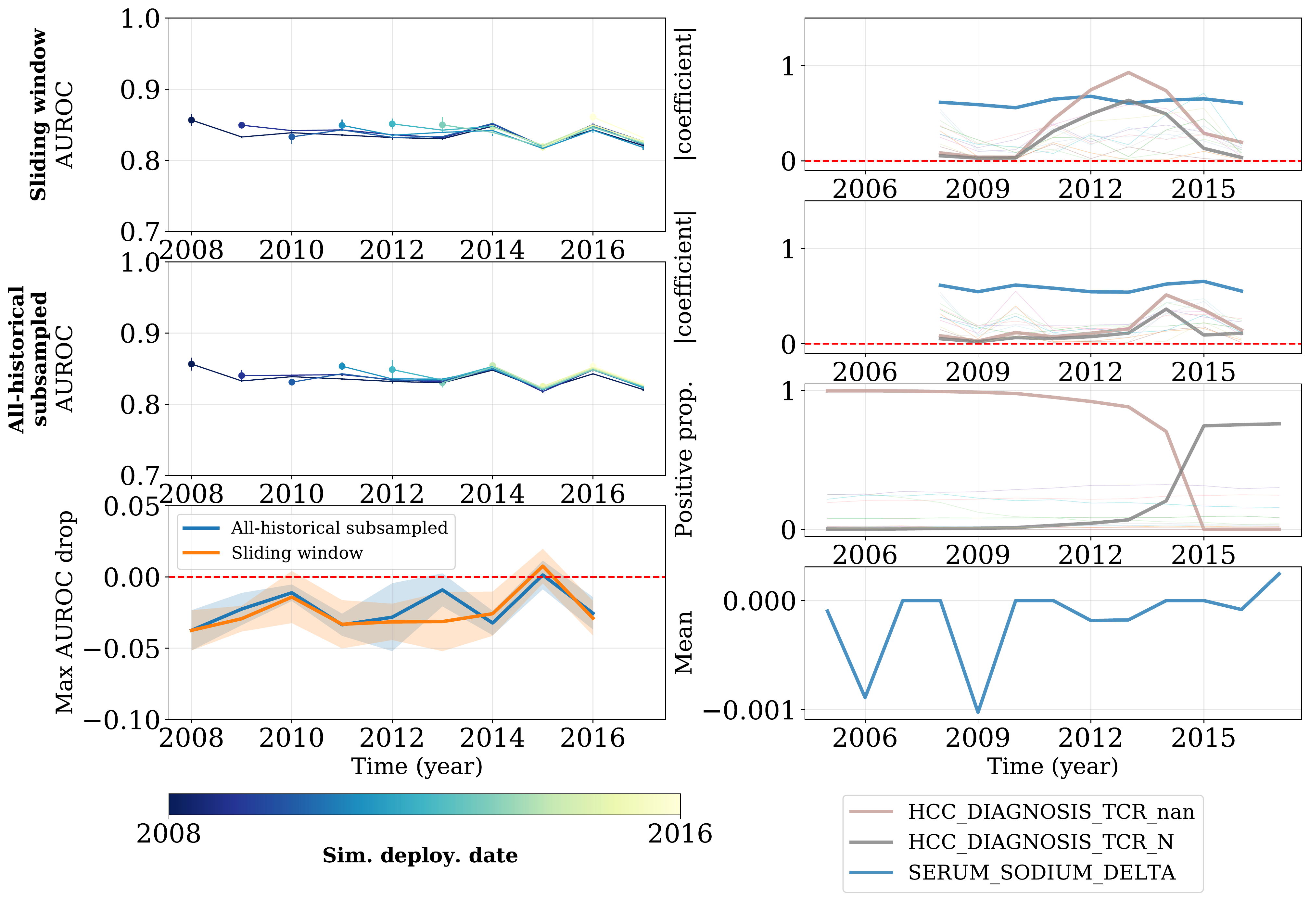}
  \caption{Diagnostic plot of OPTN (Liver). The left column includes absolute AUROC versus time for both sliding window and all-historical subsampled, and the maximum AUROC drop for each trained model. The right column provides the absolute coefficients of each trained model from both regimes, and positive proportion of the significant features over time. Although the HCC DIAGNOSIS TCR binary features change in positive proportion over time, these features were not always important, and the other important features (faded) maintain relatively stable proportions across time. Overall, model performance is quite stable over time.}
  \label{fig:diagnositic_plot_optn_liver}
\end{figure}

\section{Model performance over time (different metrics)}
\label{app:alternative_metrics}

\subsection{AUROC}
\begin{figure}[H]
  \includegraphics[width=0.85\columnwidth]{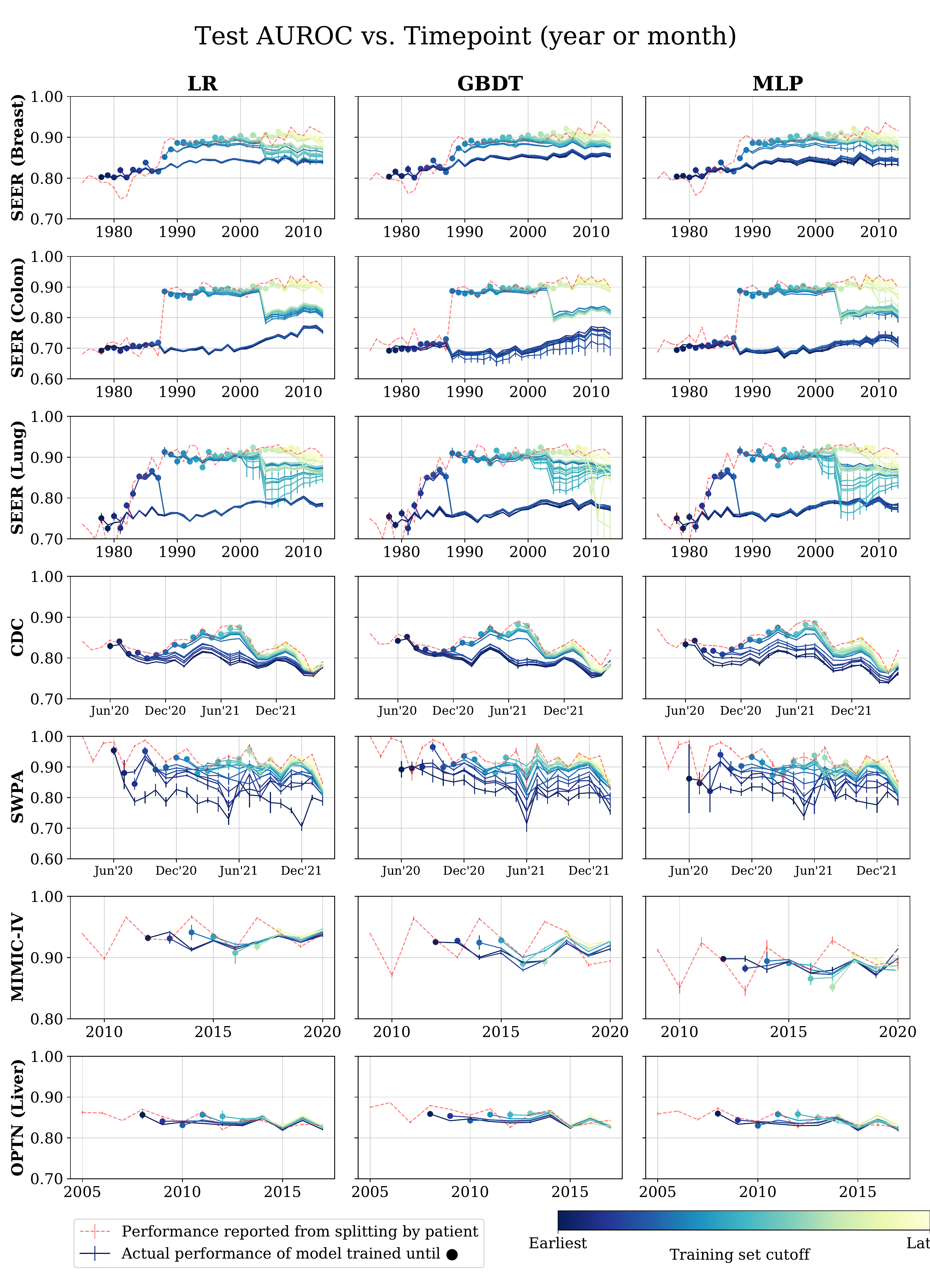}
  \caption{Absolute AUROC versus test timepoints from three model classes on all datasets.}
  \label{fig:auc_over_time}
\end{figure}

\subsection{AUPRC}
\begin{figure}[H]
  \includegraphics[width=0.85\columnwidth]{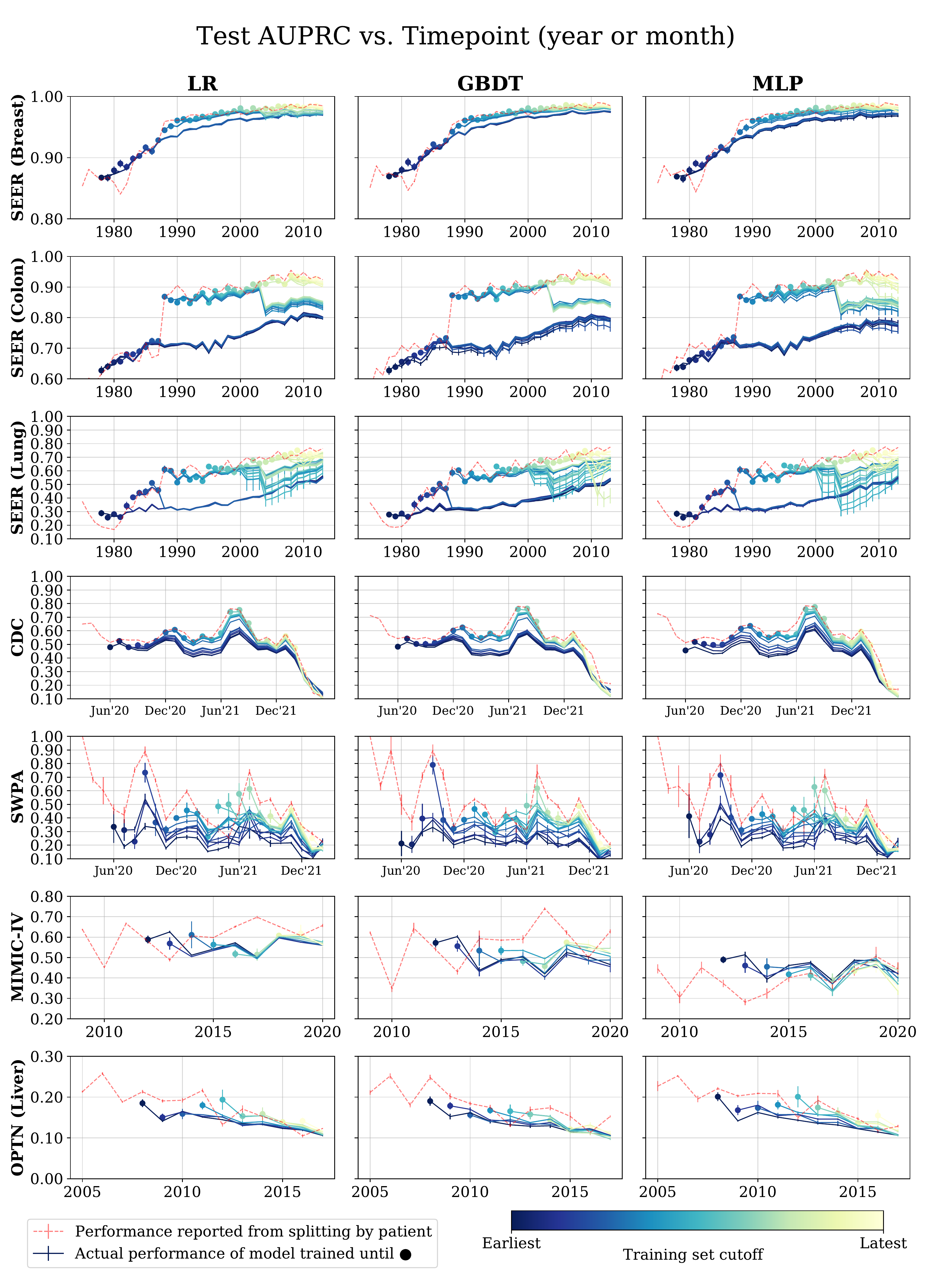}
  \caption{Absolute AUPRC versus test timepoints from three model classes on all datasets.}
  \label{fig:auprc_over_time}
\end{figure}

\subsection{Accuracy}
\begin{figure}[H]
  \includegraphics[width=0.85\columnwidth]{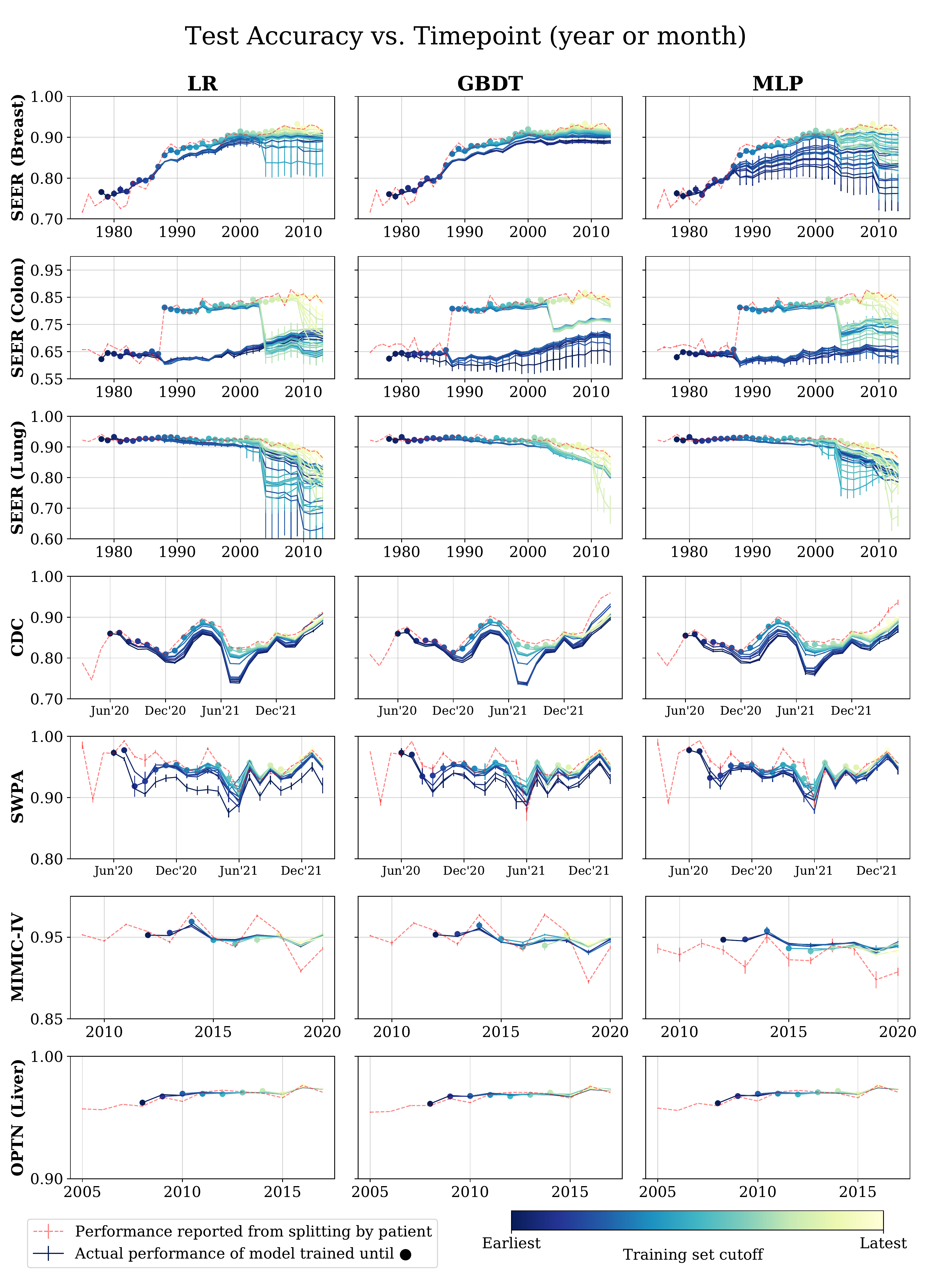}
  \caption{Absolute accuracy versus test timepoints from three model classses on all datasets.}
  \label{fig:acc_over_time}
\end{figure}

\clearpage

\subsection{Recall score}
\begin{figure}[H]
  \includegraphics[width=0.85\columnwidth]{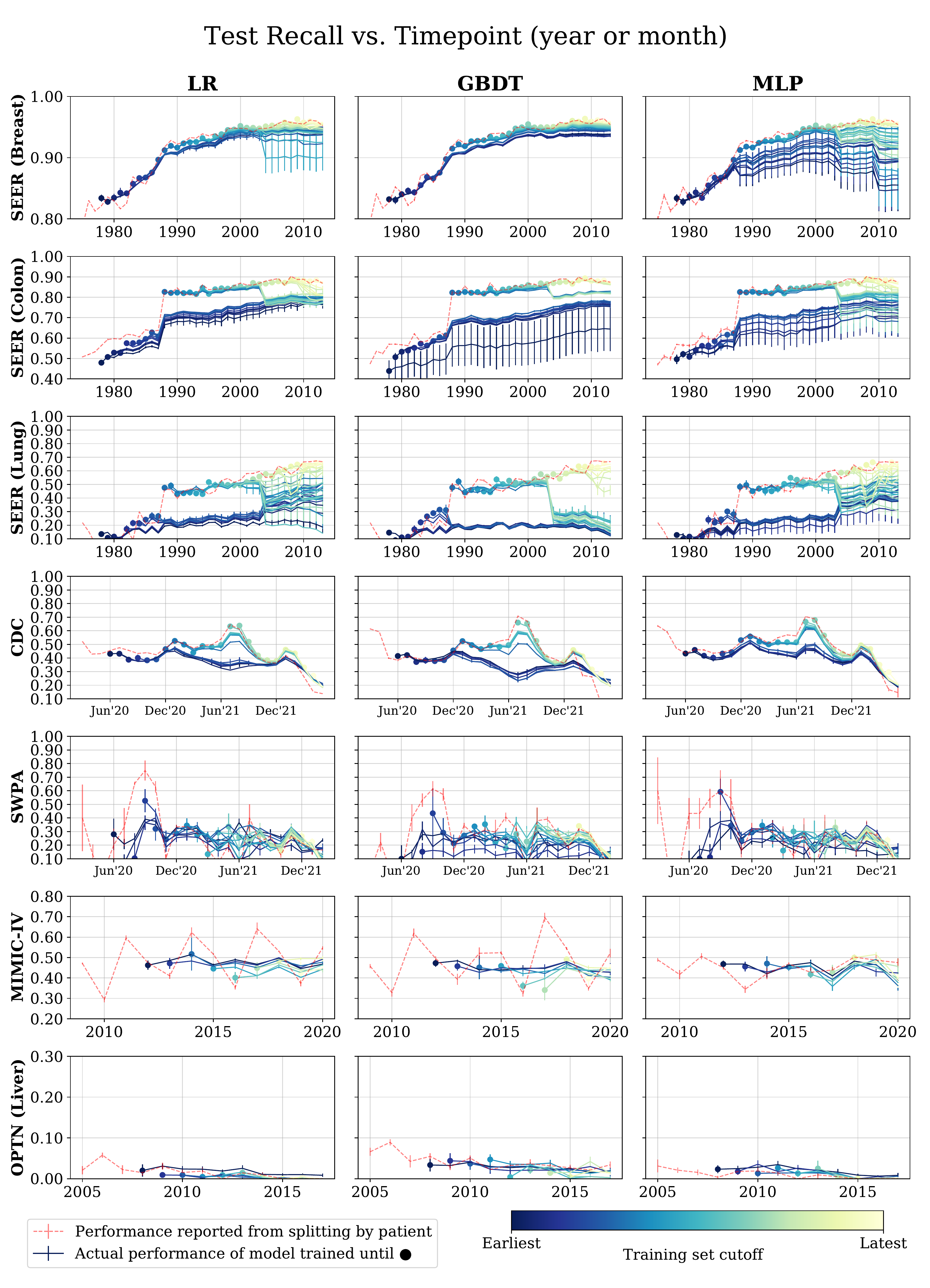}
  \caption{Absolute recall score versus test timepoints from three model classes on all datasets.}
  \label{fig:f1_over_time}
\end{figure}

\clearpage

\section{Hyperparameter Grids}\label{app:hyperparameter_grids}
\begin{table*}[ht]
\caption{Hyperparameter grid for model training.}
\label{tab:hyper_grid}
\centering
    \begin{tabular}{lc}
    \toprule
    Parameter & List of searched value \\
    \midrule
    \textbf{LR} & \\
    \hspace{1em} C & 0.01, 0.1, 1, 10, $10^2$, $10^3$, $10^4$, $10^5$ \\
    \textbf{GBDT} & \\
    \hspace{1em} n\_estimators & 50, 100 \\
    \hspace{1em} max\_depth & 3, 5 \\
    \hspace{1em} learning\_rate & 0.01, 0.1 \\
    \textbf{MLP} & \\
    \hspace{1em} hidden\_layer\_sizes & (3,), (5,) \\
    \hspace{1em} learning\_rate\_init & $10^{-4}$, $10^{-3}$, 0.01 \\
    \bottomrule
    \end{tabular}
\end{table*}

\section{Data Split Details} \label{app:datasplit_details}
\begin{table*}[ht]
\caption{Split ratio for each dataset for training, validation and testing}
\label{tab:datasplits}
\centering
    \begin{tabular}{lc}
    \toprule
    Dataset & Split ratio \\
    \midrule
    SEER (Breast) & 0.8-0.1-0.1 \\
    SEER (Colon) & 0.8-0.1-0.1 \\
    SEER (Lung) & 0.8-0.1-0.1 \\
    CDC COVID-19 & 0.8-0.1-0.1 \\
    SWPA COVID-19 & 0.5-0.25-0.25 \\
    MIMIC-IV & 0.5-0.25-0.25 \\
    OPTN (Liver) & 0.5-0.25-0.25 \\
    \bottomrule
    \end{tabular}
\end{table*}

\clearpage
\section{Complete Results for All Datasets}
\label{app:all_result}
\begin{table*}[htb]
\caption{Test AUROC from time-agnostic evaluation (all-period training).}
\label{tab:AUROC_splitting_by_patient}
\centering
    \resizebox{1\columnwidth}{!}{%
    \begin{tabular}{lccccccc}
    \toprule
    Model & SEER (Breast) & SEER (Colon) & SEER (Lung) & CDC & SWPA & MIMIC-IV & OPTN (Liver) \\
    \midrule
    \textbf{LR} &         0.887 &        0.867 &       0.896 &        0.837 &         0.914 &    \textbf{0.936} &        0.845 \\
    \textbf{GBDT} &          \textbf{0.890} &        0.871 &       0.895 &         \textbf{0.850} &         \textbf{0.926} &    0.928 &        \textbf{0.858} \\
    \textbf{MLP} &          \textbf{0.890} &        \textbf{0.873} &         \textbf{0.900} &        0.844 &         0.918 &    0.885 &        0.846 \\
    \bottomrule
    \end{tabular}%
    }
    \vspace{-0.5em}
\end{table*}

\begin{figure*}[ht]
  \includegraphics[width=1\columnwidth]{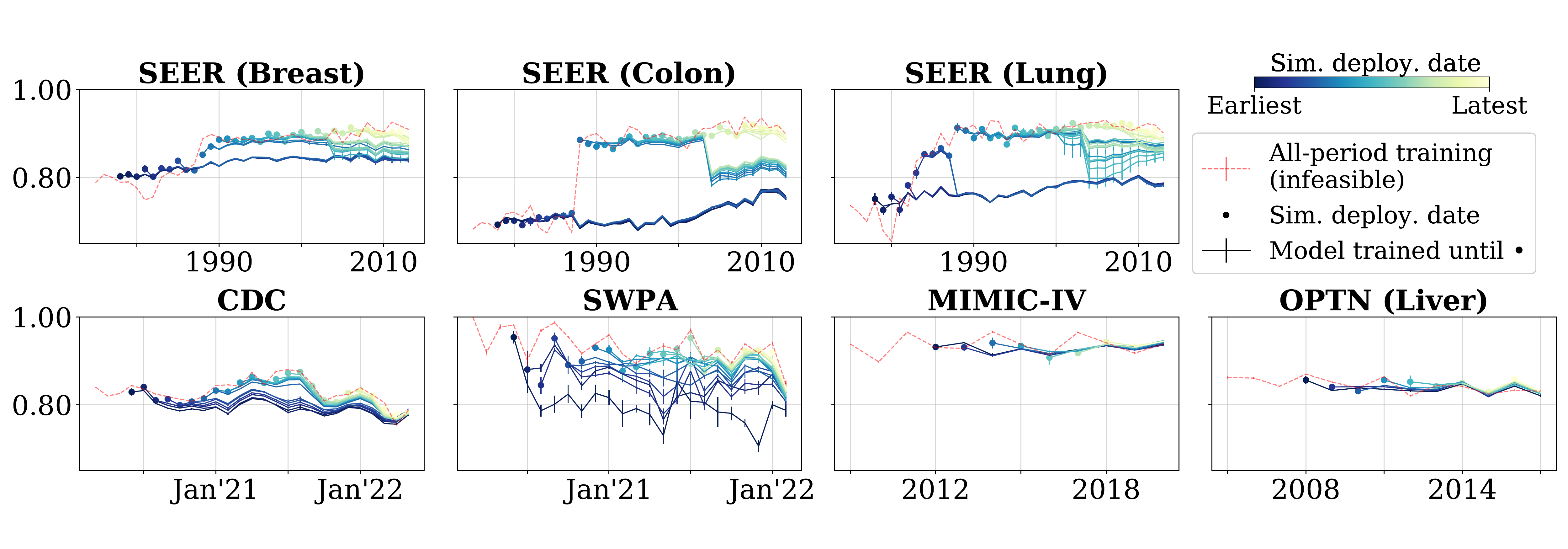}
  \vspace{-1em}
  \caption{Average Test AUROC vs. time for LR. (GBDT and MLP plots in Appendix \ref{app:alternative_metrics}.) Each solid line 
  gives the performance of a model trained up to a simulated deployment time (marked by a dot), and evaluated across future time points. 
  Error bars are the standard deviation of average test AUROC, computed over 5 random splits. The red dotted line gives per-time point test performance of a model from all-period training (infeasible, as it would train on data after the simulated deployment date). As the red dotted line mostly sits above the performance of any model that could have been previously deployed, standard all-period training tends to provide an over-optimistic estimate.}
  \label{fig:absolute_auc_over_time}
  \vspace{-1em}
\end{figure*}

\begin{figure}[ht]
  \centering
  \includegraphics[width=0.6\columnwidth]{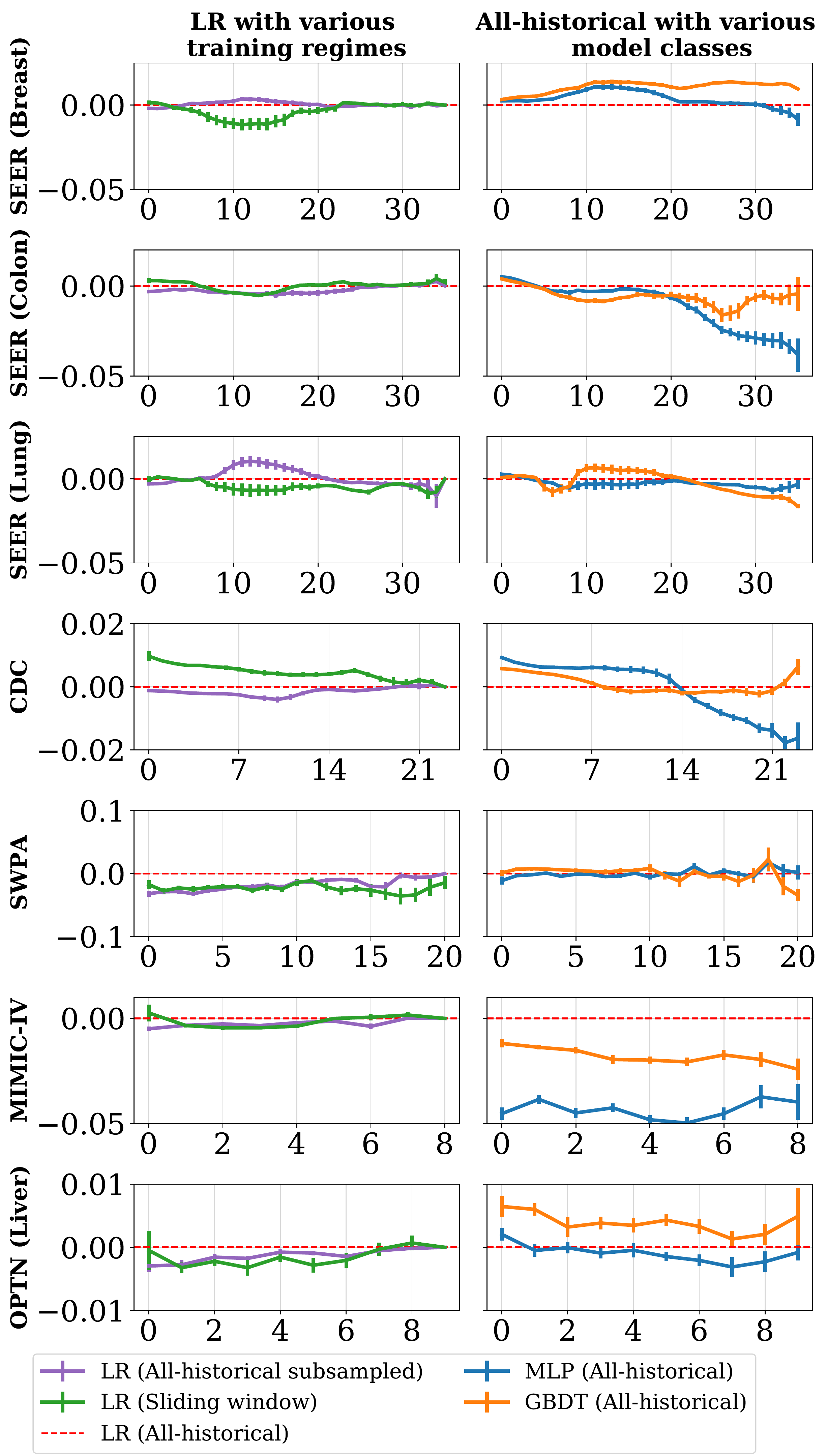}
  \vspace{-1em}
  \caption{
  $\text{AUROC} - \text{AUROC}_{\text{LR All-historical}}$ 
  across varying stalenesses of data, for different training regimes (left) and model classes (right). Error bars are $\pm$ std. dev.}
  \label{fig:window_model_comparison}
  \vspace{-2.5em}
\end{figure}

\end{document}